\documentclass[journal]{IEEEtran}
\usepackage{amsmath,bm}

\usepackage{cite}
\usepackage{multirow}
\usepackage{diagbox}
\usepackage{lipsum}
\usepackage{graphicx}

\usepackage{enumitem}
\usepackage{amssymb}
\usepackage{booktabs}

\usepackage{hyperref}
\usepackage{color,soul}
\usepackage[dvipsnames]{xcolor}

\usepackage[ruled,linesnumbered]{algorithm2e}

\ifCLASSOPTIONcompsoc
\usepackage[caption=false,font=normalsize,labelfon
t=sf,textfont=sf]{subfig}
\else
\usepackage[caption=false,font=footnotesize]{subfi
g}
\fi
\usepackage{alphalph}
\usepackage{tabularx,ragged2e}
\usepackage{threeparttable}

\usepackage{hyperref} 

\ifCLASSINFOpdf
\else
\fi
\begin{document}
%
\title{UGC-VQA: Benchmarking Blind Video Quality Assessment for User Generated Content}
%
%
%
\author{Zhengzhong~Tu,
        Yilin~Wang,
        Neil~Birkbeck,
        Balu~Adsumilli,
        and~Alan~C.~Bovik,~\IEEEmembership{Fellow,~IEEE}
\thanks{Z. Tu and A. C. Bovik are with Laboratory for Image and Video Engineering (LIVE), Department of Electrical and Computer Engineering, The University of Texas at Austin, Austin, TX, 78712, USA (email: zhengzhong.tu@utexas.edu, bovik@utexas.edu).}
\thanks{Y. Wang, N. Birkbeck, and B. Adsumilli are with YouTube Media Algorithms Team, Google LLC, Mountain View, CA, 94043, USA. (email: yilin@google.com, birkbeck@google.com, 
badsumilli@google.com)}
\thanks{This work is supported by Google.}}
\maketitle

\begin{abstract}
Recent years have witnessed an explosion of user-generated content (UGC) videos shared and streamed over the Internet, thanks to the evolution of affordable and reliable consumer capture devices, and the tremendous popularity of social media platforms. Accordingly, there is a great need for accurate video quality assessment (VQA) models for UGC/consumer videos to monitor, control, and optimize this vast content. Blind quality prediction of in-the-wild videos is quite challenging, since the quality degradations of UGC videos are unpredictable, complicated, and often commingled. Here we contribute to advancing the UGC-VQA problem by conducting a comprehensive evaluation of leading no-reference/blind VQA (BVQA) features and models on a fixed evaluation architecture, yielding new empirical insights on both subjective video quality studies and objective VQA model design. By employing a feature selection strategy on top of efficient BVQA models, we are able to extract 60 out of 763 statistical features used in existing methods to create a new fusion-based model, which we dub the \textbf{VID}eo quality \textbf{EVAL}uator (VIDEVAL), that effectively balances the trade-off between VQA performance and efficiency. Our experimental results show that VIDEVAL achieves state-of-the-art performance at considerably lower computational cost than other leading models. Our study protocol also defines a reliable benchmark for the UGC-VQA problem, which we believe will facilitate further research on deep learning-based VQA modeling, as well as perceptually-optimized efficient UGC video processing, transcoding, and streaming. To promote reproducible research and public evaluation, an implementation of VIDEVAL has been made available online: \url{https://github.com/vztu/VIDEVAL}.
\end{abstract}


%
\IEEEpeerreviewmaketitle

\section{Introduction}
%
%
%
%
\IEEEPARstart{V}{ideo} dominates the Internet. In North America, Netflix and YouTube alone account for more than fifty percent of downstream traffic, and there are many other significant video service providers. Improving the efficiency of video encoding, storage, and streaming over communication networks is a principle goal of video sharing and streaming platforms. One relevant and essential research direction is the perceptual optimization of rate-distortion tradeoffs in video encoding and streaming, where distortion (or quality) is usually modeled using video quality assessment (VQA) algorithms that can predict human judgements of video quality. This has motivated years of research on the topics of perceptual video and image quality assessment (VQA/IQA).

VQA research can be divided into two closely related categories: subjective video quality studies and objective video quality modeling. Subjective video quality research usually requires substantial resources devoted to time- and labor-consuming human studies to obtain valuable and reliable subjective data. The datasets obtained from subjective studies are invaluable for the development, calibration, and benchmarking of objective video quality models that are consistent with subjective mean opinion scores (MOS).
\begin{table*}[!t]
\setlength{\tabcolsep}{2.5pt}
\renewcommand{\arraystretch}{1.1}
\centering
\begin{threeparttable}
\caption{Evolution of popular public video quality assessment databases: from legacy lab studies of synthetically distorted video sets to large-scale crowdsourced user-generated content (UGC) video datasets with authentic distortions}
\label{table:db_comp}
\begin{tabular}{llllllllp{3.5cm}lllll}
\toprule
\textsc{Database}          & \textsc{Year} & \textsc{\#Cont} & \textsc{\#Total} & \textsc{Resolution} & \textsc{FR} & \textsc{Len} & \textsc{Format}      & \textsc{Distortion Type}                                                                  & \textsc{\#Subj} & \textsc{\#Rates}  & \textsc{Data}          & \textsc{Env}   \\
\hline\\[-1.em]
LIVE-VQA          & 2008 & 10         & 160      & 768$\times$432    & 25/50     & 10           & YUV+264 & Compression, transmission                                                   & 38       & 29       & DMOS+$\sigma$ & In-lab        \\
EPFL-PoliMI          & 2009 & 12         & 156      & CIF/4CIF    & 25/30     & 10           & YUV+264 & Compression, transmission                                                   & 40       & 34       & MOS & In-lab        \\
VQEG-HDTV         & 2010 & 49         & 740      & 1080i/p    & 25/30     & 10           & AVI         & Compression, transmission                                                   & 120      & 24       & RAW           & In-lab        \\
 IVP               & 2011 & 10         & 138      & 1080p      & 25        & 10           & YUV         & Compression, transmission                                                   & 42       & 35       & DMOS+$\sigma$ & In-lab        \\
TUM 1080p50       & 2012 & 5          & 25       & 1080p      & 50        & 10           & YUV         & Compression                                                                 & 21       & 21       & MOS           & In-lab        \\
CSIQ              & 2014 & 12         & 228      & 832$\times$480    & 24-60     & 10           & YUV         & Compression, transmission                                                   & 35       & N/A        & DMOS+$\sigma$ & In-lab        \\
CVD2014           & 2014 & 5          & 234      & 720p, 480p & 9-30      & 10-25        & AVI         & Camera capture (authentic)                                                & 210      & 30 & MOS           & In-lab        \\
MCL-V             & 2015 & 12         & 108      & 1080p      & 24-30     & 6            & YUV         & Compression, scaling                                                        & 45       & 32       & MOS           & In-lab        \\
MCL-JCV           & 2016 & 30         & 1560     & 1080p      & 24-30     & 5            & MP4         & Compression                                                                 & 150      & 50       & RAW-JND       & In-lab        \\
KoNViD-1k         & 2017 & 1200       & 1200     & 540p & 24-30     & 8            & MP4         & Diverse distortions (authentic)                                             & 642      & 114       & MOS+$\sigma$  & Crowd \\
LIVE-Qualcomm     & 2018 & 54         & 208      & 1080p      & 30        & 15           & YUV         & Camera capture (authentic)                                                & 39       & 39       & MOS           & In-lab        \\
LIVE-VQC          & 2018 & 585        & 585      & 1080p-240p & 19-30     & 10           & MP4         & Diverse distortions (authentic)                                             & 4776     & 240      & MOS           & Crowd \\
YouTube-UGC       & 2019 & 1380       & 1380     & 4k-360p    & 15-60     & 20           & MKV         & Diverse distortions (authentic)                                             &    $>$8k      &     123     &        MOS+$\sigma$       &   Crowd   \\
\bottomrule
\end{tabular}
\begin{tablenotes}[para,flushleft]
\footnotesize
\item \textsc{\#Cont}: Total number of unique contents.
\item \textsc{\#Total}: Total number of test sequences, including reference and distorted videos.
\item \textsc{Resolution}: Video resolution (p: progressive).
\item \textsc{FR}: Framerate. 
\item \textsc{Len}: Video duration/length (in seconds).
\item \textsc{Format}: Video container.
\item \textsc{\#Subj}: Total number of subjects in the study.
\item \textsc{\#Rates}: Average number of subjective ratings per video.
\item \textsc{Env}: Subjective testing environment. In-lab: study was conducted in a laboratory. Crowd: study was conducted by crowdsourcing.
\end{tablenotes}
\end{threeparttable}

\end{table*}

Hence, researchers have devoted considerable efforts on the development of high-quality VQA datasets that benefit the video quality community. Table \ref{table:db_comp} summarizes the ten-year evolution of popular public VQA databases. The first successful VQA database was the LIVE Video Quality Database \cite{seshadrinathan2010study}, which was first made publicly available in 2008. It contains $10$ pristine high-quality videos subjected to compression and transmission distortions. Other similar databases targeting simulated compression and transmission distortions have been released subsequently, including EPFL-PoliMI \cite{de2010h}, VQEG-HDTV \cite{vqeg_hdtv}, IVP \cite{ivp}, TUM 1080p50 \cite{keimel2012tum}, CSIQ \cite{vu2014vis3}, MCL-V \cite{lin2015mcl}, and MCL-JCV \cite{wang2016mcl}. All of the above mentioned datasets are based on a small set of high-quality videos, dubbed ``pristine'' or ``reference,'' then synthetically distorting them in a controlled manner. We will refer to these kinds of synthetically-distorted video sets as \textit{legacy} VQA databases. Legacy databases are generally characterized by only a small number of unique contents, each simultaneously degraded by only one or at most two synthetic distortions. For most practical scenarios, these are too simple to represent the great variety of real-world videos, and hence, VQA models derived on these databases may be insufficiently generalizable to large-scale realistic commercial VQA applications. 

Recently, there has been tremendous growth in social media, where huge volumes of  user-generated content (UGC) is shared over the media platforms such as YouTube, Facebook, and TikTok. Advances in powerful and affordable mobile devices and cloud computing techniques, combined with significant advances in video streaming have made it easy for most consumers to create, share, and view UGC pictures/videos instantaneously across the globe. Indeed, the prevalence of UGC has started to shift the focus of video quality research from \textit{legacy} synthetically-distorted databases to newer, larger-scale authentic UGC datasets, which are being used to create solutions to what we call the \textbf{\textsf{UGC-VQA problem}}. UGC-VQA studies typically follow a new design paradigm whereby: 1) All the source content is consumer-generated instead of professional-grade, thus suffers from unknown and highly diverse impairments; 2) they are only suitable for testing and comparing no-reference models, since reference videos are unavailable; 3) the types of distortions are authentic and commonly intermixed, and include but are not limited to capture impairments, editing and processing artifacts, compression, transcoding, and transmission distortions. Moreover, compression artifacts are not necessarily the dominant factors affecting video quality, unlike legacy VQA datasets and algorithms. These unpredictable perceptual degradations make perceptual quality prediction of UGC consumer videos very challenging.

Here we seek to address and gain insights into this new challenge (UGC-VQA) by first, conducting a comprehensive benchmarking study of leading video quality models on several recently released large-scale UGC-VQA databases. We also propose a new fusion-based blind VQA (BVQA) algorithm, which we call the VIDeo quality EVALuator (VIDEVAL), which is created by the processes of feature selection from existing top-performing VQA models. The empirical results show that a simple aggregation of these known models can achieve state-of-the-art (SOTA) performance. We believe that our expansive study will inspire and drive future research on BVQA modeling for the challenging UGC-VQA problem, and also pave the way towards deep learning-based solutions. 

The outline of this paper is as follows: Section \ref{sec:ugc_db} reviews and analyzes the three most recent large-scale UGC-VQA databases, while Section \ref{sec:nr_vqa} briefly surveys the development of BVQA models. We introduce the proposed VIDEVAL model in Section \ref{sec:fs}, and provide experimental results in Section \ref{sec:exp}. Finally, concluding remarks are given in Section \ref{sec:conclud}.

\begin{figure*}[!t]
\def\xheight{0.197}
\centering
\subfloat[LIVE-VQC][{LIVE-VQC}]{\includegraphics[height=\xheight\textwidth]{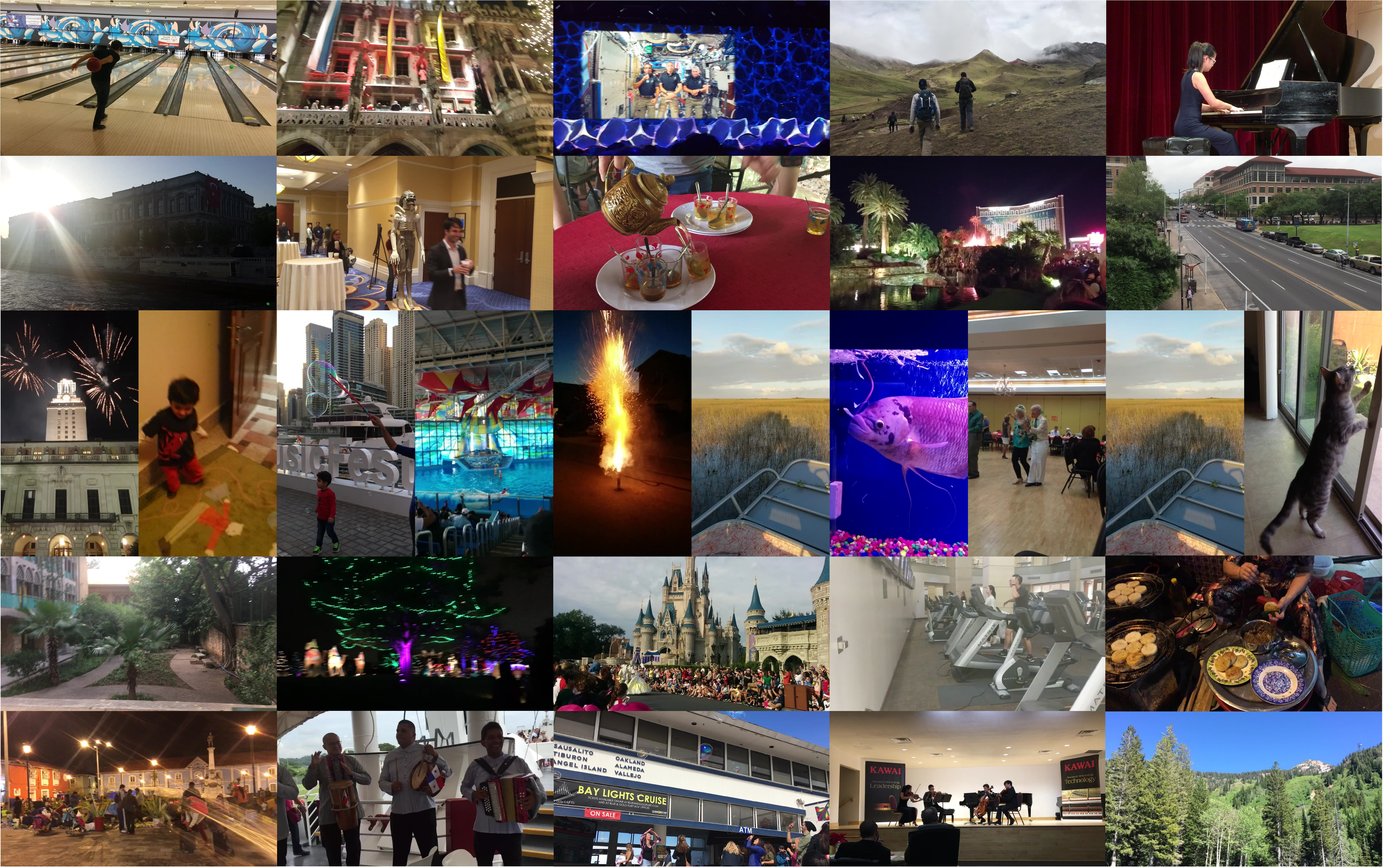} 
\label{fig:snapshota}}
\hspace{-0.284em}
\subfloat[KoNViD-1k][{KoNViD-1k}]{\includegraphics[height=\xheight\textwidth]{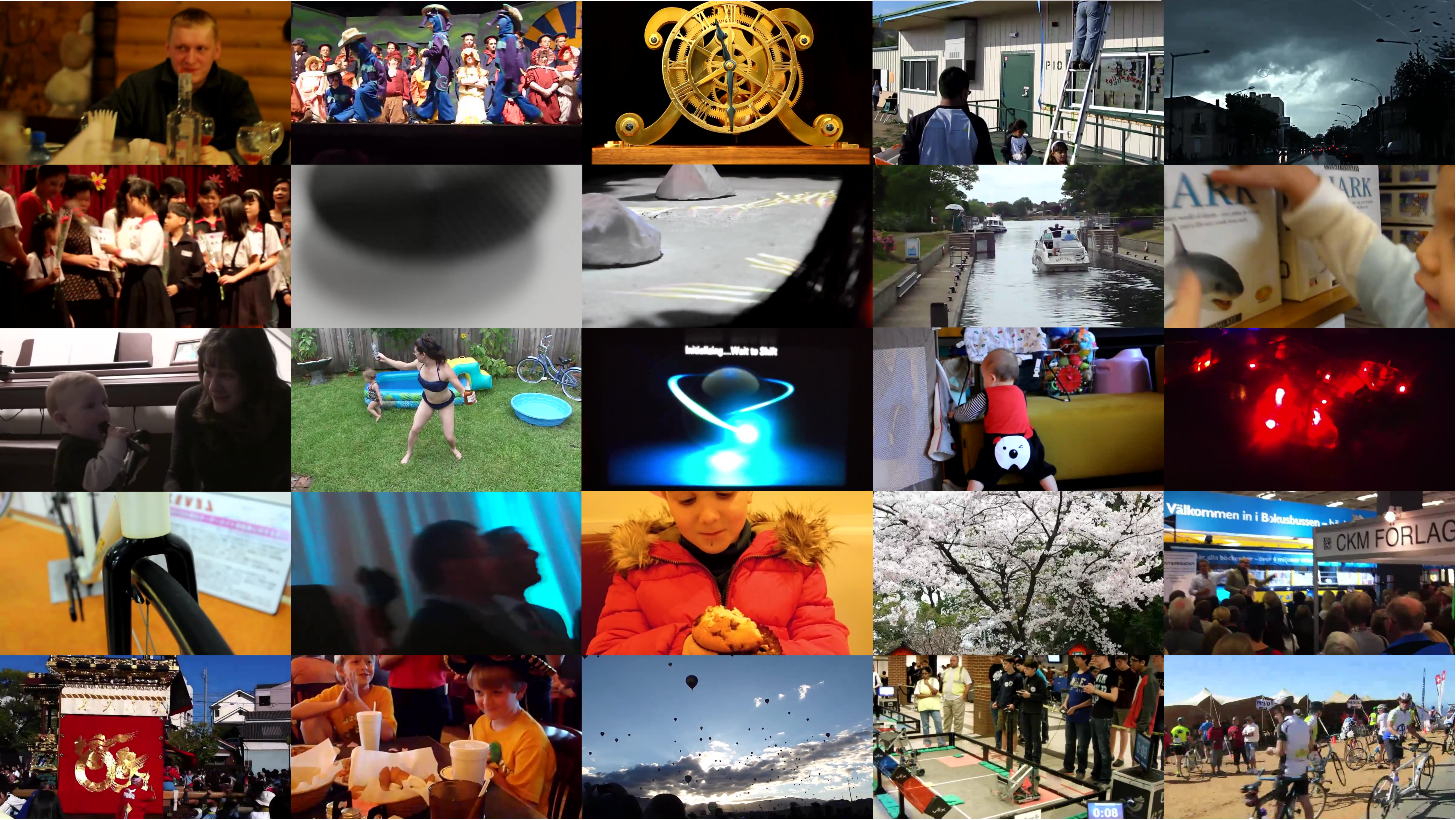} 
\label{fig:snapshotb}} 
\hspace{-0.284em}
\subfloat[YouTube-UGC][{YouTube-UGC}]{\includegraphics[height=\xheight\textwidth]{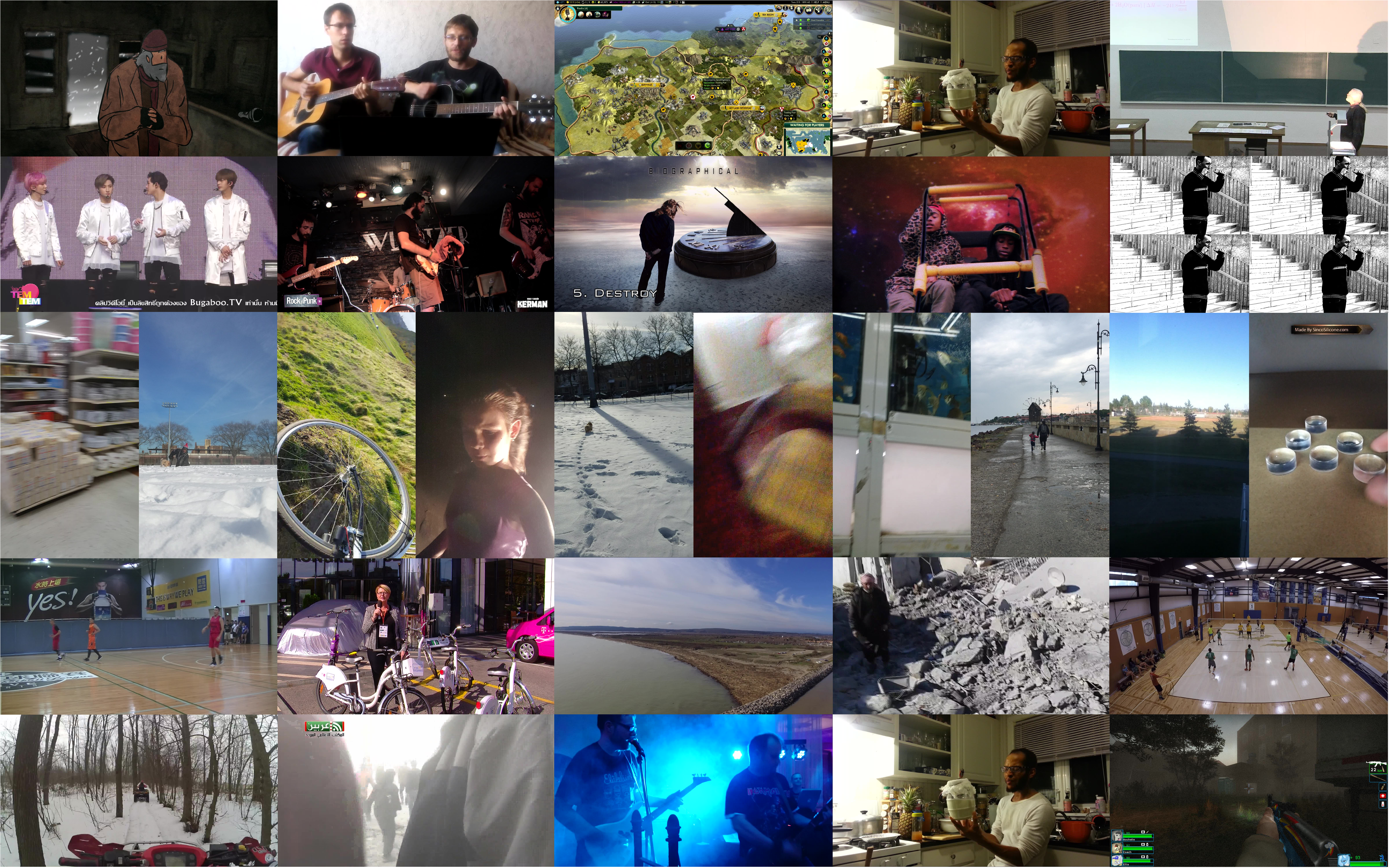} 
\label{fig:snapshotc}}
\caption{Sample frames of the video contents contained in the three large scale UGC-VQA databases: (a) LIVE-VQC \cite{sinno2018large}, (b) KoNViD-1k \cite{hosu2017konstanz}, and (c) YouTube-UGC \cite{wang2019youtube}. LIVE-VQC includes only natural contents captured by mobile devices, while KoNViD-1k and YouTube-UGC comprise of both natural videos, animations, and gaming sources. Note that YouTube-UGC video set is categorized whereas the others are not.}
\label{fig:snapshot}
\end{figure*}

\begin{table*}[!t]
\setlength{\tabcolsep}{4pt}
\renewcommand{\arraystretch}{1.1}
\centering
\caption{Public large-scale user-generated content video quality assessment (UGC-VQA) databases compared: KoNViD-1k \cite{hosu2017konstanz}, LIVE-VQC \cite{sinno2018large}, and YouTube-UGC \cite{wang2019youtube}}
\label{table:ugc_db_comp}
\begin{tabular}{lccc}
\toprule
\textsc{Database Attribute} & KoNViD-1k & LIVE-VQC  & YouTube-UGC  \\ 
\hline\\[-1.em]
Number of contents & 1200 & 585  & 1380 \\
Video sources & YFCC100m (Flickr) & Captured (mobile devices) & YouTube \\
Video resolutions & 540p & 1080p,720p,480p,etc. & 4k,1080p,720p,480p,360p \\
Video layouts & Landscape & Landscape,portrait & Landscape,portrait \\
Video framerates & 24,25,30 fr/sec & 20,24,25,30 fr/sec & 15,20,24,25,30,50,60 fr/sec \\
Video lengths & 8 seconds & 10 seconds & 20 seconds \\
Audio track included & Yes (97\%) & Yes & No \\
Testing methodology & Crowdsourcing (CrowdFlower) & Crowdsourcing (AMT) & Crowdsourcing (AMT) \\
Number of subjects & 642 & 4,776 & $>$8,000 \\
Number of ratings & 136,800 (114 votes/video) & 205,000 (240 votes/video) & 170,159 (123 votes/video) \\
Rating scale & Absolute Category Rating 1-5 & Continuous Rating 0-100 & Continuous Rating 1-5 \\
\multirow[t]{6}{*}{Content remarks} & \multirow[t]{6}{4.6cm}{Videos sampled from YFCC100m via a feature space of blur, colorfulness, contrast, SI, TI, and NIQE; Some contents irrelevant to quality research; Content was clipped from the original and resized to 540p.}  & \multirow[t]{6}{4.6cm}{Videos manually captured by certain people; Content including many camera motions; Content including some night scenes that are prone to be outliers; Resolutions not uniformly distributed.} & \multirow[t]{6}{4.6cm}{Videos sampled from YouTube via a feature space of spatial, color, temporal, and chunk variation; Contents categorized into $15$ classes, including HDR, screen content, animations, and gaming videos.}  \\ \\ \\ \\ \\ \\ 
\multirow[t]{5}{*}{Study remarks} & \multirow[t]{5}{4.6cm}{Study did not account for or remove videos on which stalling events occurred when viewed; test methodology prone to unreliable individual scores.}  & \multirow[t]{5}{4.6cm}{Distribution of MOS values slightly skewed towards higher scores; standard deviation statistics of MOS were not provided.} & \multirow[t]{5}{4.6cm}{Distribution of MOS values slightly skewed towards higher values; three additional chunk MOS scores with standard deviation were provided.} \\ \\ \\ \\ \\ 
\bottomrule

\end{tabular}
\end{table*}

\section{UGC-VQA Databases}
\label{sec:ugc_db}

The first UGC-relevant VQA dataset containing authentic distortions was introduced as the Camera Video Database (CVD2014) \cite{nuutinen2016cvd2014}, which consists of videos with in-the-wild distortions from 78 different video capture devices, followed by the similar LIVE-Qualcomm Mobile In-Capture Database \cite{ghadiyaram2017capture}. These two databases, however, only modeled (camera) capture distortions on small numbers of not very diverse unique contents. Inspired by the first successful massive online crowdsourcing study of UGC picture quality \cite{ghadiyaram2015massive}, the authors of \cite{hosu2017konstanz} created the KoNViD-1k video quality database, the first such resource for UGC videos. It consists of 1,200 public-domain videos sampled from the YFCC100M dataset \cite{thomee2015yfcc100m}, and was annotated by 642 crowd-workers. LIVE-VQC \cite{sinno2018large} was another large-scale UGC-VQA database with 585 videos, crowdsourced on Amazon Mechanical Turk to collect human opinions from 4,776 unique participants. The most recently published UGC-VQA database is the YouTube-UGC Dataset \cite{wang2019youtube} comprising 1,380 20-second video clips sampled from millions of YouTube videos, which were rated by more than 8,000 human subjects. Table \ref{table:ugc_db_comp} summarizes the main characteristics of the three large-scale UGC-VQA datasets studied, while Figure \ref{fig:snapshot} shows some representative snapshots of the source sequences for each database, respectively.

\subsection{Content Diversity and MOS Distribution}

\begin{figure*}[!t]
\captionsetup[subfigure]{justification=centering}
\centering
\def\xwidth{0.133}
\def\hswidth{-0.4em}
\subfloat[Brightness][{Brightness}]{\includegraphics[height=\xwidth\textwidth]{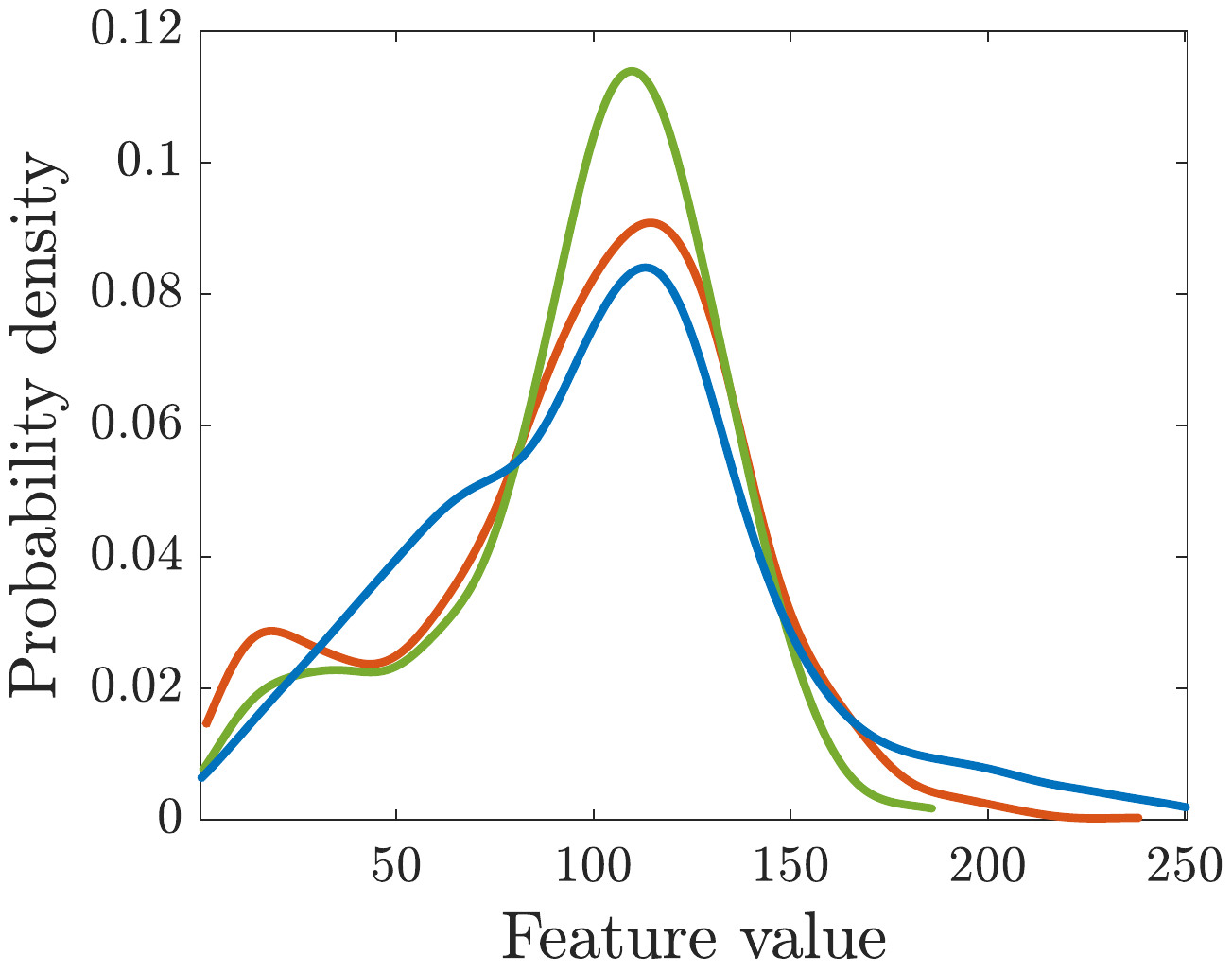} 
\label{fig:ind_feat_disa}} 
\hspace{\hswidth}
\subfloat[Contrast][{Contrast}]{\includegraphics[height=\xwidth\textwidth]{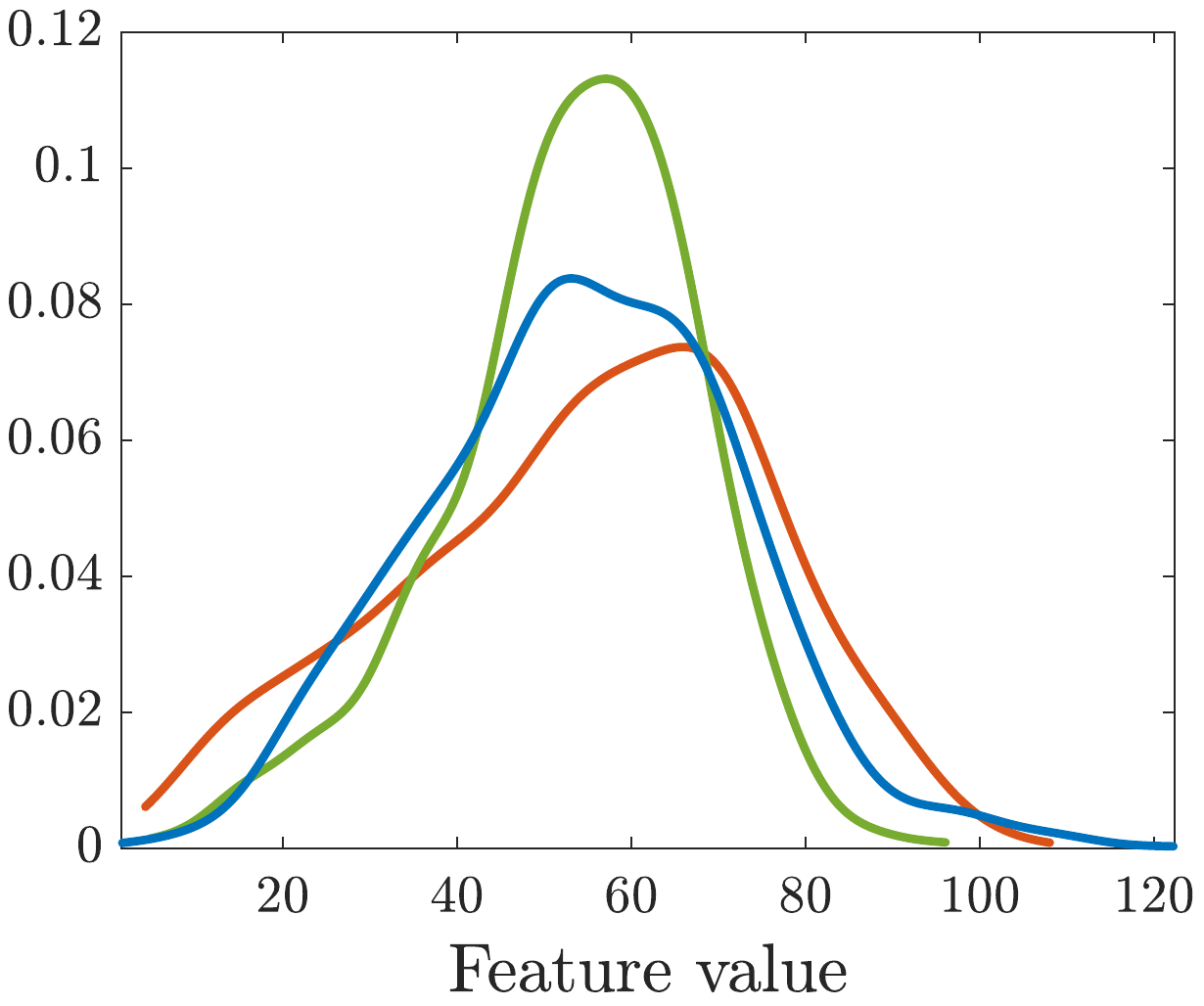} 
\label{fig:ind_feat_disb}} 
\hspace{\hswidth}
\subfloat[Colorfulness][{Colorfulness}]{\includegraphics[height=\xwidth\textwidth]{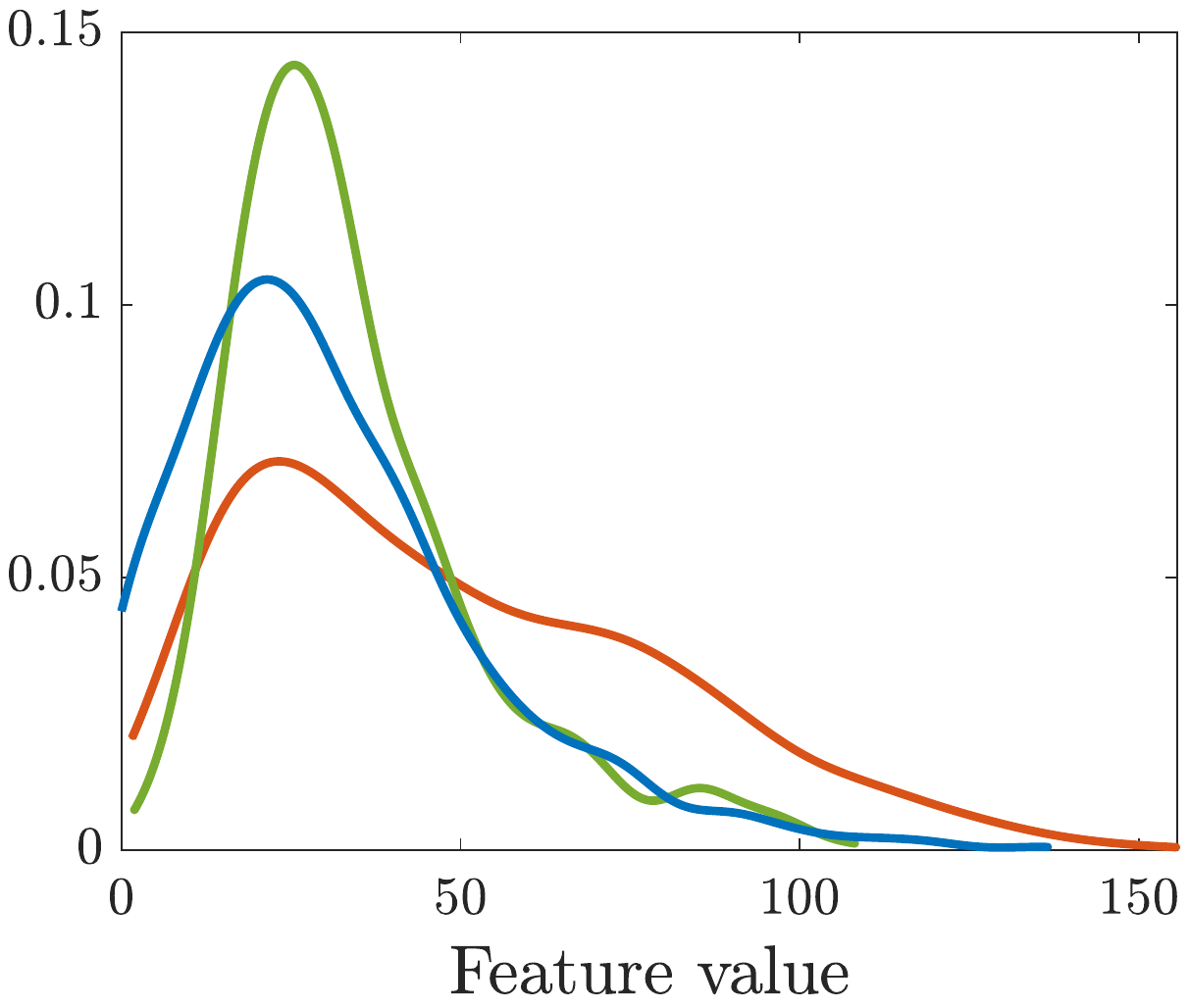} 
\label{fig:ind_feat_disc}}
\hspace{\hswidth}
\subfloat[Sharpness][{Sharpness}]{\includegraphics[height=\xwidth\textwidth]{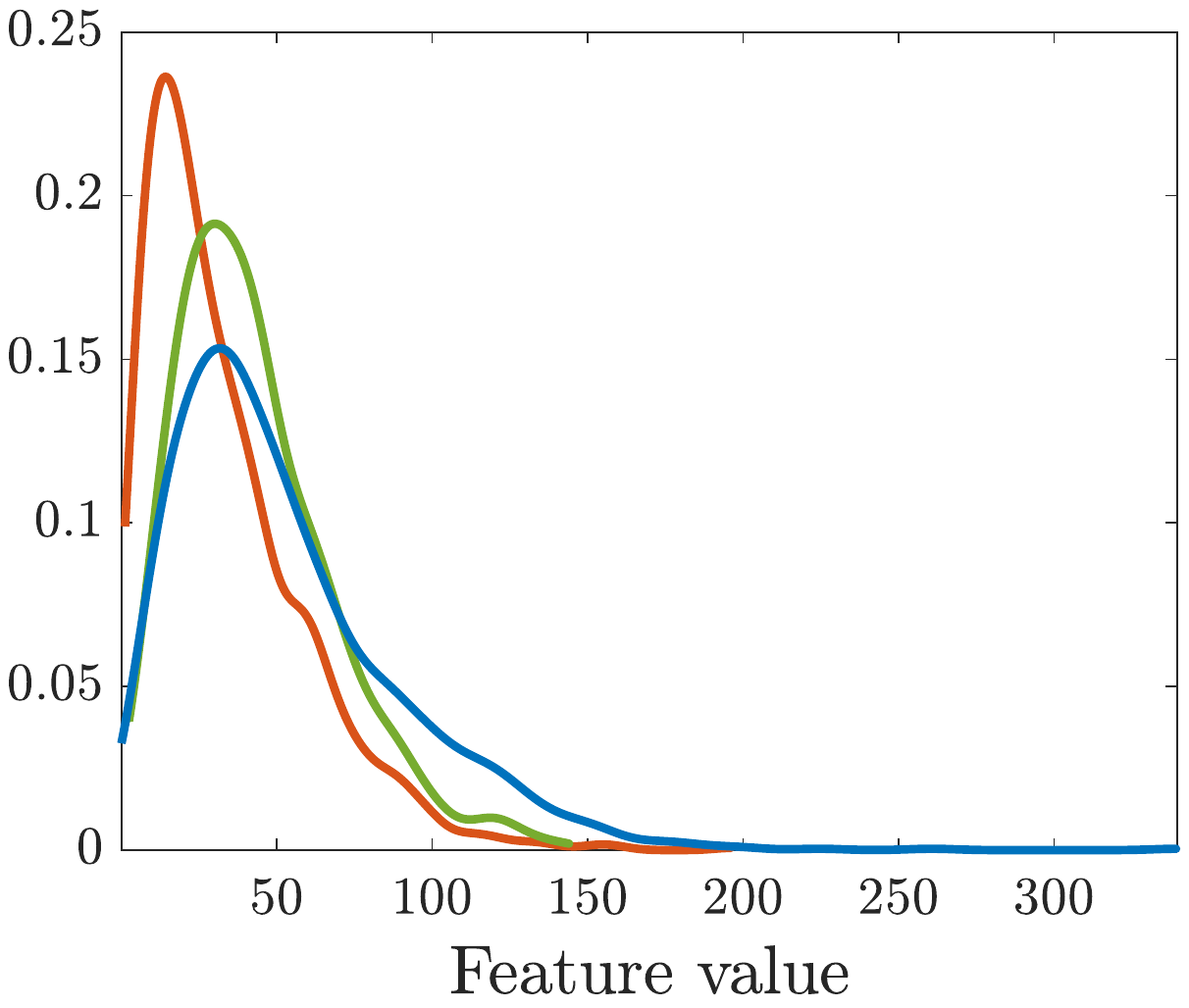} 
\label{fig:ind_feat_disd}}
\hspace{\hswidth}
\subfloat[SI][{SI}]{\includegraphics[height=\xwidth\textwidth]{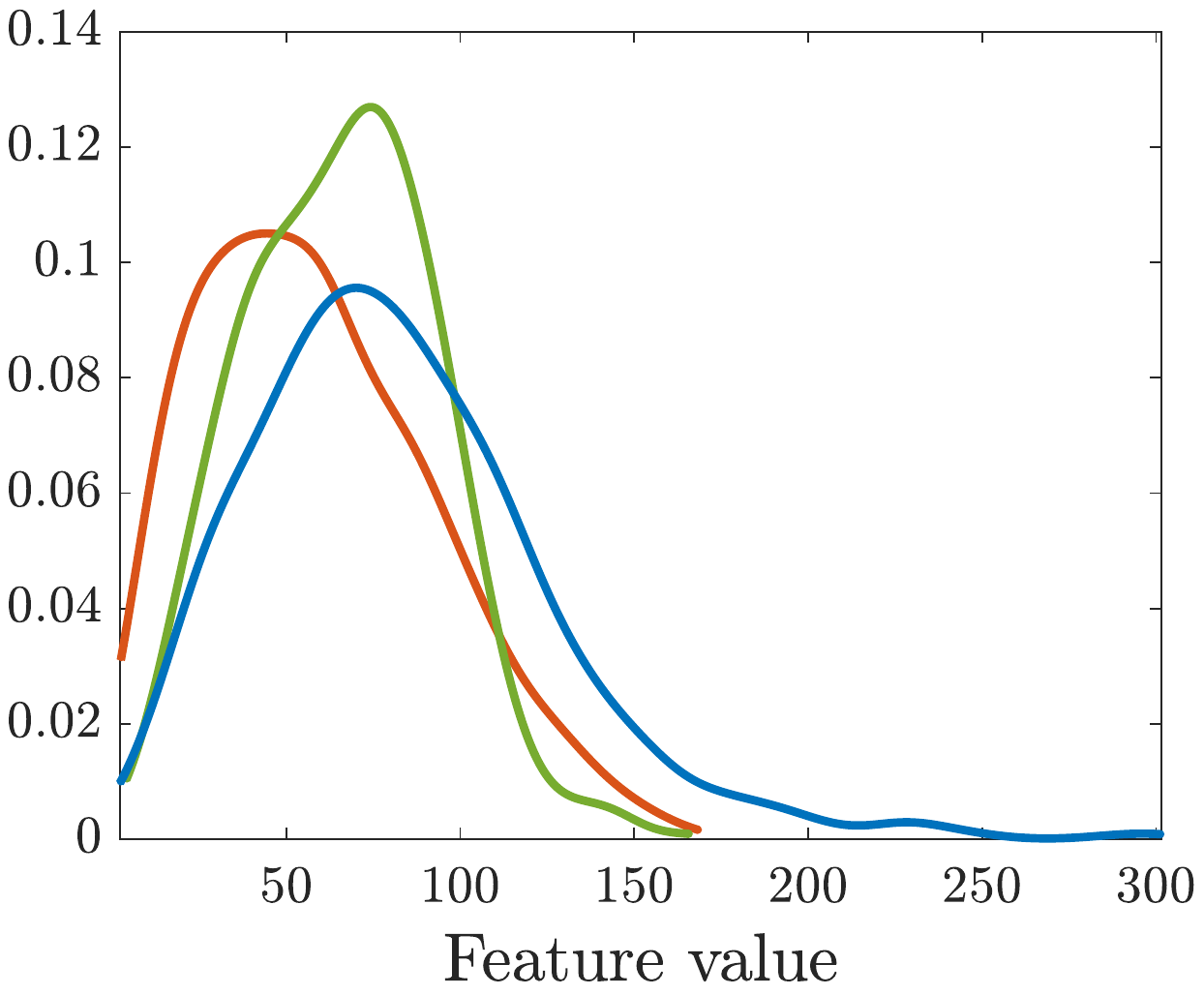} 
\label{fig:ind_feat_dise}}
\hspace{\hswidth}
\subfloat[TI][{TI}]{\includegraphics[height=\xwidth\textwidth]{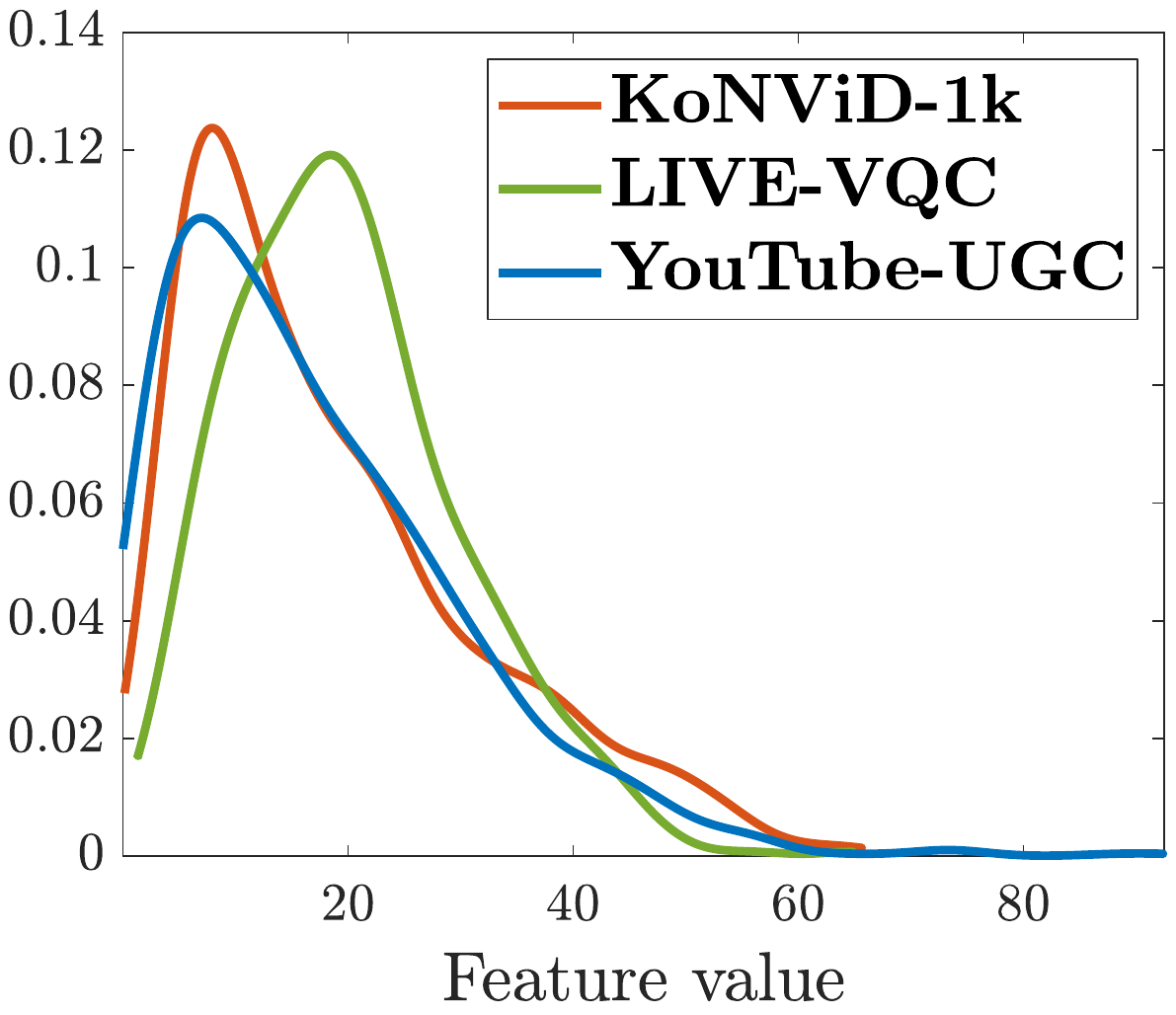} 
\label{fig:ind_feat_disf}}
\caption{Feature distribution comparisons among the three considered UGC-VQA databases: KoNViD-1k, LIVE-VQC, and YouTube-UGC.}
\label{fig:ind_feat_dis}
\end{figure*}

\begin{figure}[!t]
\centering
\def\xwidth{0.12}
\def\hswidth{-0.em}
\def\xlinewidth{0.255}
\def\xem{1pt}
\def\yem{3pt}
\footnotesize
\setlength{\tabcolsep}{1.5pt}
\renewcommand{\arraystretch}{1.0}
\begin{tabular}{ccc}

  \includegraphics[ height=\xlinewidth\linewidth, keepaspectratio]{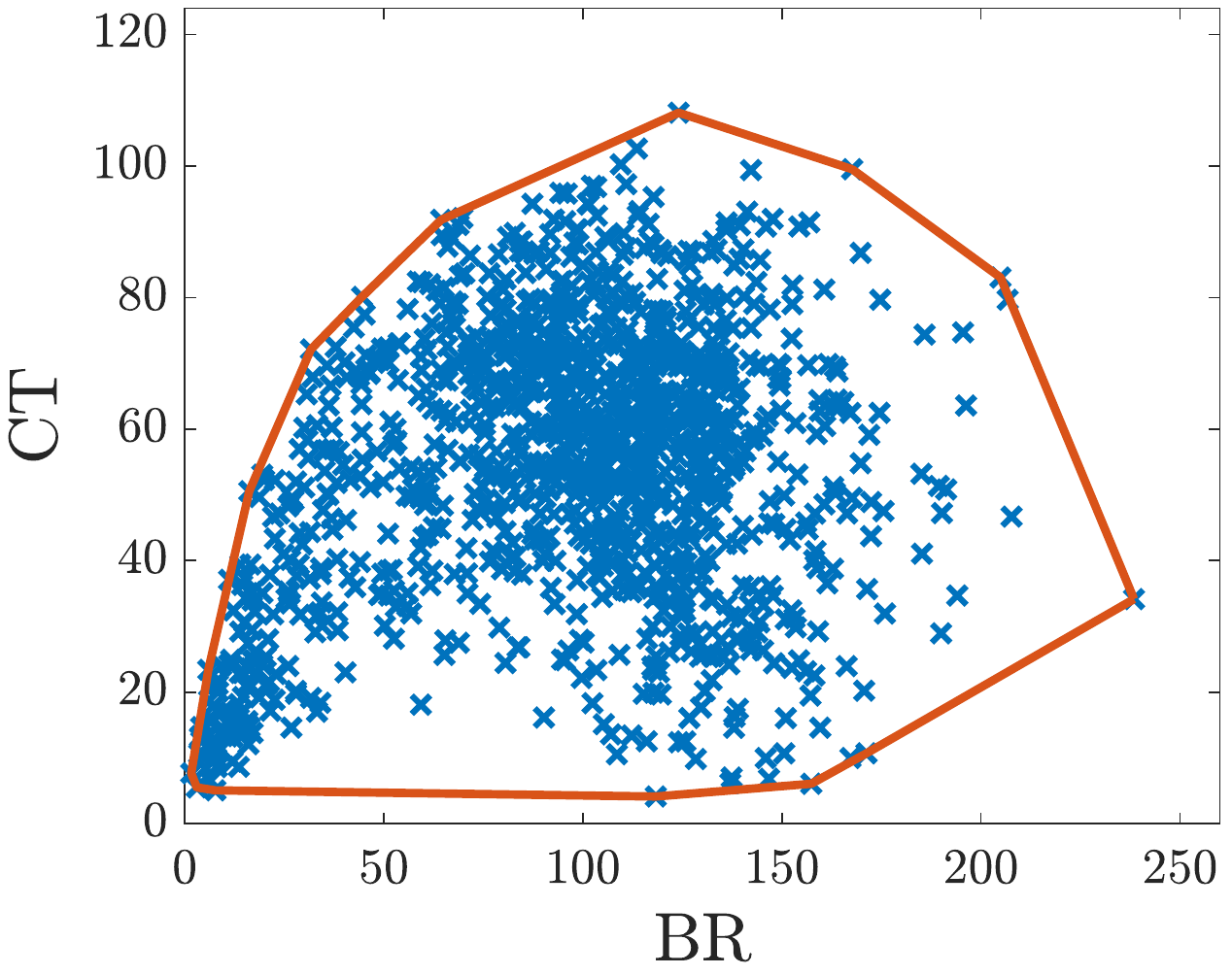} &
  \includegraphics[ height=\xlinewidth\linewidth, keepaspectratio]{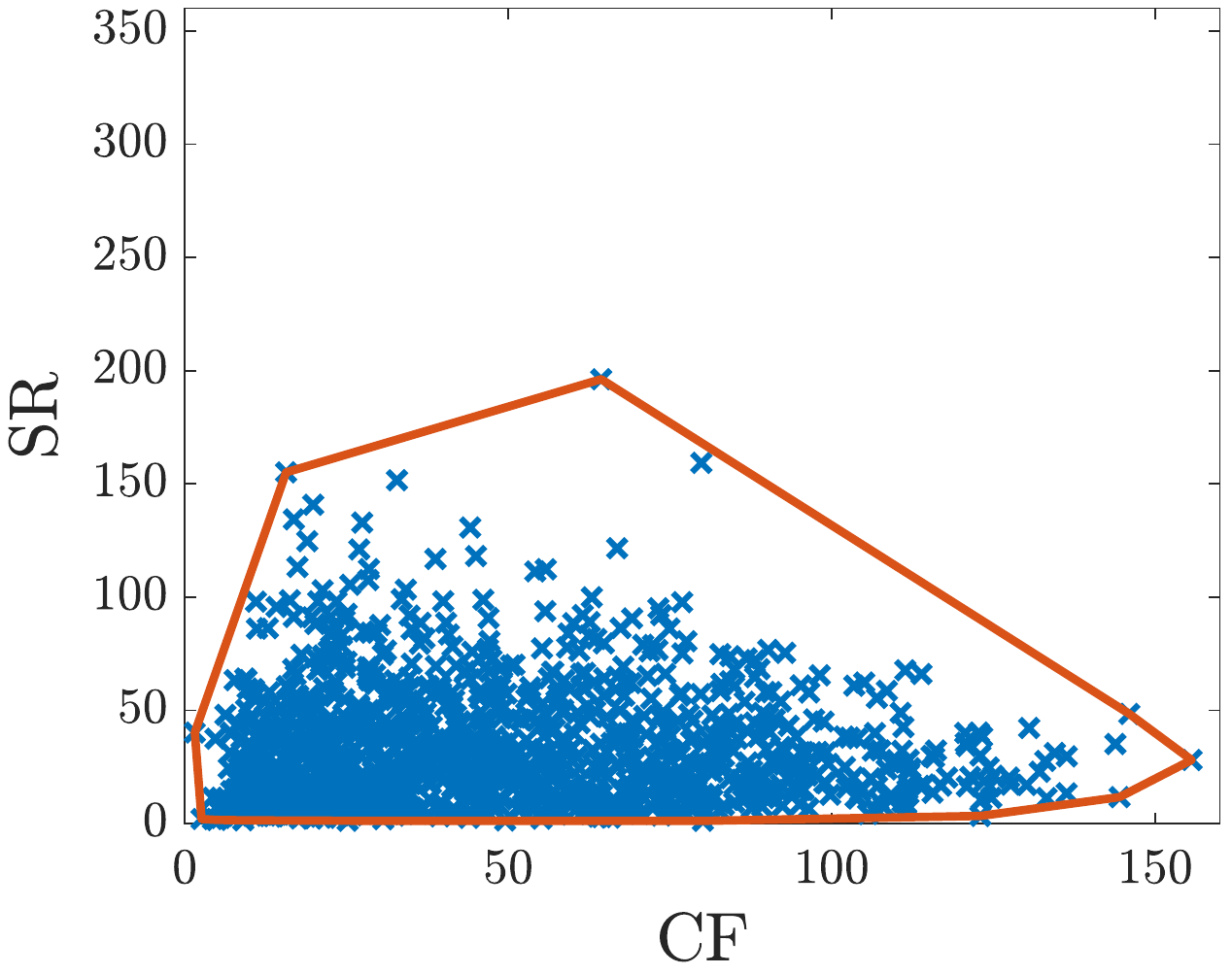} &
  \includegraphics[ height=\xlinewidth\linewidth, keepaspectratio]{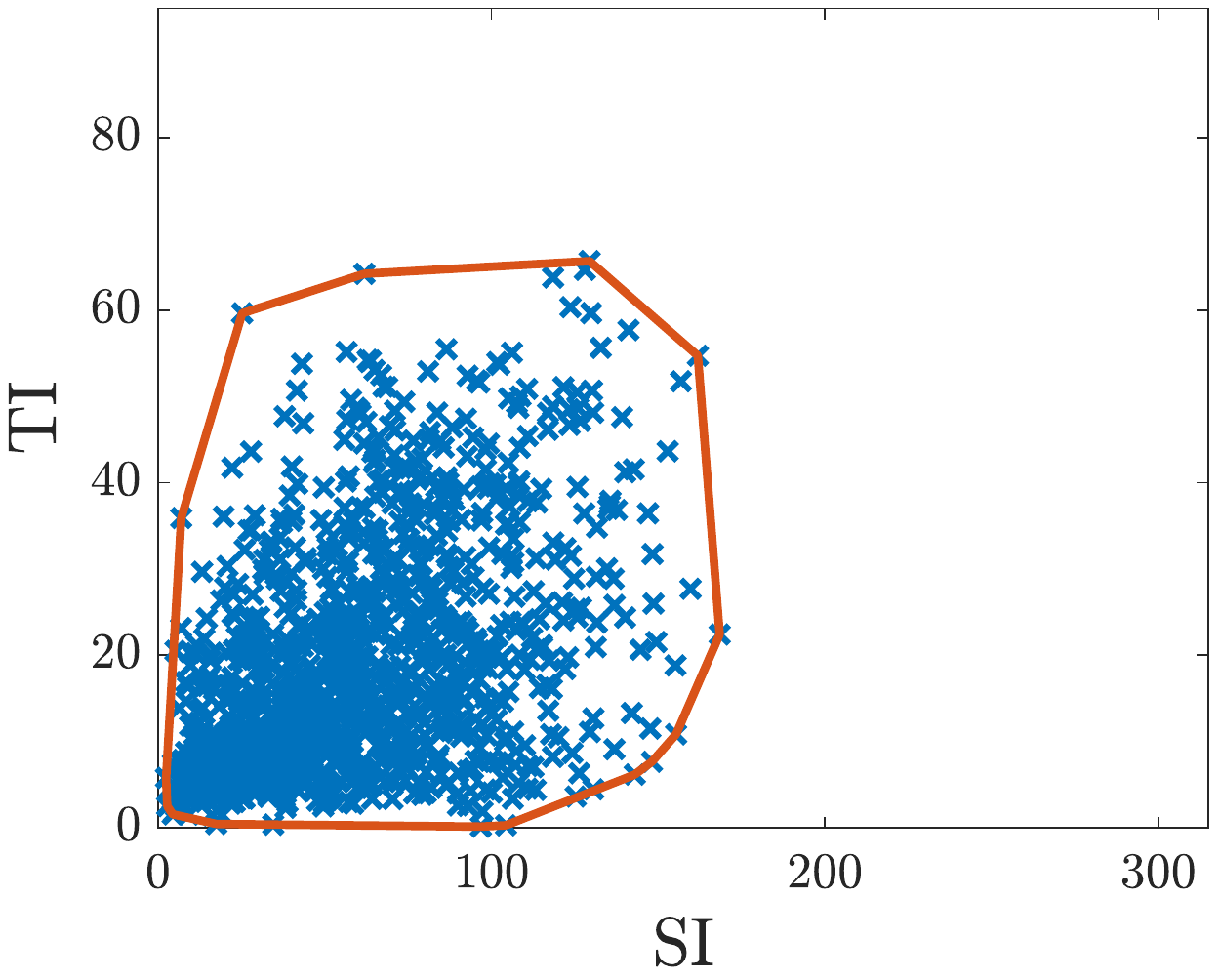} \\[\xem]
 \multicolumn{3}{c}{(a) {KoNViD-1k} } \\[\yem]
  \includegraphics[ height=\xlinewidth\linewidth, keepaspectratio]{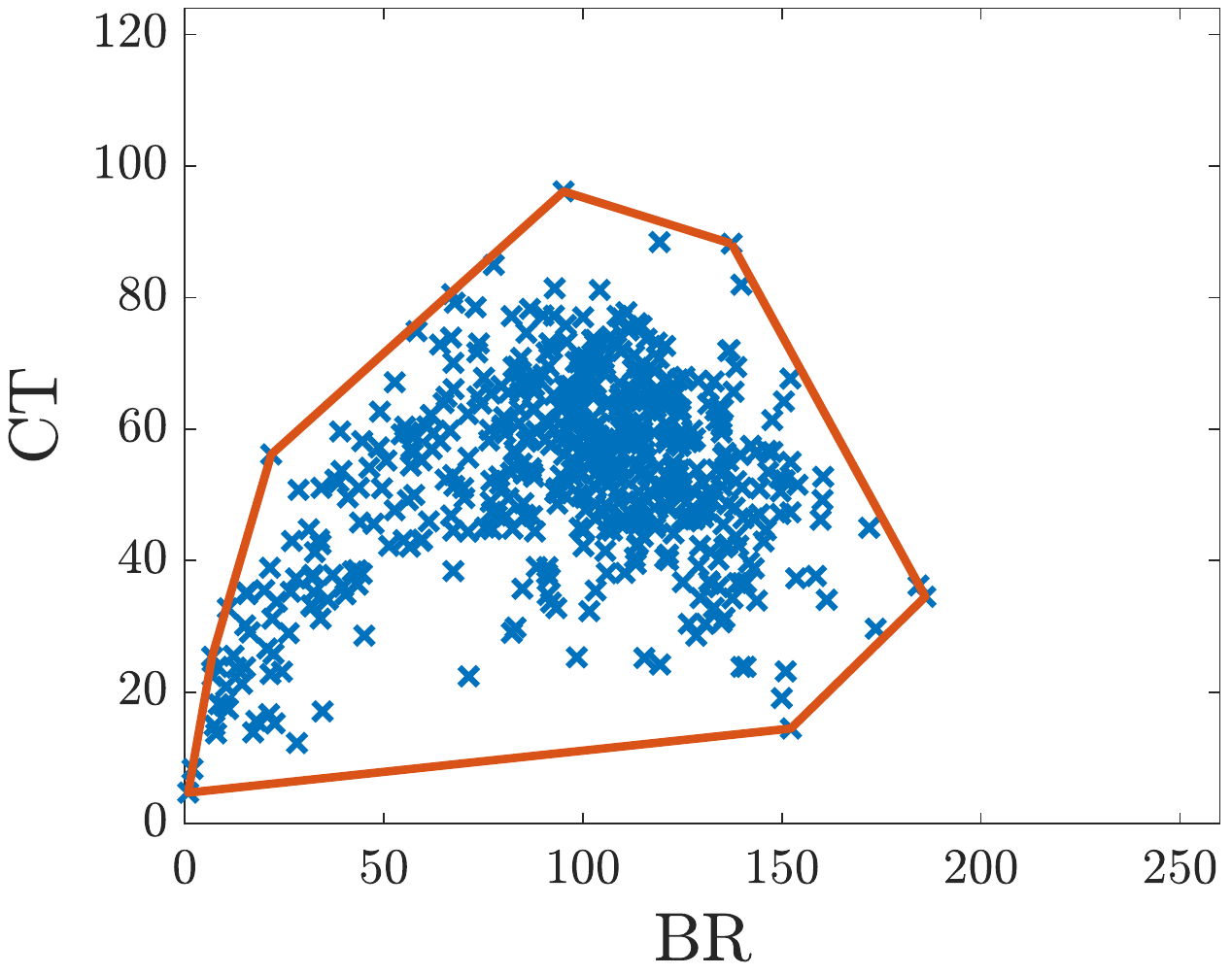} &
  \includegraphics[ height=\xlinewidth\linewidth, keepaspectratio]{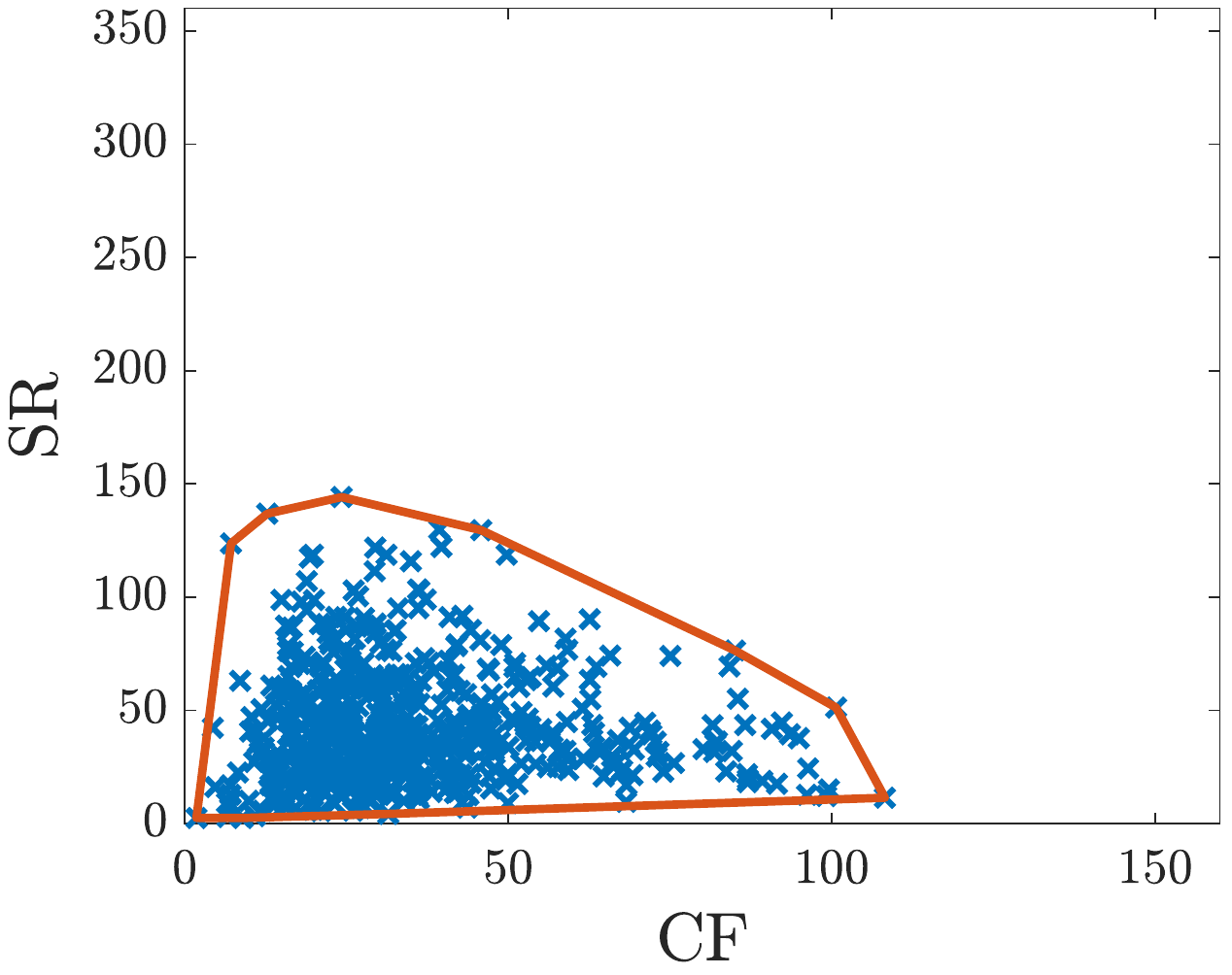} &
  \includegraphics[ height=\xlinewidth\linewidth, keepaspectratio]{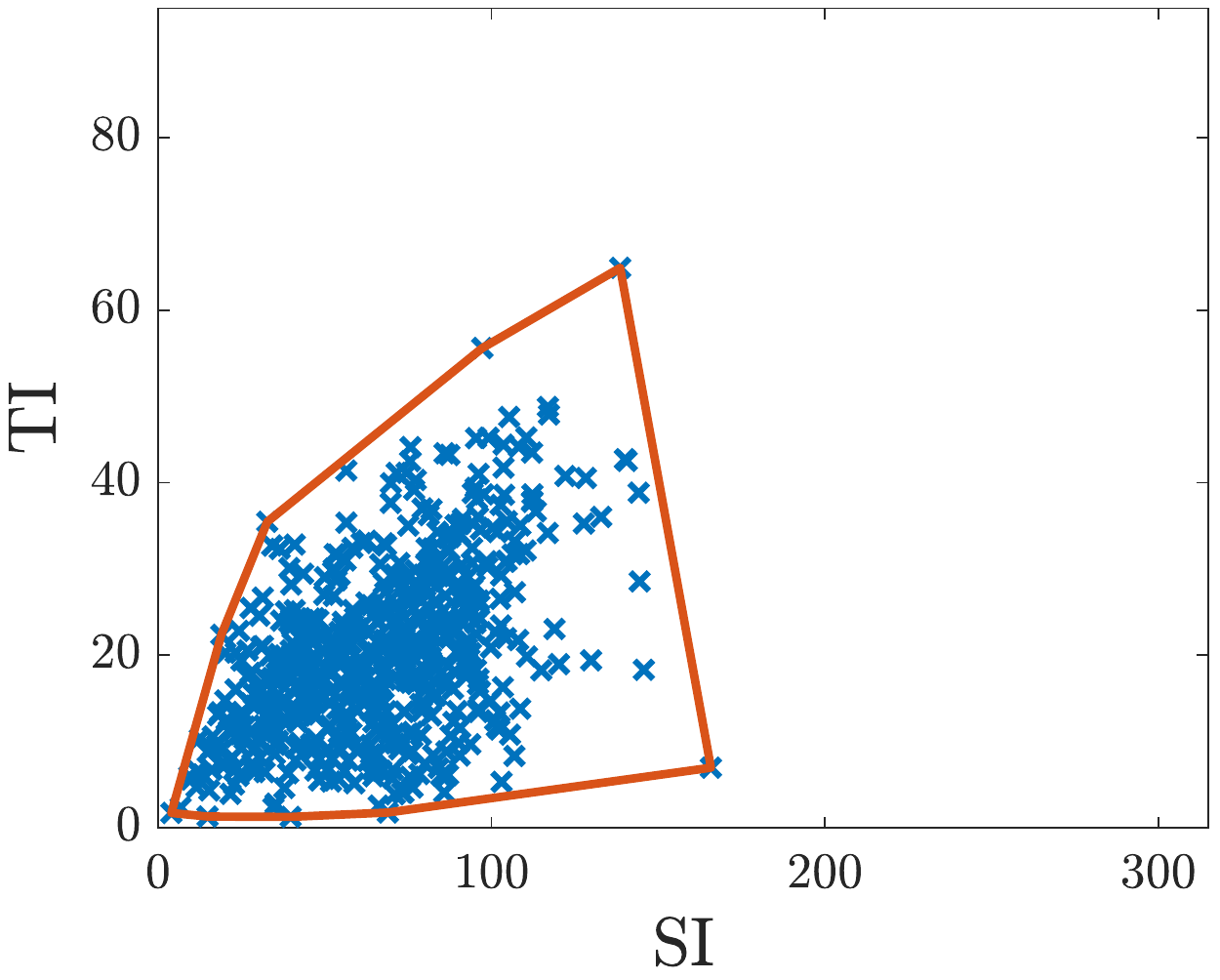}  \\[\xem]
  \multicolumn{3}{c}{(b) LIVE-VQC}    \\[\yem]
    \includegraphics[ height=\xlinewidth\linewidth, keepaspectratio]{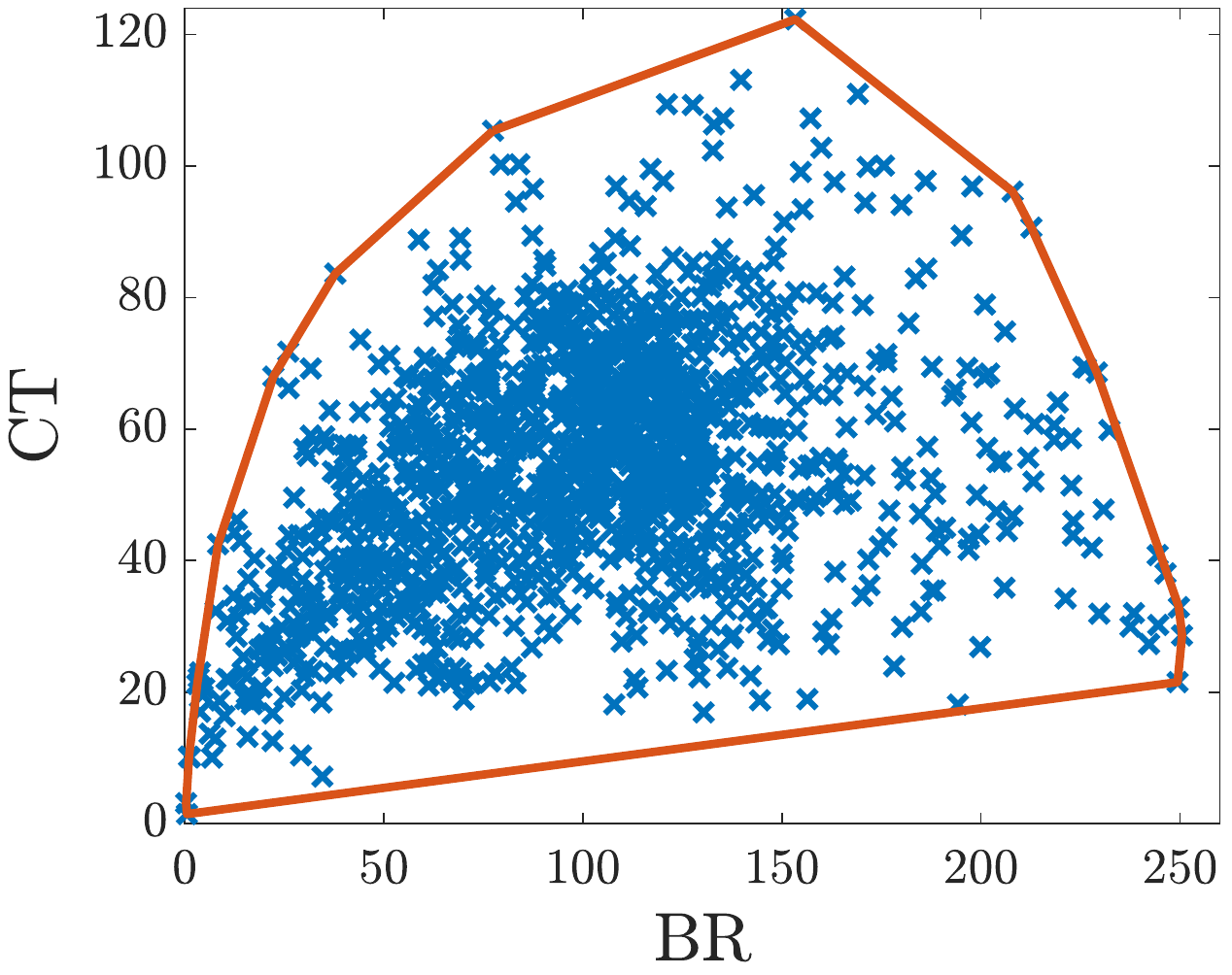} &
  \includegraphics[ height=\xlinewidth\linewidth, keepaspectratio]{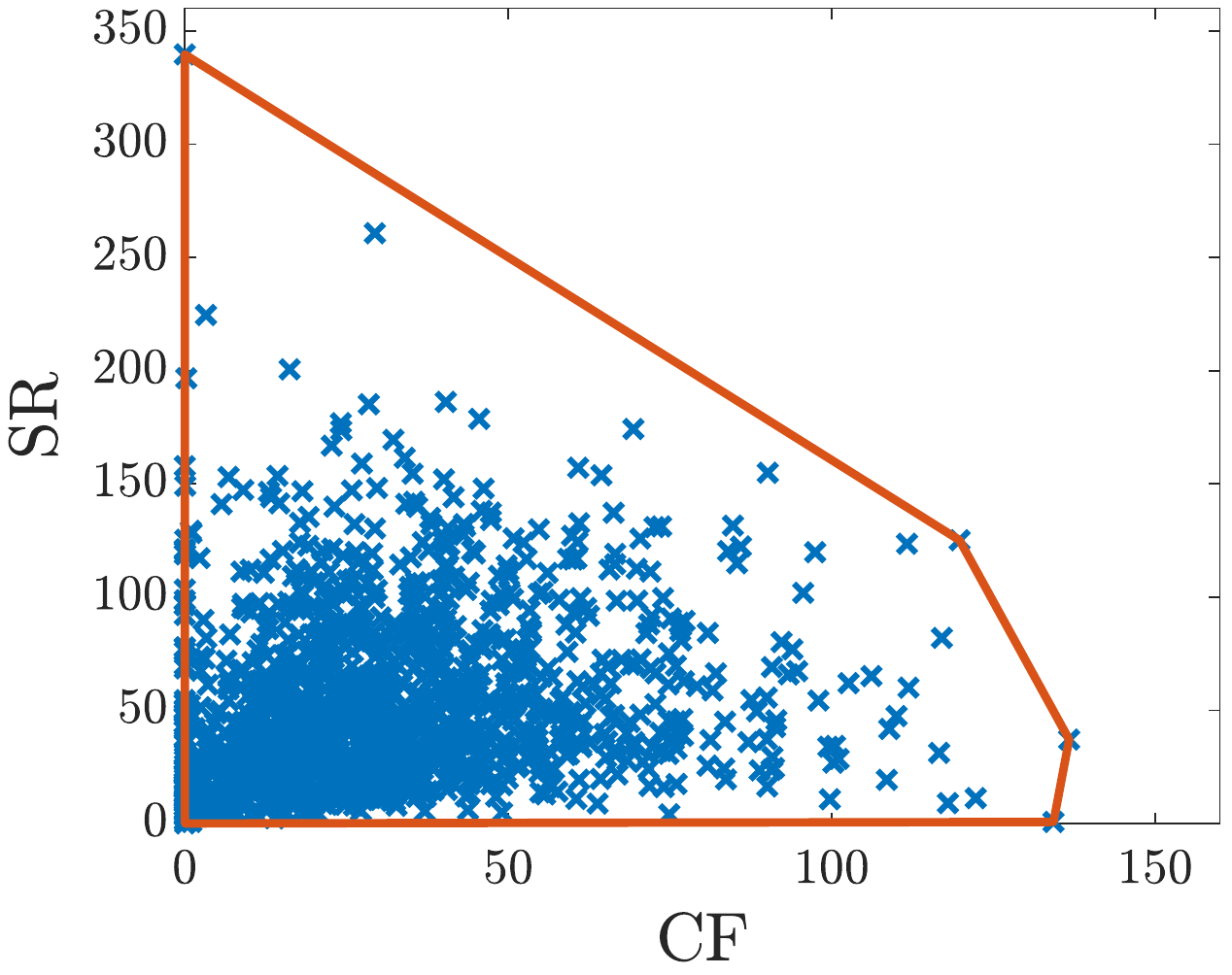} &
  \includegraphics[ height=\xlinewidth\linewidth, keepaspectratio]{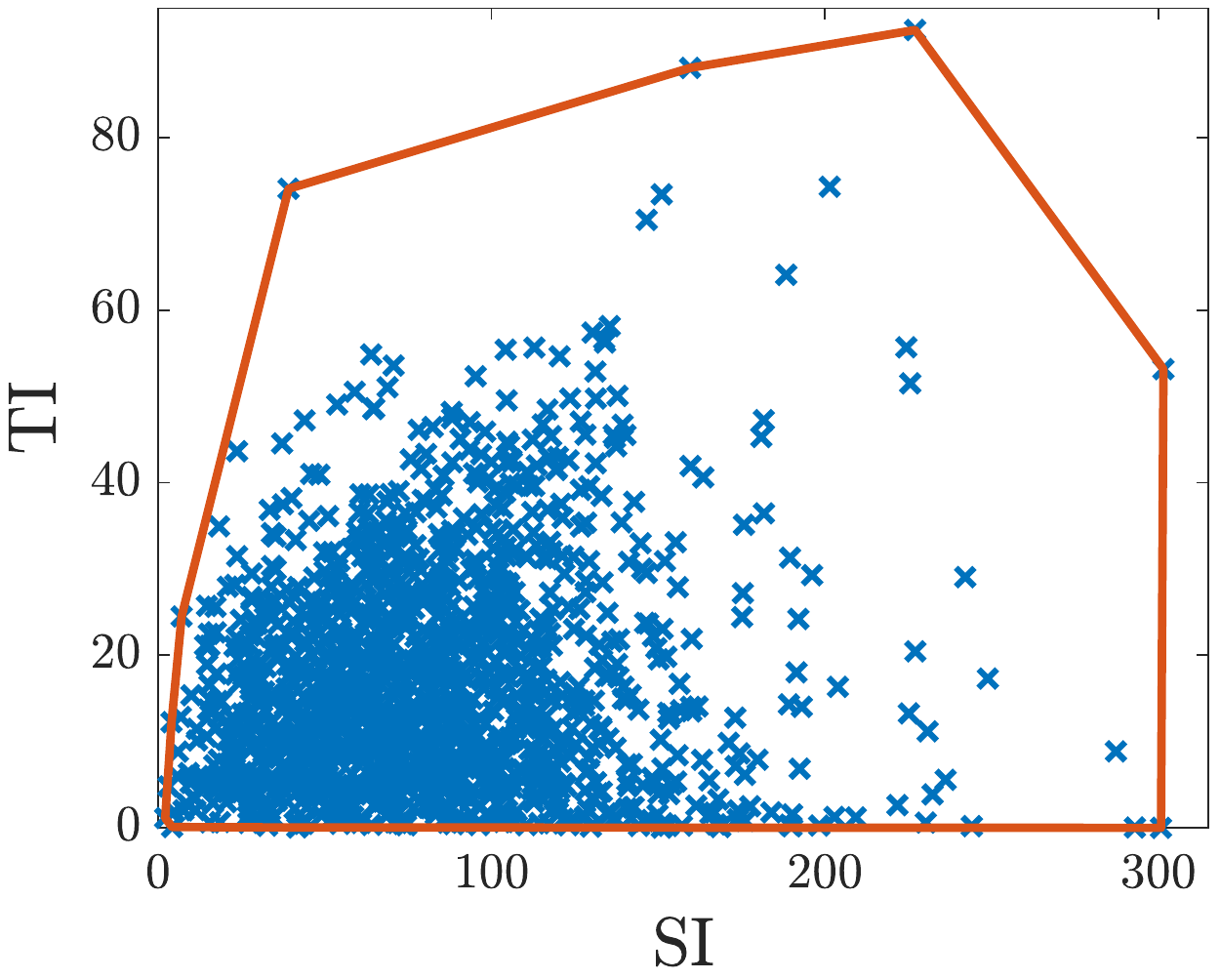}  \\[\xem]
     \multicolumn{3}{c}{(c) YouTube-UGC}    \\
\end{tabular}

\caption{Source content (blue `x') distribution in paired feature space with corresponding convex hulls (orange boundaries). Left column: BR$\times$CT, middle column: CF$\times$SR, right column: SI$\times$TI.}
\label{fig:feat_cvx_hull}
\end{figure}

As a way of characterizing the content diversity of the videos in each database, Winkler \cite{winkler2012analysis} suggested three quantitative attributes related to spatial activity, temporal activity, and colorfulness. Here we expand the set of attributes to include six low-level features including brightness, contrast, colorfulness \cite{hasler2003measuring}, sharpness, spatial information (SI), and temporal information (TI), thereby providing a larger visual space in which to plot and analyze content diversities of the three UGC-VQA databases. To reasonably limit the computational cost, each of these features was calculated on every 10th frame, then was averaged over frames to obtain an overall feature representation of each content. For simplicity, we denote the features as $\{\mathrm{C}_i\},i=1,2,...,6$. Figure \ref{fig:ind_feat_dis} shows the fitted kernel distribution of each selected feature. We also plotted the convex hulls of paired features, to show the feature coverage of each database, in Figure \ref{fig:feat_cvx_hull}. To quantify the coverage and uniformity of these databases over each defined feature space, we computed the relative range and uniformity of coverage \cite{winkler2012analysis}, where the relative range is given by:
\begin{equation}
\label{eq:relative_range}
\mathrm{R}_i^k=\frac{\max(\mathrm{C}_i^k)-\min(\mathrm{C}_i^k)}{\max_{k} (\mathrm{C}_i^k)},
\end{equation}
where $\mathrm{C}_i^k$ denotes the feature distribution of database $k$ for a given feature dimension $i$, and $\max_{k}(\mathrm{C}_i^k)$ specifies the maximum value for that given dimension across all databases. 

Uniformity of coverage measures how uniformly distributed the videos are in each feature dimension. We computed this as the entropy of the $\mathrm{B}$-bin histogram of $\mathrm{C}_i^k$ over all sources for each database indexed $k$:
\begin{equation}
\label{eq:uniformity}
\mathrm{U}^k_i=-\sum_{b=1}^\mathrm{B} p_b \log_\mathrm{B} p_b,
\end{equation}
where $p_b$ is the normalized number of sources in bin $b$ at feature $i$ for database $k$. The higher the uniformity the more uniform the database is. Relative range and uniformity of coverage are plotted in Figure \ref{fig:relative_range} and Figure \ref{fig:uniformity}, respectively, quantifying the intra- and inter-database differences in source content characteristics.

We also extracted 4,096-dimensional VGG-19 \cite{simonyan2014very} deep features and embedded these features into 2D subspace using t-SNE \cite{maaten2008visualizing} to further compare content diversity, as shown in Figure \ref{fig:tsne}. Apart from content diversity expressed in terms of visual features, the statistics of the subjective ratings are another important attribute of each video quality database. The main aspect considered in the analysis here is the distributions of mean opinion scores (MOS), as these are indicative of the quality range of the subjective judgements. The analysis of standard deviation of MOS is not presented here since it is not provided in LIVE-VQC. Figure \ref{fig:mos_dist} displays the histogram of MOS distributions for the three UGC-VQA databases. 

\subsection{Observations}
\label{ssec:observation}

\begin{figure}[!t]
\def\xheight{0.823}
\centering
\includegraphics[width=\xheight\linewidth]{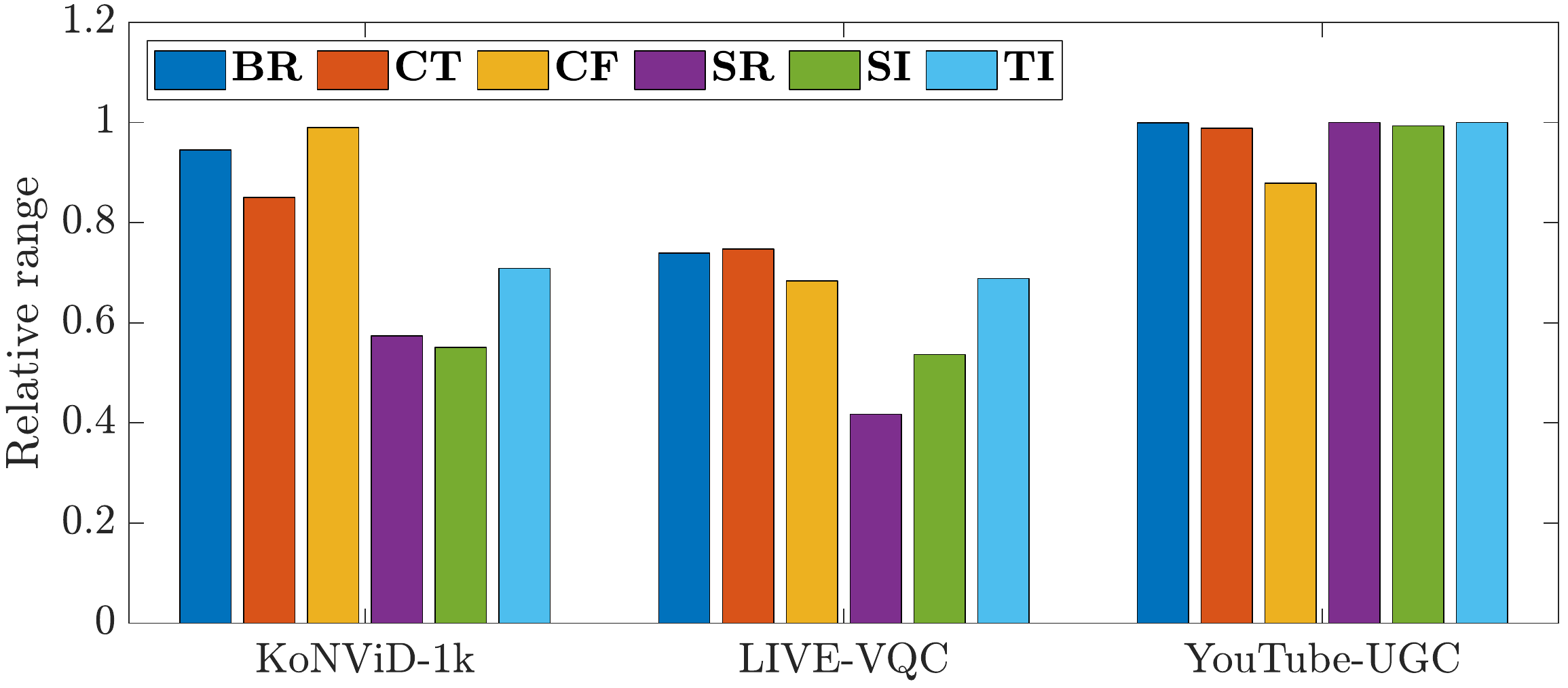}
\caption{Relative range $\mathrm{R}^k_i$ comparisons of the selected six features calculated on the three UGC-VQA databases: KoNViD-1k, LIVE-VQC, and YouTube-UGC.}
\label{fig:relative_range}
\end{figure}

\begin{figure}[!t]
\def\xheight{0.823}
\centering
\includegraphics[width=\xheight\linewidth]{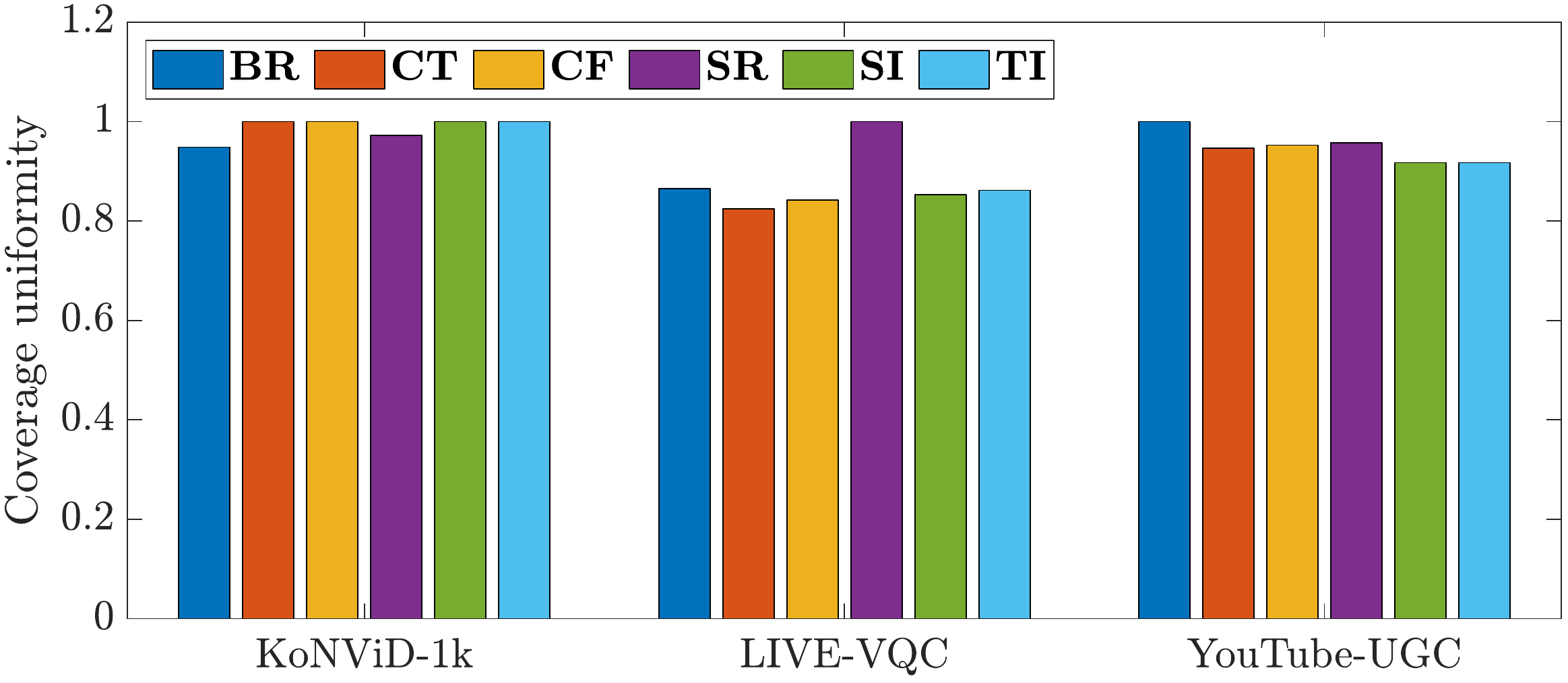}
\caption{Comparison of coverage uniformity $\mathrm{U}^k_i$ of the selected six features computed on the three UGC-VQA databases: KoNViD-1k, LIVE-VQC, and YouTube-UGC.}
\label{fig:uniformity}
\end{figure}

\begin{figure*}[!t]
\centering
\def\xwidth{0.16}
\def\hswidth{10pt}
\subfloat[KoNViD-1k][{KoNViD-1k}]{\includegraphics[height=\xwidth\textwidth]{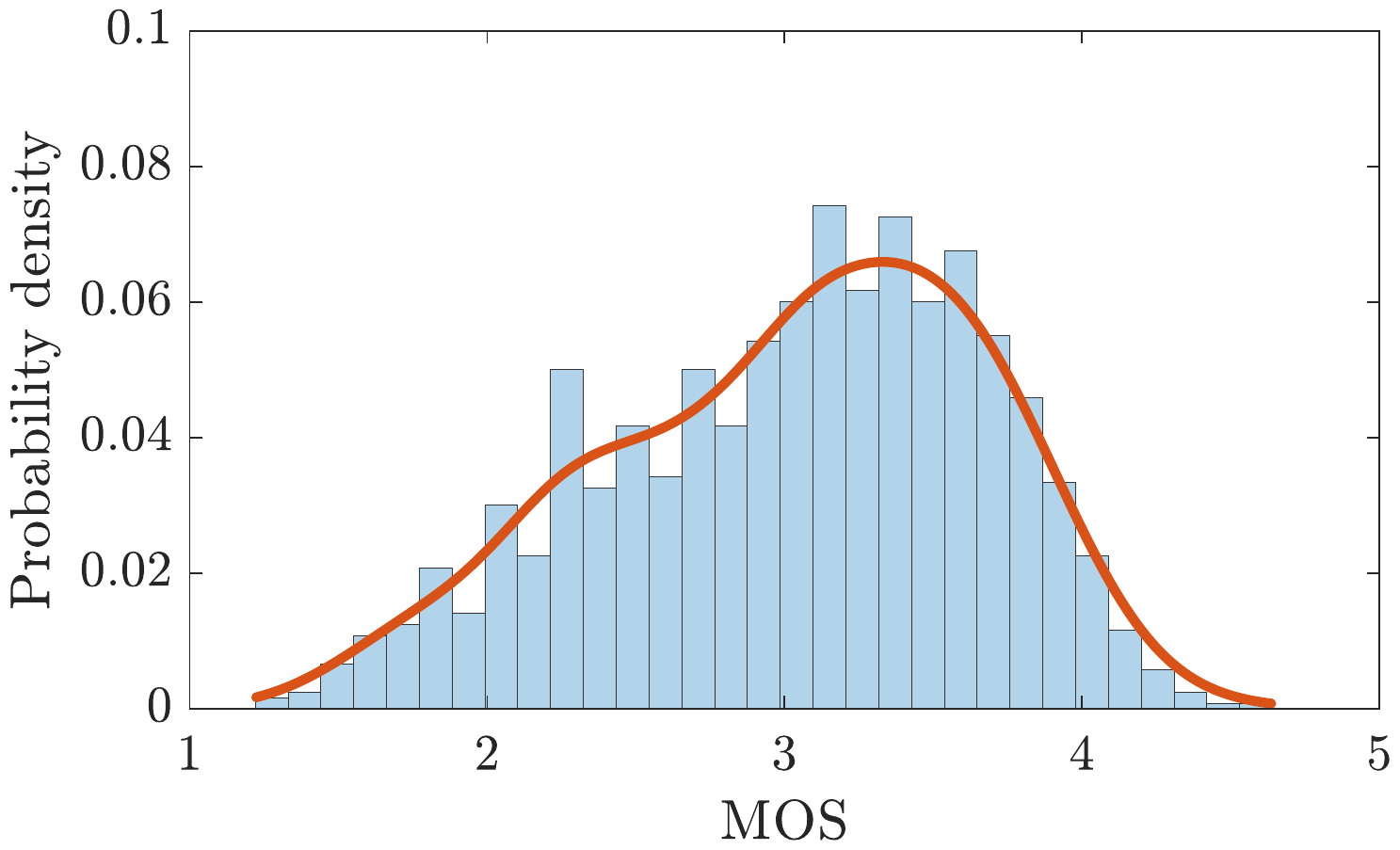} 
\label{fig:mos_dist-a}} 
\hspace{\hswidth}
\subfloat[LIVE-VQC][{LIVE-VQC}]{\includegraphics[height=\xwidth\textwidth]{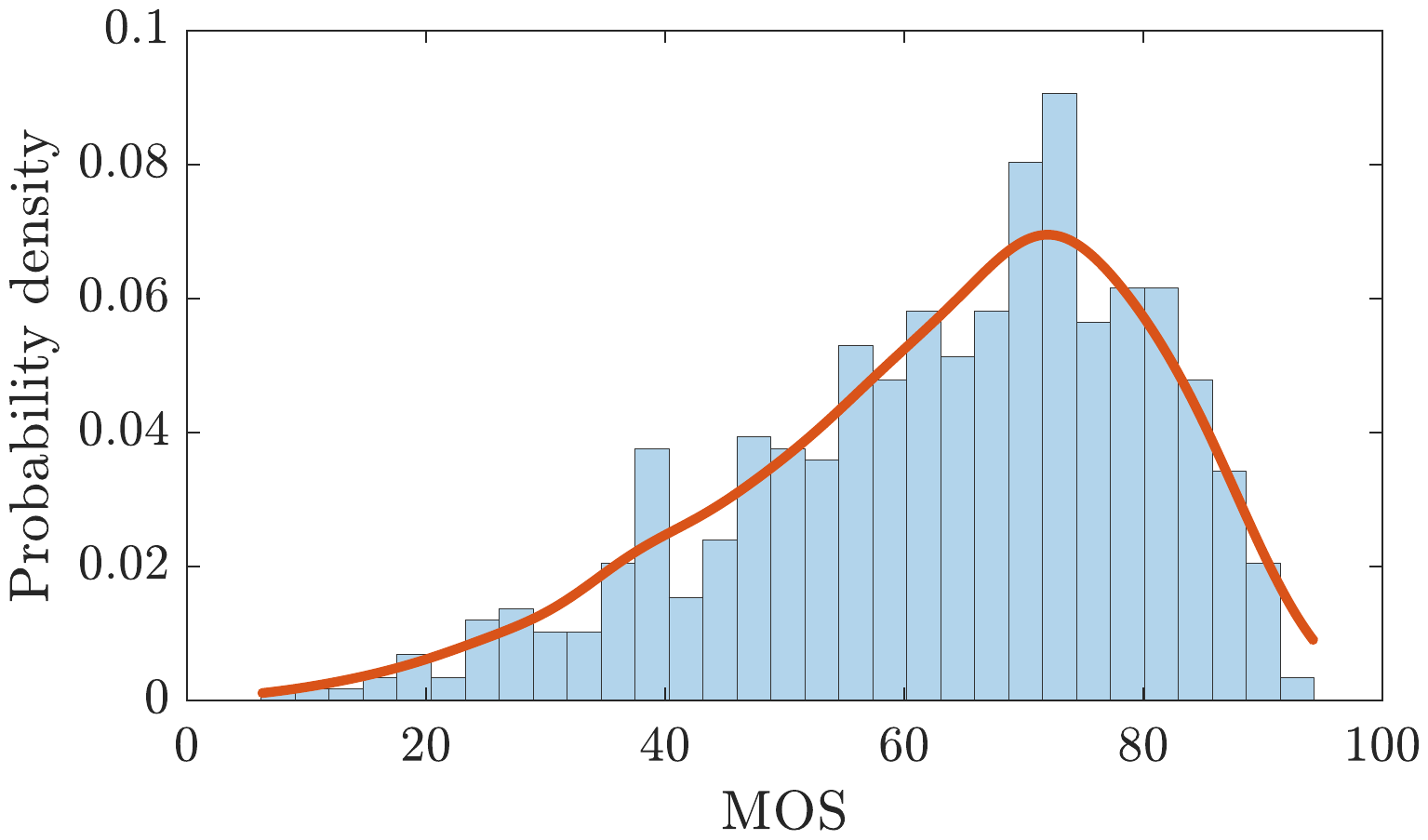} 
\label{fig:mos_dist-b}} 
\hspace{\hswidth}
\subfloat[YouTube-UGC][{{YouTube-UGC}}]{\includegraphics[height=\xwidth\textwidth]{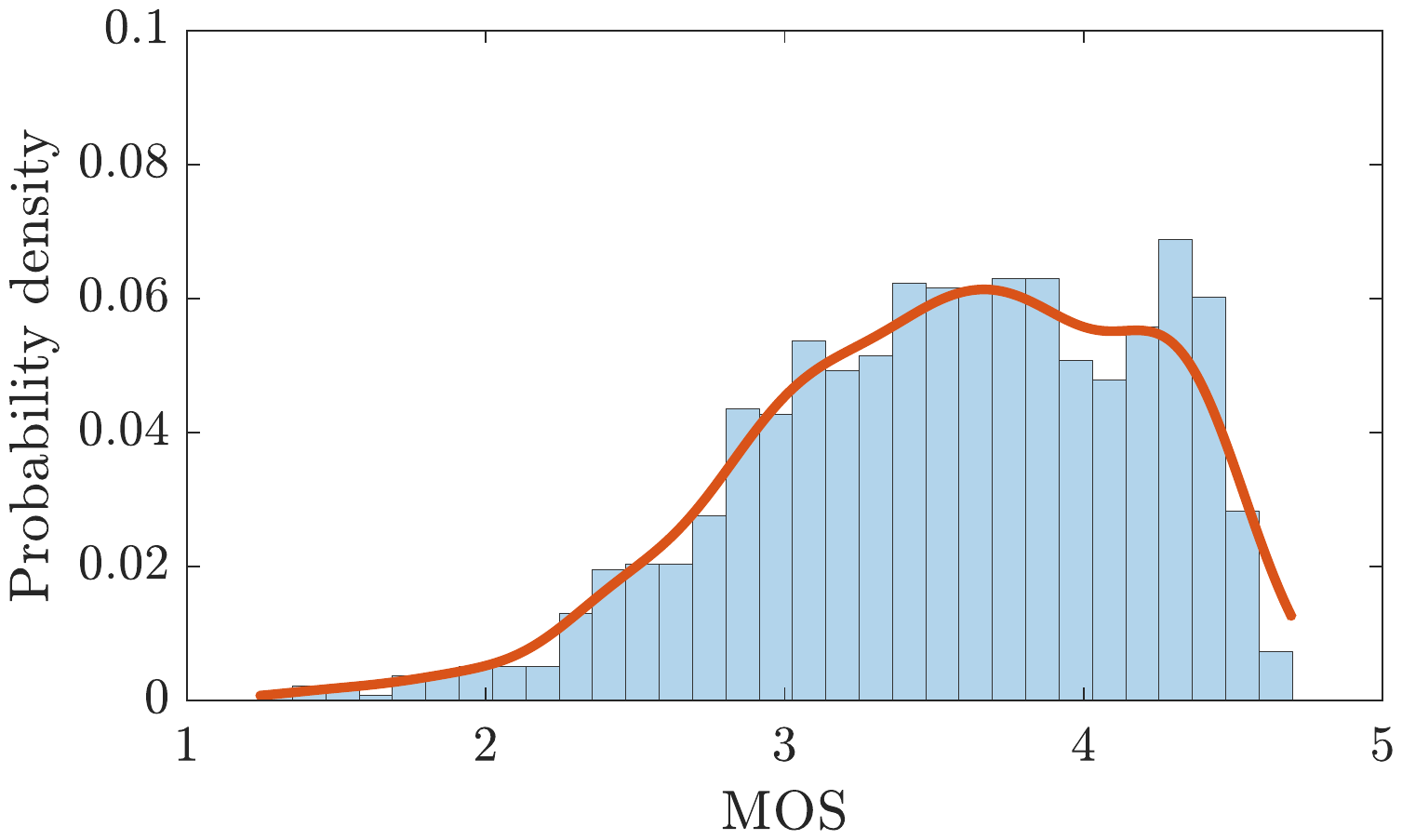}
\label{fig:mos_dist-c}}
\caption{MOS histograms and the fitted kernel distributions of the three UGC-VQA databases: KoNViD-1k, LIVE-VQC, and YouTube-UGC.}
\label{fig:mos_dist}
\end{figure*}

\begin{figure}[!t]
\def\xheight{0.6}
\centering
\includegraphics[width=\xheight\linewidth]{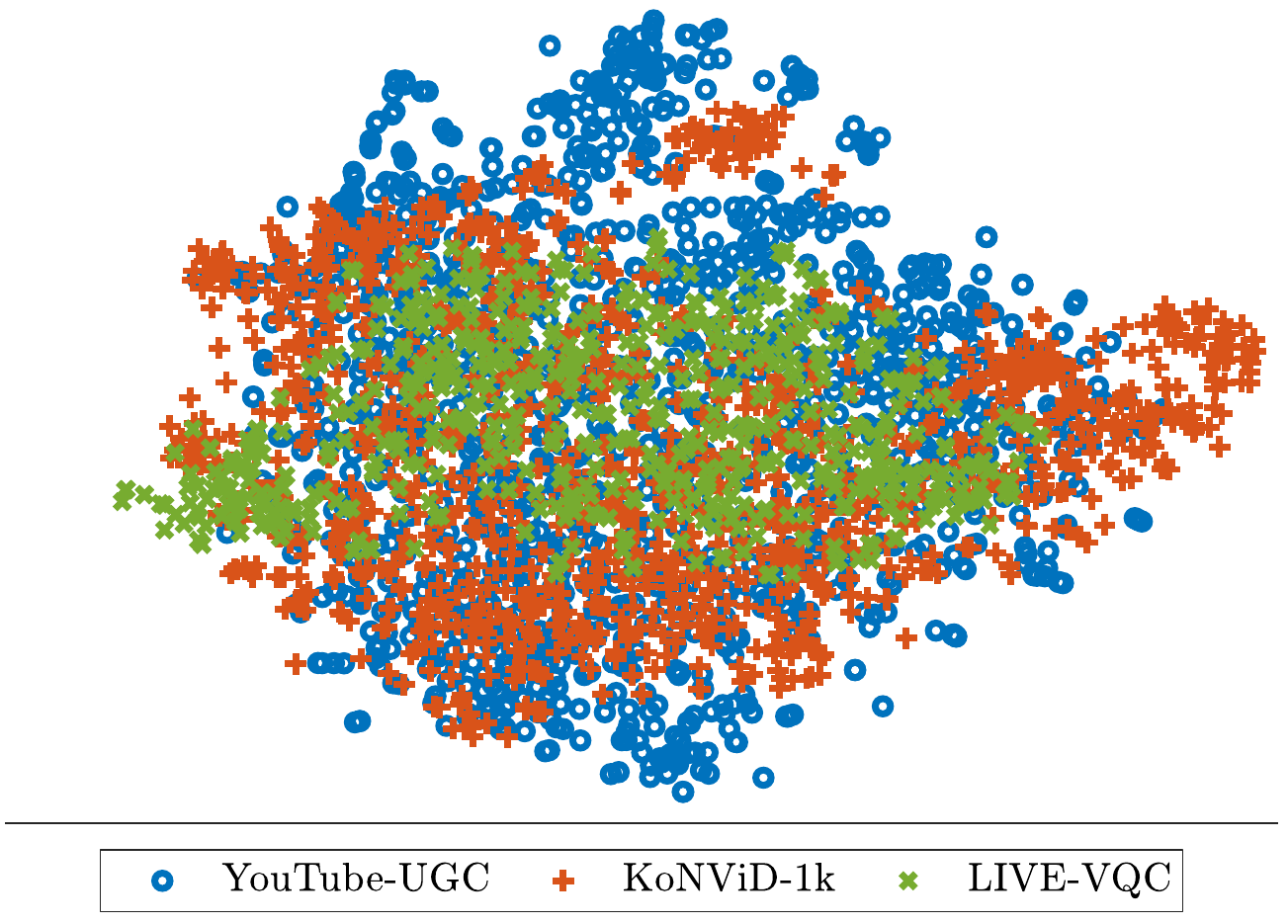}
\caption{VGG-19 deep feature embedding via t-SNE \cite{maaten2008visualizing} on KoNViD-1k, LIVE-VQC, and YouTube-UGC, respectively.}
\label{fig:tsne}
\end{figure}

We make some observations from the above plots. As may be seen in Figures \ref{fig:ind_feat_disa} and \ref{fig:ind_feat_disb}, and the corresponding convex hulls in Figure \ref{fig:feat_cvx_hull}, KoNViD-1k and YouTube-UGC exhibit similar coverage in terms of brightness and contrast, while LIVE-VQC adheres closer to middle values. Regarding colorfulness, KoNViD-1k shows a skew towards higher scores than the other two datasets, which is consistent with the observations that Flickr users self-characterize as either professional video/photographers or as dedicated amateurs. On the sharpness and SI histograms, YouTube-UGC is spread most widely, while KoNViD-1k is concentrated on lower values. Another interesting finding from the TI statistics: LIVE-VQC is distributed more towards higher values than YouTube-UGC and KoNViD-1k, consistent with our observation that videos in LIVE-VQC were captured in the presence of larger and more frequent camera motions. We will revisit this interesting aspect of TI when evaluating the BVQA models in Section \ref{sec:exp}. The visual comparison in Figure \ref{fig:tsne} shows that YouTube-UGC and KoNViD-1k span a wider range of VGG-19 feature space than does LIVE-VQC, indicating significant content diversity differences. Figure \ref{fig:mos_dist} shows the MOS distributions: all three databases have right-skewed MOS distributions, with KoNViD-1k less so, and LIVE-VQC and YouTube-UGC more so. The overall ranges and uniformity comparisons in Figures \ref{fig:relative_range}, \ref{fig:uniformity}, and \ref{fig:tsne} suggest that constructing a database by crawling and sampling from a large content repository is likely to yield a more content-diverse, uniformly-distributed dataset than one created from pictures or videos captured directly from a set of user cameras. Both cases may be argued to be realistic in some scenario.

\section{UGC-VQA Models}
\label{sec:nr_vqa}

The goal of subjective video quality studies is to motivate the development of automatic objective video quality models. Conventionally, objective video quality assessment can be classified into three main categories: full-reference (FR), reduced-reference (RR), and no-reference (NR) models. FR-VQA models require the availability of an entire pristine source video to measure visual differences between a target signal and a corresponding reference \cite{wang2004video, vmaf,sheikh2005information, chen2020perceptual}, while RR-VQA models only make use of a limited amount of reference information \cite{wang2005reduced,soundararajan2012video}. Some popular FR-VQA models, including PSNR, SSIM \cite{wang2004image}, and VMAF \cite{vmaf} have already been successfully and massively deployed to optimize streaming and shared/uploaded video encoding protocols by leading video service providers. NR-VQA or BVQA models, however, rely solely on analyzing the test stimuli without the benefit of any corresponding ``ground truth'' pristine signal. It is obvious that only BVQA models are appropriate for the UGC-VQA problem. Here we briefly review the evolution of BVQA models, from conventional handcrafted feature-based approaches, on to convolutional neural network-based models.

\subsection{Conventional Feature-Based BVQA Models}

Almost all of the earliest BVQA models have been `distortion specific,' meaning they were designed to quantify a specific type of distortion such as blockiness \cite{wang2000blind}, blur \cite{marziliano2002no}, ringing \cite{ feng2006measurement}, banding \cite{ wang2016perceptual, tu2020bband, tu2020adaptive}, or noise \cite{amer2005fast, norkin2018film} in distorted videos, or to assess multiple specific coincident distortion types caused by compression or transmission impairments \cite{caviedes2017no, keimel2009no}. More recent top-performing BVQA models are almost exclusively learning-based, leveraging a set of generic quality-aware features, combined to conduct quality prediction by machine learning regression \cite{ moorthy2011blind, mittal2012no, saad2014blind, kundu2017no, ghadiyaram2017perceptual, korhonen2019two, ye2012unsupervised, pei2015image, tu2021rapique}. Learning-based BVQA models are more versatile and generalizable than `distortion specific' models, in that the selected features are broadly perceptually relevant, while powerful regression models can adaptively map the features onto quality scores learned from the data in the context of a specific application.

The most popular BVQA algorithms deploy perceptually relevant, low-level features based on simple, yet highly regular parametric bandpass models of good-quality scene statistics \cite{ruderman1994statistics}. These natural scene statistics (NSS) models predictably deviate in the presence of distortions, thereby characterizing perceived quality degradations \cite{sheikh2006image}. Successful blind picture quality assessment (BIQA) models of this type have been developed in the wavelet (BIQI \cite{moorthy2010two}, DIIVINE \cite{moorthy2011blind}, C-DIIVINE \cite{zhang2014c}), discrete cosine transform (BLIINDS \cite{saad2010dct}, BLIINDS-II \cite{saad2012blind}), curvelet \cite{liu2014no}, and spatial intensity domains (NIQE \cite{mittal2012making}, BRISQUE \cite{mittal2012no}), and have further been extended to video signals using natural bandpass space-time video statistics models \cite{li2016spatiotemporal,mittal2015completely,saad2014blind, sinno2019spatio}, among which the most well-known model is the Video-BLIINDS \cite{saad2014blind}. Other extensions to empirical NSS include the joint statistics of the gradient magnitude and Laplacian of Gaussian responses in the spatial domain (GM-LOG \cite{xue2014blind}), in log-derivative and log-Gabor spaces (DESIQUE \cite{zhang2013no}), as well as in the gradient domain of LAB color transforms (HIGRADE \cite{kundu2017no}). The FRIQUEE model \cite{ghadiyaram2017perceptual} has been observed to achieve SOTA performance both on UGC/consumer video/picture databases like LIVE-Challenge \cite{ghadiyaram2015massive}, CVD2014 \cite{nuutinen2016cvd2014}, and KoNViD-1k \cite{hosu2017konstanz} by leveraging a bag of NSS features drawn from diverse color spaces and perceptually motivated transform domains.

Instead of using NSS-inspired feature descriptors, methods like CORNIA \cite{ye2012unsupervised} employ unsupervised learning techniques to learn a dictionary (or codebook) of distortions from raw image patches, and was further extended to Video CORNIA \cite{xu2014no} by applying an additional temporal hysteresis pooling \cite{seshadrinathan2011temporal} of learned frame-level quality scores. Similar to CORNIA, the authors of \cite{xu2016blind} proposed another codebook-based general-purpose BVQA method based on High Order Statistics Aggregation (HOSA), requiring only a small codebook, yet yielding promising performance.

A very recent handcrafted feature-based BVQA model is the ``two level'' video quality model (TLVQM) \cite{korhonen2019two}, wherein a two-level feature extraction mechanism is adopted to achieve efficient computation of a set of carefully-defined impairment/distortion-relevant features. Unlike NSS features, TLVQM selects a comprehensive feature set comprising of empirical motion statistics, specific artifacts, and aesthetics. TLVQM does require that a large set of parameters (around 30) be specified, which may affect performance on datasets or application scenarios it has not been exposed to. The model currently achieves SOTA performance on three UGC video quality databases, CVD2014 \cite{nuutinen2016cvd2014}, KoNViD-1k \cite{hosu2017konstanz}, and LIVE-Qualcomm \cite{ghadiyaram2017capture}, at a reasonably low complexity, as reported by the authors.

\subsection{Deep Convolutional Neural Network-Based BVQA Models}

Deep convolutional neural networks (CNNs or ConvNets) have been shown to deliver standout performance on a wide variety of low-level computer vision applications. Recently, the release of several ``large-scale'' (in the context of IQA/VQA research) subjective quality databases \cite{ghadiyaram2015massive, hosu2017konstanz} have sped the application of deep CNNs to perceptual quality modeling. For example, several deep learning picture-quality prediction methods were proposed in \cite{kang2014convolutional, kim2017deep, ying2019patches, chen2020proxiqa}. To conquer the limits of data scale, they either propose to conduct patch-wise training \cite{kang2014convolutional, bosse2016deep, kim2017deep} using global scores, or by pretraining deep nets on ImageNet \cite{deng2009imagenet}, then fine tuning. Several authors report SOTA performance on legacy synthetic distortion databases \cite{sheikh2006statistical,ponomarenko2013color} or on naturally distorted databases \cite{ghadiyaram2015massive, hosu2020koniq}.

Among the applications of deep CNNs to blind video quality prediction, Kim \cite{kim2018deep} proposed a deep video quality assessor (DeepVQA) to learn the spatio-temporal visual sensitivity maps via a deep ConvNet and a convolutional aggregation network. The V-MEON model \cite{liu2018end} used a multi-task CNN framework which jointly optimizes a 3D-CNN for feature extraction and a codec classifier using fully-connected layers to predict video quality. Zhang \cite{zhang2018blind} leveraged transfer learning to develop a general-purpose BVQA framework based on weakly supervised learning and a resampling strategy. In the VSFA model \cite{li2019quality}, the authors applied a pre-trained image classification CNN as a deep feature extractor and integrated the frame-wise deep features using a gated recurrent unit and a subjectively-inspired temporal pooling layer, and reported leading performance on several natural video databases \cite{nuutinen2016cvd2014, hosu2017konstanz, ghadiyaram2017capture}. These SOTA deep CNN-based BVQA models \cite{kim2018deep, liu2018end, zhang2018blind, li2019quality} produce accurate quality predictions on legacy (single synthetic distortion) video datasets \cite{seshadrinathan2010study, vu2014vis3}, but struggle on recent in-the-wild UGC databases \cite{nuutinen2016cvd2014, ghadiyaram2017capture,hosu2017konstanz}.

\begin{table}[!t]
\setlength{\tabcolsep}{4pt}
\renewcommand{\arraystretch}{1.1}
\centering
\caption{Summary of the initial feature set and the finalized VIDEVAL subset after feature selection.}
\label{table:feat_sum}
\begin{threeparttable}
\begin{tabular}{llcc}
\toprule
\textsc{Feature Name} &  \textsc{Feature Index} & \textsc{\#($\mathcal{F}_\mathrm{INIT}$)}  & \textsc{\#($\mathrm{VIDEVAL}$)} \\ \hline\\[-1.em]
BRISQUE\textsubscript{avg}  & $f_1-f_{36}$  & 36 & 3   \\
BRISQUE\textsubscript{std}  &  $f_{37}-f_{72}$  &  36  & 1 \\
GM-LOG\textsubscript{avg} &  $f_{73}-f_{112}$  & 40  &  4  \\
GM-LOG\textsubscript{std} &  $f_{113}-f_{152}$    &  40  & 5 \\
HIGRADE-GRAD\textsubscript{avg} &  $f_{153}-f_{188}$    &  36  & 8 \\
HIGRADE-GRAD\textsubscript{std} &  $f_{189}-f_{224}$    &   36 &  1 \\
FRIQUEE-LUMA\textsubscript{avg}  &    $f_{225}-f_{298}$    &  74  & 4  \\
FRIQUEE-LUMA\textsubscript{std}  &    $f_{299}-f_{372}$    & 74   & 8  \\
FRIQUEE-CHROMA\textsubscript{avg}  &    $f_{373}-f_{452}$    &  80  & 10 \\
FRIQUEE-CHROMA\textsubscript{std}  &    $f_{453}-f_{532}$    &  80  &  1 \\
FRIQUEE-LMS\textsubscript{avg}  &    $f_{533}-f_{606}$    &  74  & 1 \\
FRIQUEE-LMS\textsubscript{std}  &    $f_{607}-f_{680}$    & 74   & 0  \\
FRIQUEE-HS\textsubscript{avg}  &    $f_{681}-f_{684}$    &  4 & 0  \\
FRIQUEE-HS\textsubscript{std}  &    $f_{685}-f_{688}$    & 4   & 0  \\
TLVQM-LCF\textsubscript{avg}  &    $f_{689}-f_{710}$    & 22   &  5 \\
TLVQM-LCF\textsubscript{std}  &    $f_{711}-f_{733}$    & 23   &  3 \\
TLVQM-HCF  &    $f_{734}-f_{763}$    &  30 & 6 \\ \hline\\[-1.em]
$\mathcal{F}_\mathrm{ALL}$     &   $f_1-f_{763}$  &  763  &  60 \\
\bottomrule
\end{tabular}
\begin{tablenotes}[para,flushleft]
\footnotesize
\item $^\star$All the spatial features are calculated every two frames and aggregated into a single feature vector within 1-sec chunks. The overall feature vector for the whole video is then obtained by averaging all the chunk-wise feature vectors. Subscript \textit{avg} means within-chunk average pooling, whereas subscript \textit{std} means within-chunk standard deviation pooling.
\end{tablenotes}
\end{threeparttable}
\end{table}

\section{Feature Fused VIDeo Quality EVALuator (VIDEVAL)}
\label{sec:fs}

We have just presented a diverse set of BVQA models designed from a variety of perspectives, each either based on scene statistics, or motivated by visual impairment heuristics. As might be expected, and as we shall show later, the performances of these models differ, and also vary on different datasets. We assume that the features extracted from different models may represent statistics of the signal in different perceptual domains, and henceforce, a selected fusion of BVQA models may be expected to deliver better consistency against subjective assessment, and also to achieve more reliable performance across different databases and use cases. This inspired our new feature fused VIDeo quality EVALuator (VIDEVAL), as described next. 

\begin{figure}[!t]
\def\xwidth{0.8}
\def\y{1}
\centering
\includegraphics[width=\xwidth\linewidth]{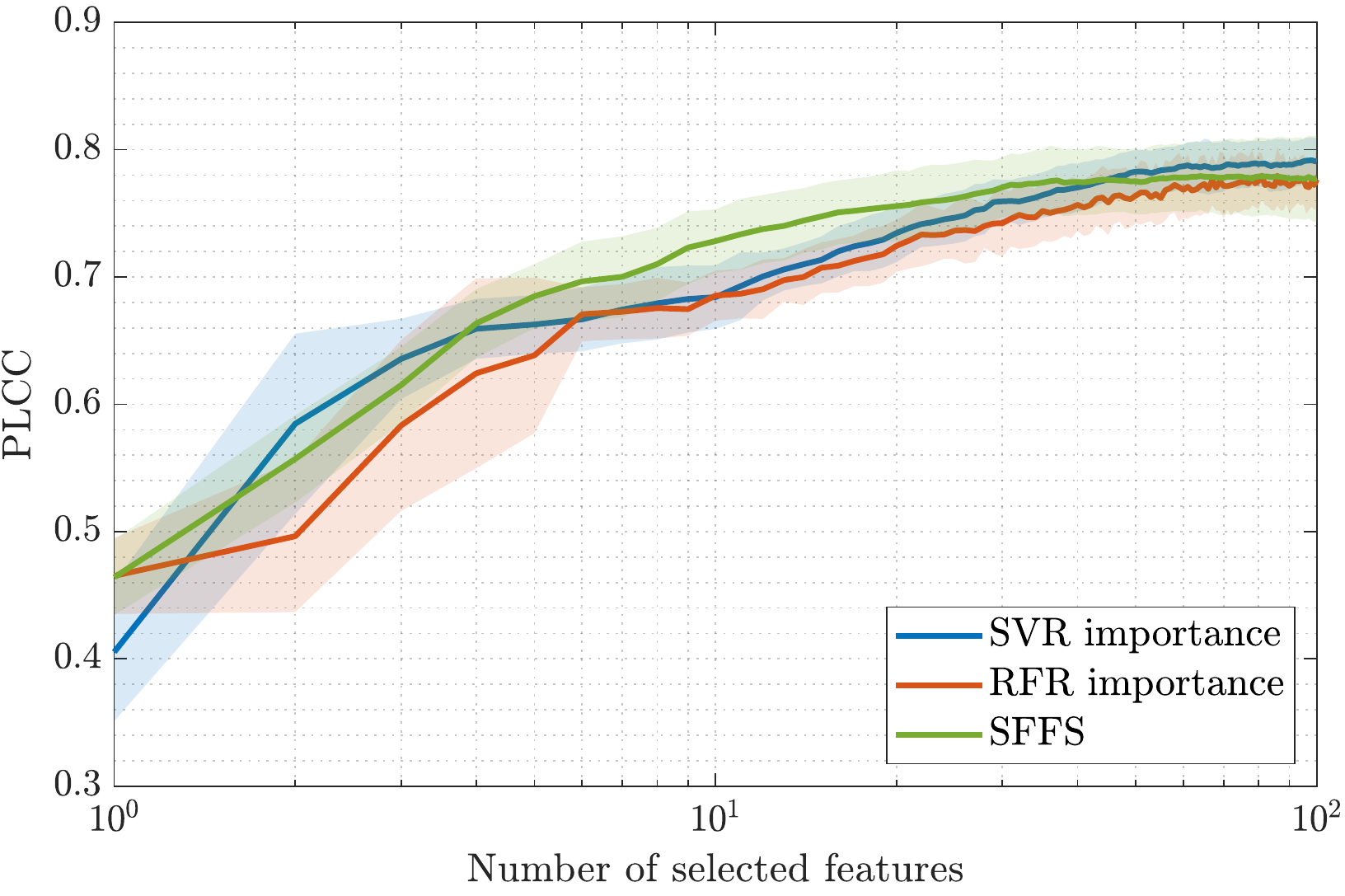} 
\caption{{Feature selection performance (PLCC) of three selected algorithms as a function of $k$ on the All-Combined\textsubscript{c} dataset. The shaded error bar denotes the standard deviation of PLCC over 10 iterations.}}
\label{fig:feat_sel}
\end{figure}

\begin{figure}[!t]
\def\xwidth{0.8}
\def\y{1}
\centering
\includegraphics[width=\xwidth\linewidth]{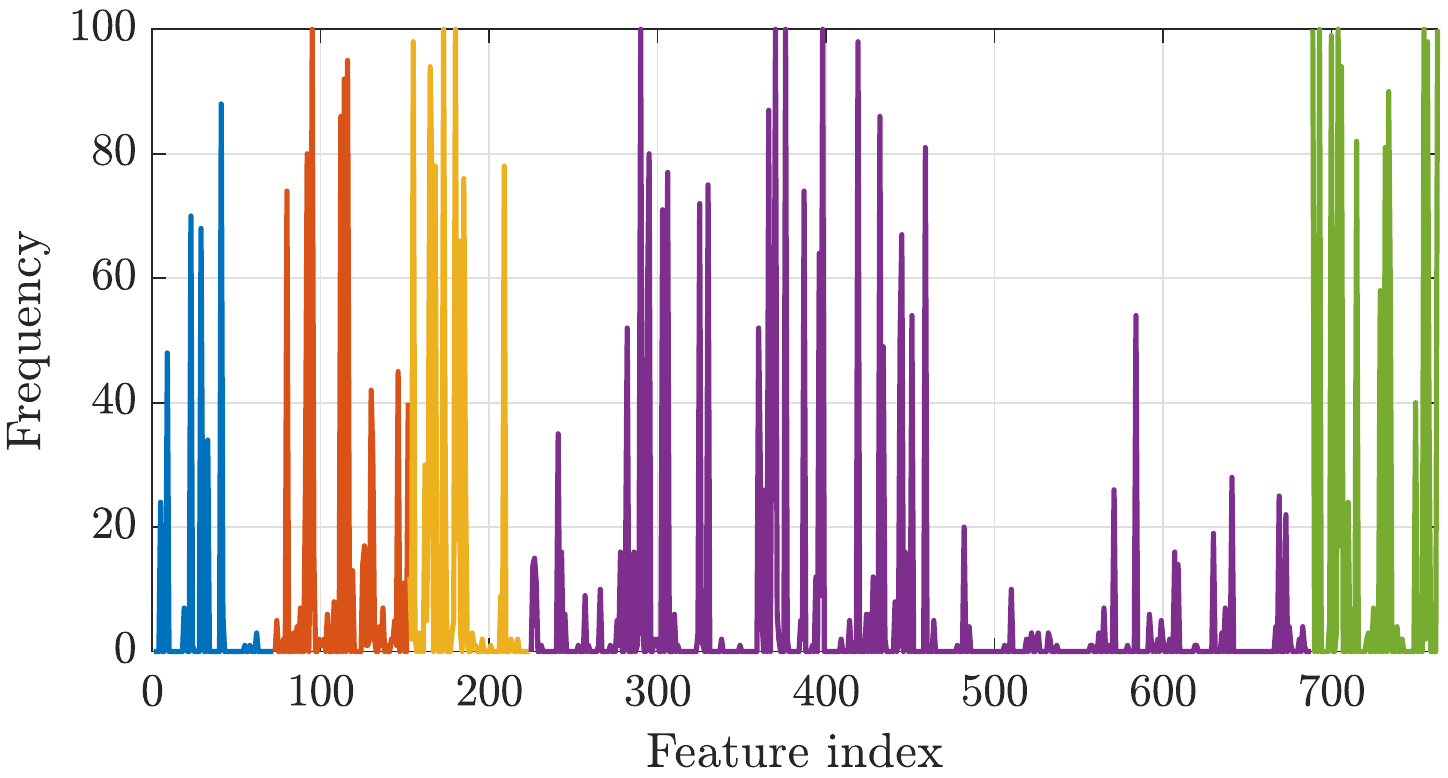} 
\caption{Visualization of the second step in feature selection: frequency of each feature being selected over 100 iterations of train-test splits using SVR importance selection method with fixed $k=60$.}
\label{fig:feat_import}
\end{figure}

We begin by constructing an initial feature set on top of existing high-performing, compute-efficient BVQA models and features, distilled through a feature selection program. The goal of feature selection is to choose an optimcal or sub-optimal feature subset $\mathcal{F}_{k}\in \mathbb{R}^k$ from the initial feature set $\mathcal{F}_{\mathrm{INIT}}\in \mathbb{R}^\mathrm{N}$ (where $k<{N}$) that achieves nearly top performance but with many fewer features.

\subsection{Feature Extraction}
\label{ssec:feat_extract}

We construct an initial feature set by selecting features from existing top-performing BVQA models. For practical reasons, we ignore features with high computational cost, e.g., certain features from DIIVINE, BLIINDS, C-DIIVINE, and V-BLIINDS. We also avoid using duplicate features in different models, such as the BRISQUE-like features in HIGRADE, and the C-DIIVINE features in V-BLIINDS. This filtering process yields the initial feature candidates, which we denote as BRISQUE, GM-LOG, HIGRADE-GRAD, FRIQUEE-LUMA, FRIQUEE-CHROMA, FRIQUEE-LMS, FRIQUEE-HS, TLVQM-LCF, and TLVQM-HCF. 

Inspired by the efficacy of standard deviation pooling as first introduced in GMSD \cite{xue2013gradient} and later also used in TLVQM \cite{korhonen2019two}, we calculate these spatial features every second frame within each sequentially cut non-overlapping one-second chunk, then we enrich the feature set by applying average and standard deviation pooling of frame-level features within each chunk, based on the hypothesis that the variation of spatial NSS features also correlates with the temporal properties of the video. Finally, all the chunk-wise feature vectors are average pooled \cite{tu2020comparative} across all the chunks to derive the final set of features for the entire video. Table \ref{table:feat_sum} indexes and summarizes the selected features in the initial feature set, yielding an overall 763-dimensional feature vector, $\mathcal{F}_\mathrm{INIT}\in\mathbb{R}^{763}$.

\begin{figure}[!t]
\centering
\def\xwidth{0.4}
\def\hswidth{0ex}
\subfloat[][{}]{\includegraphics[height=\xwidth\linewidth]{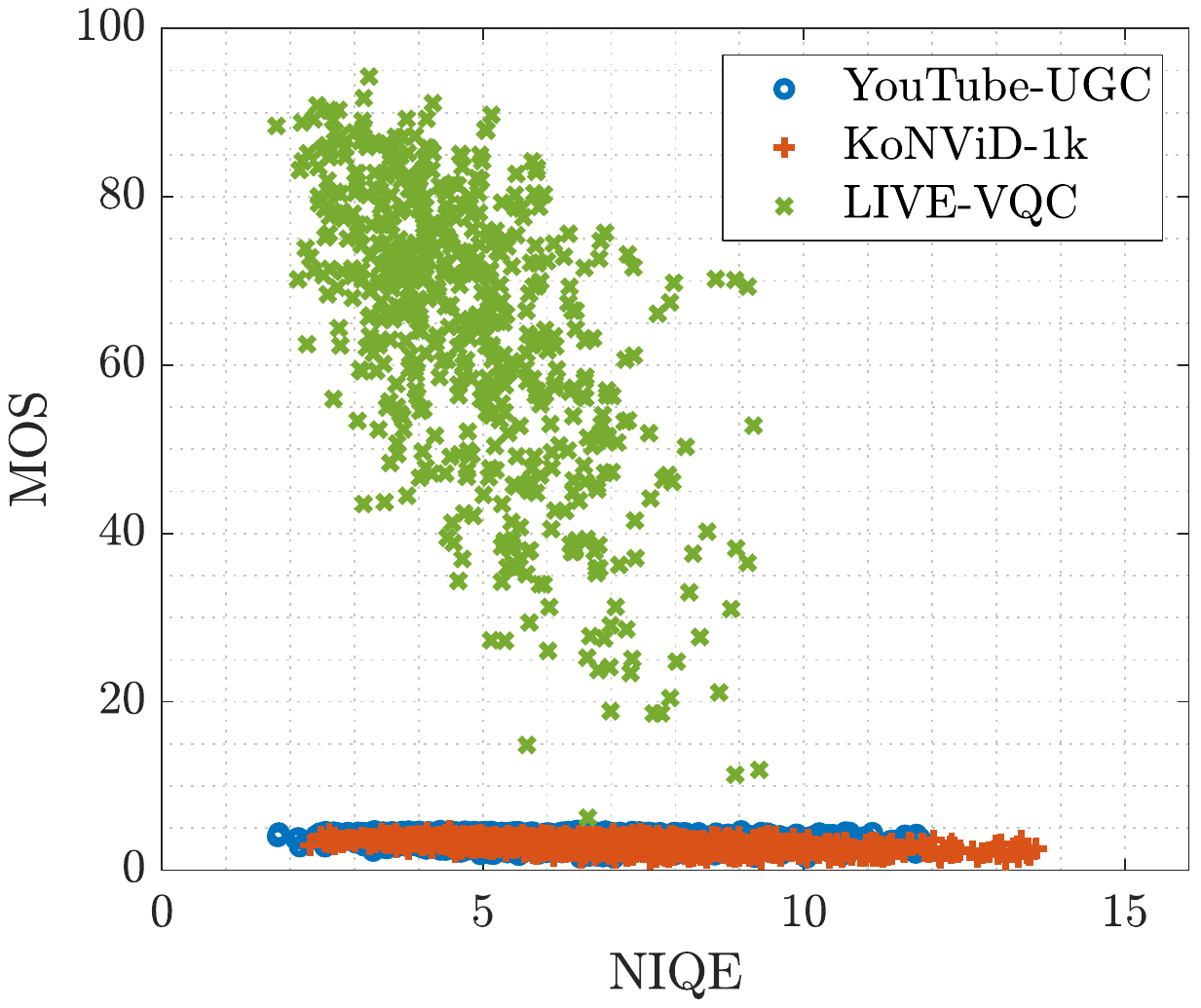} 
\label{fig:inlsa-a}} 
\hspace{\hswidth}
\subfloat[][{}]{\includegraphics[height=\xwidth\linewidth]{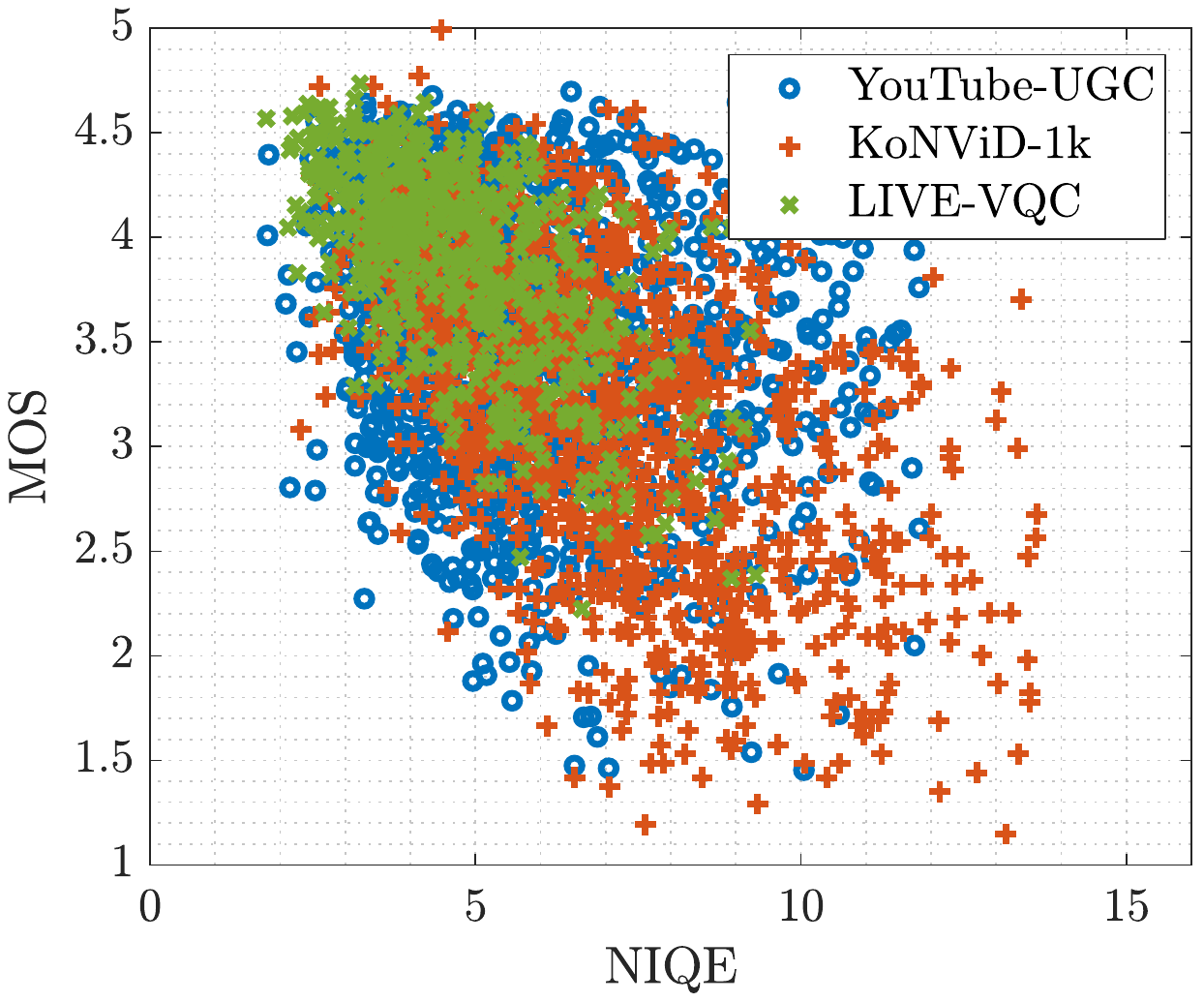} 
\label{fig:inlsa-b}} 
\caption{Scatter plots of MOS versus NIQE scores (a) before, and (b) after INLSA calibration \cite{pinson2003objective} using YouTube-UGC as the reference set.}
\label{fig:inlsa}
\end{figure}

\subsection{Feature Selection}
\label{ssec:feat_select}

We deploy two types of feature selection algorithms to distill the initial feature set. The first method is a model-based feature selector that utilizes a machine learning model to suggest features that are important. We employed the popular random forest (RF) to fit a regression model and eliminate the least significant features sorted by permutation importance. We also trained a support vector machine (SVM) with the linear kernel to rank the features, as a second model selector. Another sub-optimal solution is to apply a greedy search approach to find a good feature subset. Here we employed Sequential Forward Floating Selection (SFFS), and used SVM as the target regressor with its corresponding mean squared error between the predictions and MOS as the cost function. The mean squared error is calculated by cross-validation measures of predictive accuracy to avoid overfitting.

One problem with feature selection is that we do not know \textit{a priori} what $k$ to select, i.e., how many features are needed. Therefore, we conducted a two-step feature selection procedure. First, we evaluated the feature selection methods as a function of $k$ via 10 train-test iterations, to select the best algorithm with corresponding optimal $k$. Figure \ref{fig:feat_sel} shows the median PLCC (defined in Section \ref{ssec:eval_proto}) performance with respect to $k$ for different feature selection models, based on which we finally chose the SVM importance method with $k=60$ in our next experiments. In the second step, we applied the best feature selection algorithm with the fixed best $k$ over 100 random train-test splits. On each iteration, a subset is selected from the feature selector, based on which the frequency of each feature over the iterations is counted, and the $j$ most frequently occurring features are included into the final feature set. Figure \ref{fig:feat_import} shows the frequency of each feature being selected over 100 random splits in the second step. This selection process is implemented on a combined dataset constructed from three independent databases, as described in Section \ref{ssec:eval_proto}. Table \ref{table:feat_sum} summarizes the results of the feature selection procedure (SVR importance with $k=60$), yielding the final proposed VIDEVAL model.

\begin{figure}[!t]
\centering
\def\xwidth{0.98}
\includegraphics[width=\xwidth\linewidth]{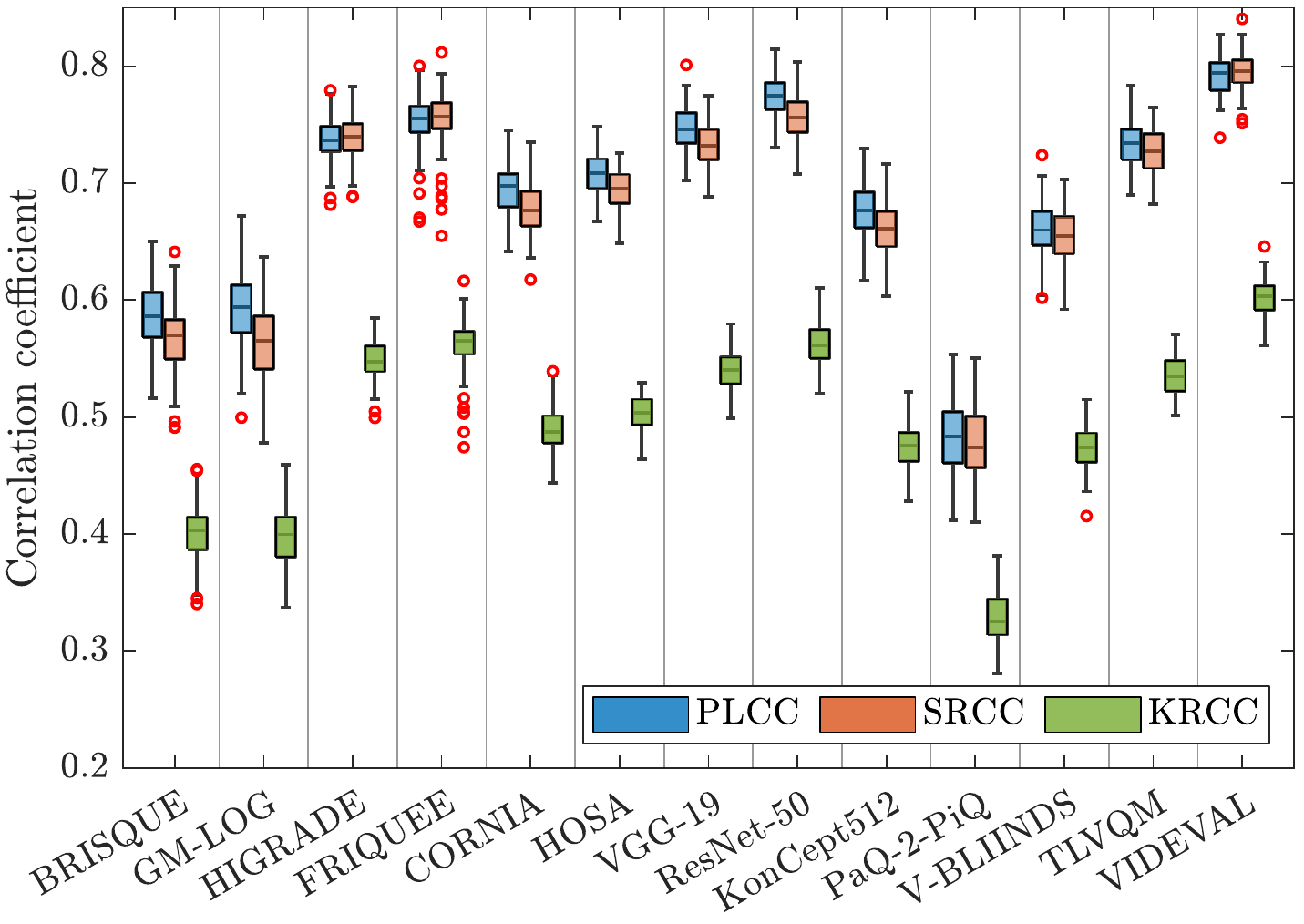}
\caption{Box plots of PLCC, SRCC, and KRCC of evaluated learning-based BVQA algorithms on the All-Combined\textsubscript{c} dataset over 100 random splits. For each box, median is the central box, and the edges of the box represent 25th and 75th percentiles, while red circles denote outliers.}
\label{fig:all_boxplot}
\end{figure}

\begin{table}[!t]
\setlength{\tabcolsep}{4pt}
\renewcommand{\arraystretch}{1.1}
\centering
\caption{Performance comparison of evaluated opinion-unaware ``completely blind'' BVQA models.}
\label{table:compete-blind}
\begin{tabular}{llcccc}
\toprule
\textsc{Dataset}  & \textsc{Model} \textbackslash\ \textsc{Metric} & SRCC$\uparrow$   & KRCC$\uparrow$    & PLCC$\uparrow$     & RMSE$\downarrow$   \\
\hline\\[-1.em]
\multirow{3}{*}{KoNViD} & NIQE (1 fr/sec)    & 0.5417 & 0.3790  & 0.5530  & 0.5336  \\
& ILNIQE (1 fr/sec)  & 0.5264  & 0.3692  & 0.5400  & 0.5406  \\
& VIIDEO  & 0.2988 & 0.2036  & 0.3002 & 0.6101  \\
\hline\\[-1.em]
\multirow{3}{*}{LIVE-C}   & NIQE (1 fr/sec)    & 0.5957  & 0.4252  & 0.6286 & 13.110  \\
& ILNIQE (1 fr/sec)  & 0.5037 & 0.3555 & 0.5437  & 14.148  \\
& VIIDEO  & 0.0332 & 0.0231 & 0.2146 & 16.654  \\
\hline\\[-1.em]
\multirow{3}{*}{YT-UGC}
& NIQE (1 fr/sec) & 0.2379 & 0.1600 & 0.2776 & 0.6174  \\
& ILNIQE (1 fr/sec) & 0.2918 & 0.1980 & 0.3302 & 0.6052  \\
& VIIDEO  & 0.0580 & 0.0389  & 0.1534  & 0.6339 \\
\hline\\[-1.em]
\multirow{3}{*}{All-Comb}
& NIQE (1 fr/sec) & 0.4622  & 0.3222 & 0.4773  & 0.6112 \\
& ILNIQE (1 fr/sec) & 0.4592 & 0.3213  & 0.4741  & 0.6119  \\
& VIIDEO  & 0.1039  & 0.0688  & 0.1621 & 0.6804 \\
\bottomrule
\end{tabular}
\end{table}

\begin{table*}[!t]
\setlength{\tabcolsep}{2.7pt}
\renewcommand{\arraystretch}{1.1}
\centering
\begin{threeparttable}
\caption{Performance comparison of evaluated BVQA models on the four benchmark datasets. The \underline{\textbf{underlined}} and \textbf{boldfaced} entries indicate the best and top three performers on each database for each performance metric, respectively.}
\label{table:eval_svr}
\begin{tabular}{lcccccccccc}
\toprule
\textsc{Dataset} & \multicolumn{4}{c}{KoNViD-1k} & & \multicolumn{4}{c}{LIVE-VQC} \\ \cline{2-5}\cline{7-10}\\[-1.em]

\textsc{Model} \textbackslash\ \textsc{Metric} & \textsc{SRCC$\uparrow$ (std)}    & \textsc{KRCC$\uparrow$ (std)}    & \textsc{PLCC$\uparrow$ (std)}     & \textsc{RMSE$\downarrow$ (std)} & & \textsc{SRCC$\uparrow$ (std)}    & \textsc{KRCC$\uparrow$ (std)}    & \textsc{PLCC$\uparrow$ (std)}     & \textsc{RMSE$\downarrow$ (std)}   \\ 
\hline\\[-1.em]
BRISQUE (1 fr/sec) & 0.6567 (.035) & 0.4761 (.029) & 0.6576 (.034) & 0.4813 (.022) & & 0.5925 (.068) & 0.4162 (.052) & 0.6380 (.063) & 13.100 (.796) \\
GM-LOG (1 fr/sec)  & 0.6578 (.032) & 0.4770 (.026) & 0.6636 (.031) & 0.4818 (.022) & & 0.5881 (.068) & 0.4180 (.052) & 0.6212 (.063) & 13.223 (.822) \\
HIGRADE (1 fr/sec) & 0.7206 (.030) & 0.5319 (.026) & 0.7269 (.028) & 0.4391 (.018) & & 0.6103 (.068) & 0.4391 (.054) & 0.6332 (.065) & 13.027 (.904) \\
FRIQUEE (1 fr/sec) & 0.7472 (.026) & 0.5509 (.024) & 0.7482 (.025) & 0.4252 (.017) & & 0.6579 (.053) & 0.4770 (.043) & 0.7000 (.058) & 12.198 (.914) \\
CORNIA (1 fr/sec)  & 0.7169 (.024) & 0.5231 (.021) & 0.7135 (.023) & 0.4486 (.018) & & 0.6719 (.047) & 0.4849 (.039) & 0.7183 (.042) & 11.832 (.700) \\
HOSA (1 fr/sec)    & 0.7654 (.022) & 0.5690 (.021) & 0.7664 (.020) & 0.4142 (.016) & & 0.6873 (.046) & 0.5033 (.039) & {\textbf{0.7414 (.041)}} & {\textbf{11.353 (.747)}} \\
VGG-19 (1 fr/sec) & {\textbf{0.7741 (.028)}} & {\textbf{0.5841 (.027)}} & {\textbf{0.7845 (.024)}} & {\textbf{0.3958 (.017)}} & & 0.6568 (.053) & 0.4722 (.044) & 0.7160 (.048) & 11.783 (.696) \\
ResNet-50 (1 fr/sec) & \textbf{\underline{0.8018} (.025)} & {\textbf{\underline{0.6100} (.024)}} & {\textbf{\underline{0.8104} (.022)}} & {\textbf{\underline{0.3749} (.017)}} & & 0.6636 (.051) & 0.4786 (.042) & 0.7205 (.043) &  11.591 (.733) \\
KonCept512 (1 fr/sec) & 0.7349 (.025) & 0.5425 (.023) & 0.7489 (.024) & 0.4260 (.016) & & 0.6645 (.052) & 0.4793 (.045) & 0.7278 (.046) & 11.626 (.767) \\
PaQ-2-PiQ (1 fr/sec) & 0.6130 (.032) & 0.4334 (.026) & 0.6014 (.033) & 0.5148 (.019)  & & 0.6436 (.045) & 0.4568 (.035) & 0.6683 (.044) & 12.619 (.848) \\
V-BLIINDS          & 0.7101 (.031) & 0.5188 (.026) & 0.7037 (.030) & 0.4595 (.023) & & {\textbf{0.6939 (.050)}} & {\textbf{0.5078 (.042)}} & 0.7178 (.050) & 11.765 (.828) \\
TLVQM              & 0.7729 (.024) & 0.5770 (.022) & 0.7688 (.023) & 0.4102 (.017) & & {\textbf{\underline{0.7988} (.036)}} & {\textbf{\underline{0.6080} (.037)}} & {\textbf{\underline{0.8025} (.036)}} & {\textbf{\underline{10.145} (.818)}} \\
VIDEVAL & \textbf{0.7832 (.021)} & \textbf{0.5845 (.021)} & \textbf{0.7803 (.022)} & \textbf{0.4026 (.017)} & & \textbf{0.7522 (.039)} & \textbf{0.5639 (.036)} & \textbf{0.7514 (.042)} & \textbf{11.100 (.810)}    \\
\midrule
\textsc{Dataset} & \multicolumn{4}{c}{YouTube-UGC} & & \multicolumn{4}{c}{All-Combined\textsubscript{c}\hyperlink{all_exp}{$^\dagger$}} \\ \cline{2-5}\cline{7-10}\\[-1.em]

\textsc{Model} \textbackslash\ \textsc{Metric} & \textsc{SRCC$\uparrow$ (std)}    & \textsc{KRCC$\uparrow$ (std)}    & \textsc{PLCC$\uparrow$ (std)}     & \textsc{RMSE$\downarrow$ (std)} & & \textsc{SRCC$\uparrow$ (std)}    & \textsc{KRCC$\uparrow$ (std)}    & \textsc{PLCC$\uparrow$ (std)}     & \textsc{RMSE$\downarrow$ (std)}   \\
\hline\\[-1.em]
BRISQUE (1 fr/sec) & 0.3820 (.051) & 0.2635 (.036) & 0.3952 (.048) & 0.5919 (.021) & & 0.5695 (.028) & 0.4030 (.022) & 0.5861 (.027) & 0.5617 (.016) \\
GM-LOG (1 fr/sec)  & 0.3678 (.058) & 0.2517 (.041) &  0.3920 (.054) &  0.5896 (.022) & & 0.5650 (.029) & 0.3995 (.022) & 0.5942 (.030) & 0.5588 (.014) \\
HIGRADE (1 fr/sec) & {\textbf{0.7376 (.033)}} & {\textbf{0.5478 (.028)}} &  {\textbf{0.7216 (.033)}} & {\textbf{0.4471 (.024)}} & & 0.7398 (.018) & 0.5471 (.016) & {0.7368 (.019)} & {0.4674 (.015)} \\
FRIQUEE\hyperlink{fri_exp}{$^{\star}$} (1 fr/sec) & {\textbf{{0.7652} (.030)}}  & {\textbf{{0.5688} (.026)}} & {\textbf{{0.7571} (.032)}} & {\textbf{{0.4169} (.023)}} & & {\textbf{{0.7568} (.023)}} & {\textbf{{0.5651} (.021)}} & {\textbf{0.7550 (.022)}} & {\textbf{0.4549 (.018)}} \\
CORNIA (1 fr/sec)  & 0.5972 (.041) & 0.4211 (.032) & 0.6057 (.039) & 0.5136 (.024) & & 0.6764 (.021) & 0.4846 (.017) & 0.6974 (.020) & 0.4946 (.013) \\
HOSA (1 fr/sec)   & 0.6025 (.034) & 0.4257 (.026) & 0.6047 (.034) & 0.5132 (.021) & & 0.6957 (.018) & 0.5038 (.015) & 0.7082 (.016) & 0.4893 (.013) \\
VGG-19 (1 fr/sec) & 0.7025 (.028) & 0.5091 (.023) & 0.6997 (.028) & 0.4562 (.020) & & 0.7321 (.018) & 0.5399 (.016) & 0.7482 (.017) & {0.4610 (.013)}  \\
ResNet-50 (1 fr/sec) & 0.7183 (.028) & 0.5229 (.024) & 0.7097 (.027) & 0.4538 (.021) & & {\textbf{0.7557 (.017)}} & {\textbf{0.5613 (.016)}} & {\textbf{{0.7747} (.016)}} & {\textbf{{0.4385} (.013)}} \\
KonCept512 (1 fr/sec) & 0.5872 (.039) & 0.4101 (.030) & 0.5940 (.041) & 0.5135 (.022)  & & 0.6608 (.022) & 0.4759 (.018) & 0.6763 (.022) & 0.5091 (.014) \\
PaQ-2-PiQ (1 fr/sec) & 0.2658 (.047) & 0.1778 (.032) & 0.2935 (.049) & 0.6153 (.019) & &    0.4727 (.029) & 0.3242 (.021) & 0.4828 (.029) & 0.6081 (.015) \\
V-BLIINDS          &  0.5590 (.049) & 0.3899 (.036) & 0.5551 (.046) & 0.5356 (.022) & & 0.6545 (.023) & 0.4739 (.019) & 0.6599 (.023) & 0.5200 (.016) \\
TLVQM              & 0.6693 (.030) &  0.4816 (.025) & {0.6590} ({.030}) & {0.4849} ({.022}) & & {0.7271 (.018)} & {0.5347 (.016)} & {0.7342 (.018)} & {0.4705 (.013)} \\
VIDEVAL\hyperlink{fri_exp}{$^{\star}$} & \textbf{\underline{0.7787} (.025)} & \textbf{\underline{0.5830} (.023)} &  \textbf{\underline{0.7733} (.025)} & \textbf{\underline{0.4049} (.021)} & & {\textbf{\underline{0.7960} (.015)}} & {\textbf{\underline{0.6032} (.014)}} & {\textbf{\underline{0.7939} (.015)}} & {\textbf{\underline{0.4268} (.015)}} \\

\bottomrule
\end{tabular}

\begin{tablenotes}[para,flushleft]
\footnotesize
\item \hypertarget{fri_exp}{$^\star$}FRIQUEE and VIDEVAL were evaluated on a subset of 1,323 color videos in YouTube-UGC, denoted YouTube-UGC\textsubscript{c}, since it yields numerical errors when calculating on the remaining $57$ grayscale videos. For the other BVQA models evaluated, no significant difference was observed when evaluated on YouTube-UGC\textsubscript{c} versus YouTube-UGC, and hence we still report the results on YouTube-UGC. \\
\item \hypertarget{all_exp}{$^\dagger$}For a fair comparison, we only combined and calibrated (via INLSA \cite{pinson2003objective}) all the color videos from these three databases to obtain the combined dataset, i.e., All-Combined\textsubscript{c} (3,108)\! $=$\! KoNViD-1k (1,200)\! $+$\! LIVE-VQC (585)\! $+$\! YouTube-UGC\textsubscript{c} (1,323).
\end{tablenotes}
\end{threeparttable}
\end{table*}

\begin{figure*}[!t]
\captionsetup[subfigure]{justification=centering}
\centering
\def\hswidth{0em}
\def\xwidth{0.188}
\subfloat[BRISQUE]{\includegraphics[width=\xwidth\textwidth]{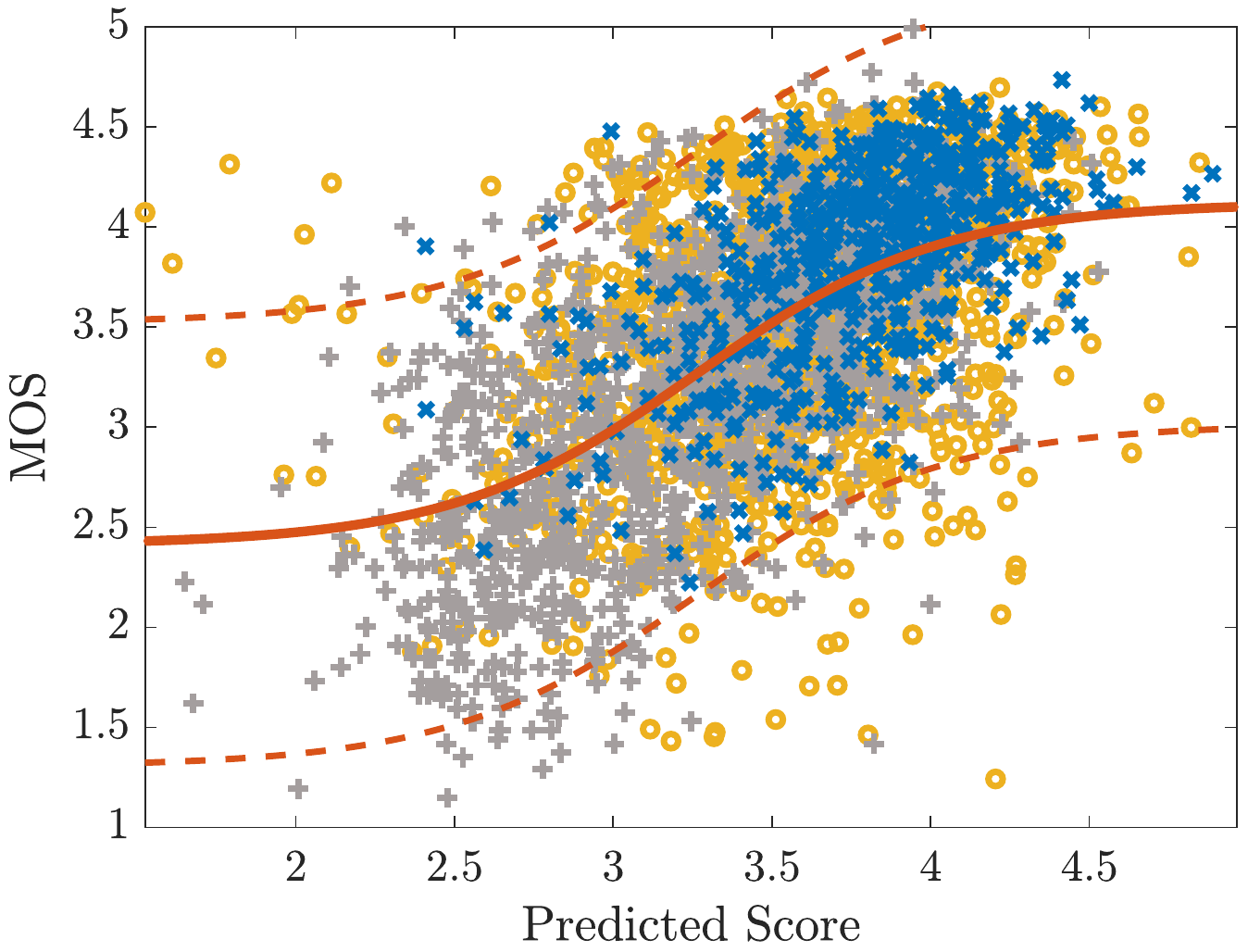}
\label{fig1a}}
\hspace{\hswidth}
\subfloat[GM-LOG]{\includegraphics[width=\xwidth\textwidth]{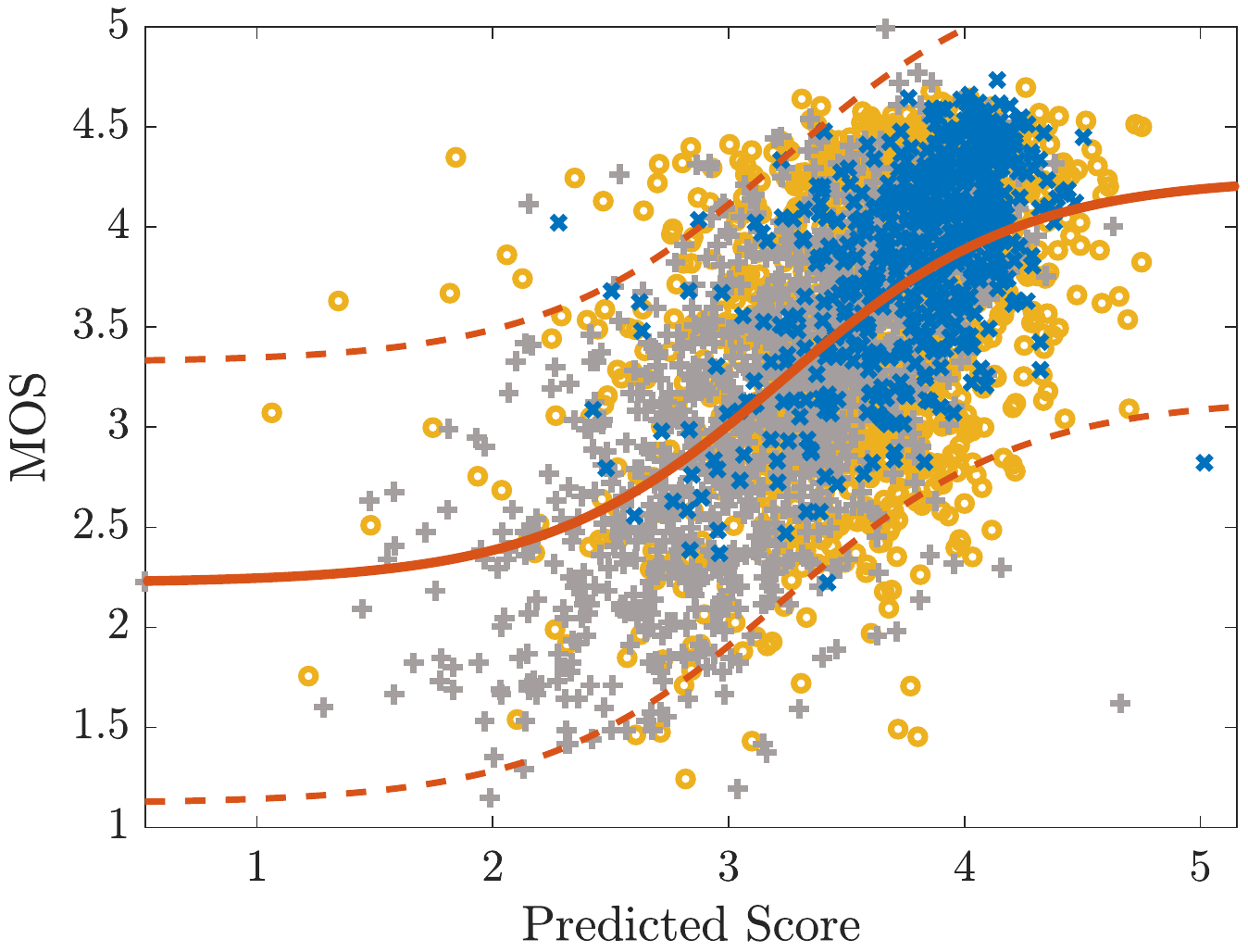}
\label{fig1b}}
\hspace{\hswidth}
\subfloat[HIGRADE]{\includegraphics[width=\xwidth\textwidth]{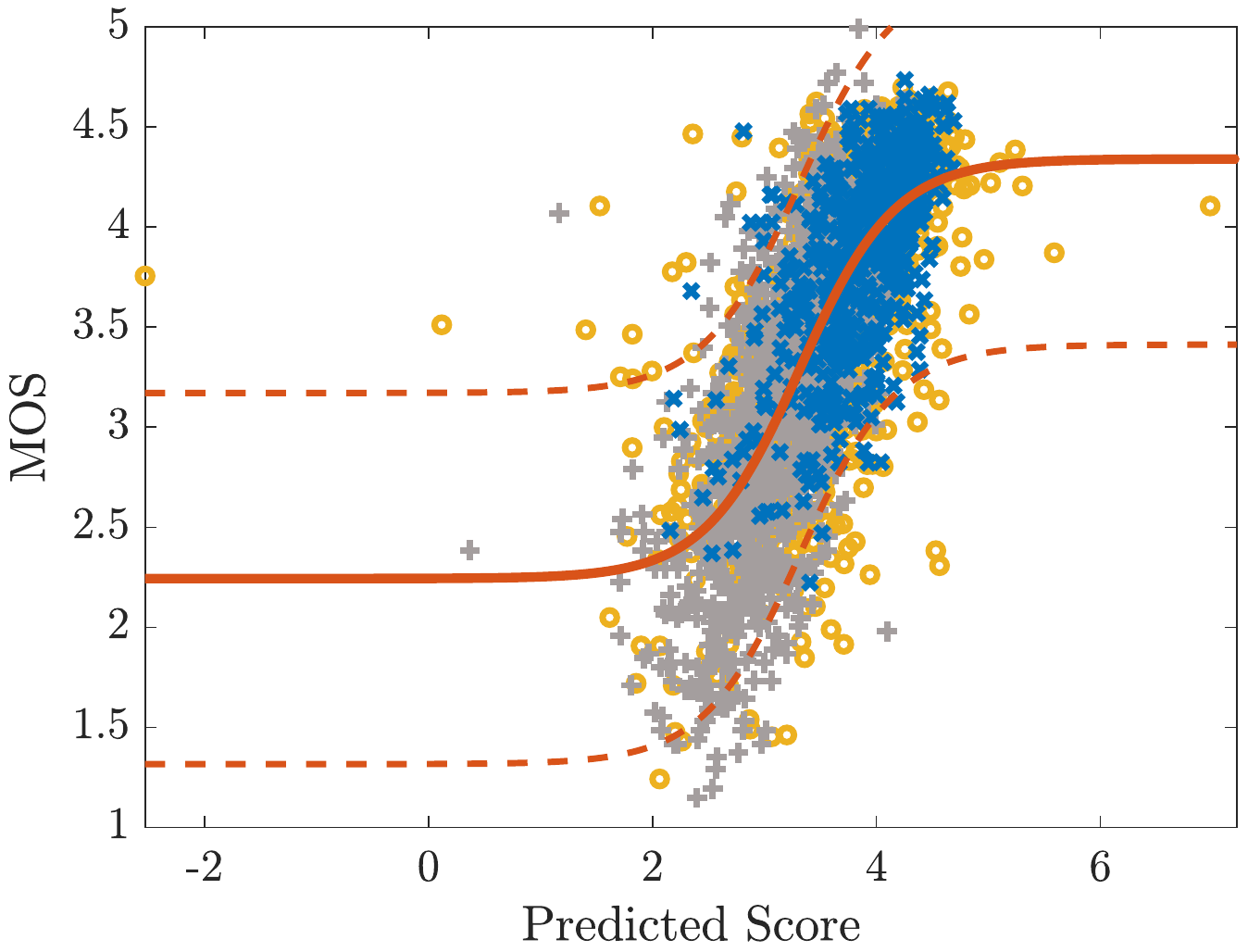}
\label{fig1c}}
\hspace{\hswidth}
\subfloat[FRIQUEE]{\includegraphics[width=\xwidth\textwidth]{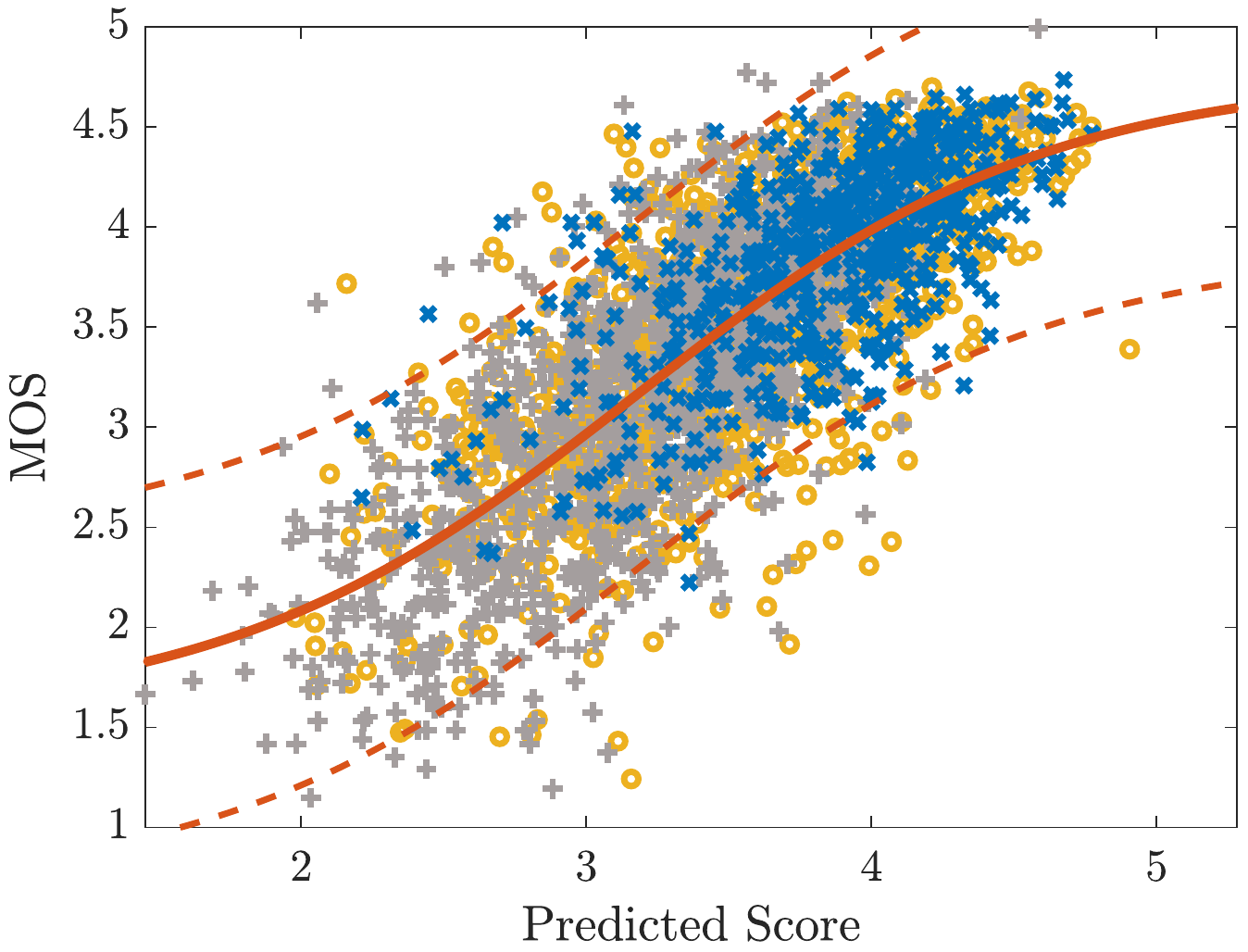}
\label{fig1d}}
\subfloat[CORNIA]{\includegraphics[width=\xwidth\textwidth]{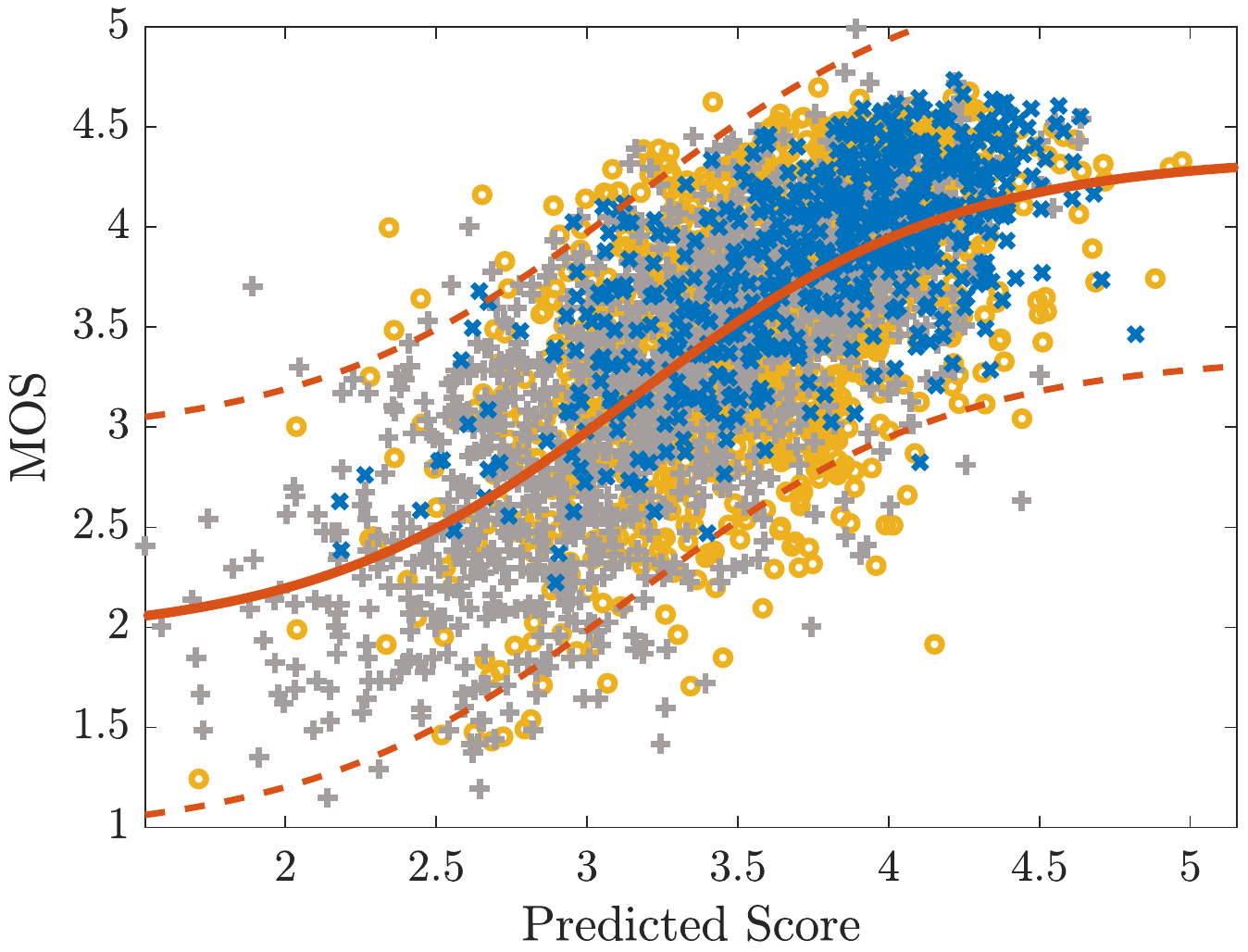}
\label{fig1e}} \\[-2ex]

\subfloat[HOSA]{\includegraphics[width=\xwidth\textwidth]{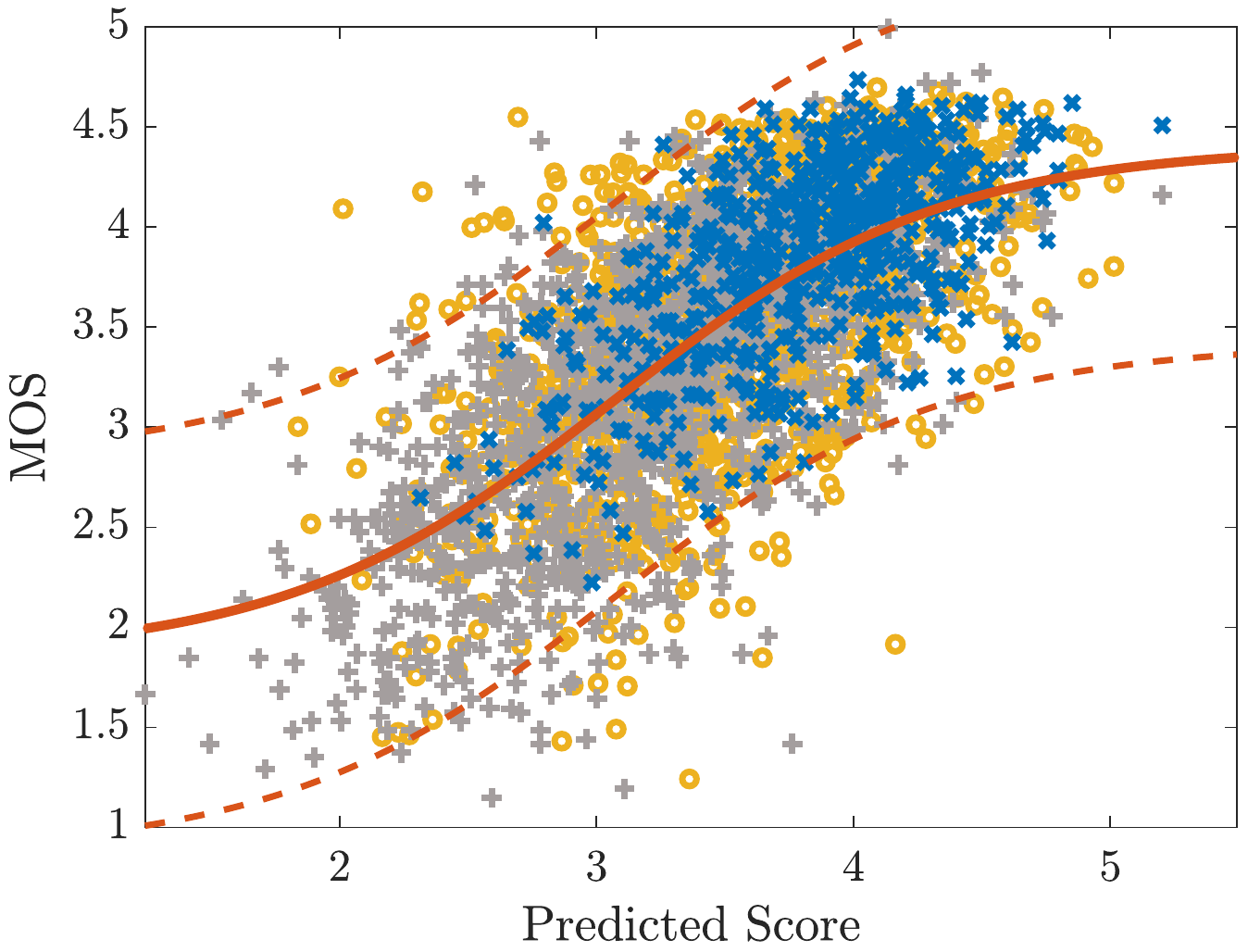}
\label{fig1f}}
\hspace{\hswidth}
\subfloat[VGG-19]{\includegraphics[width=\xwidth\textwidth]{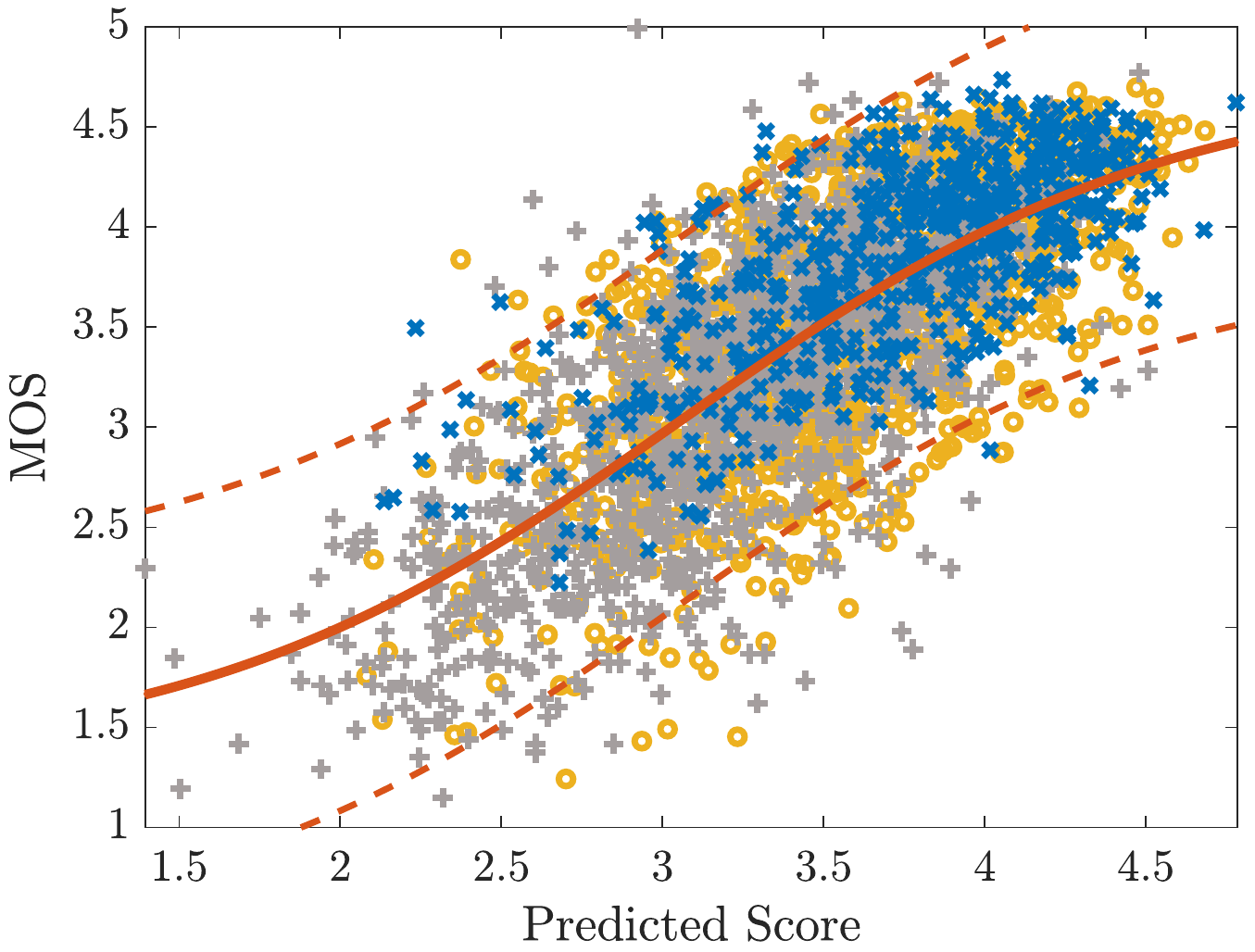}
\label{fig1g}}
\hspace{\hswidth}
\subfloat[ResNet-50]{\includegraphics[width=\xwidth\textwidth]{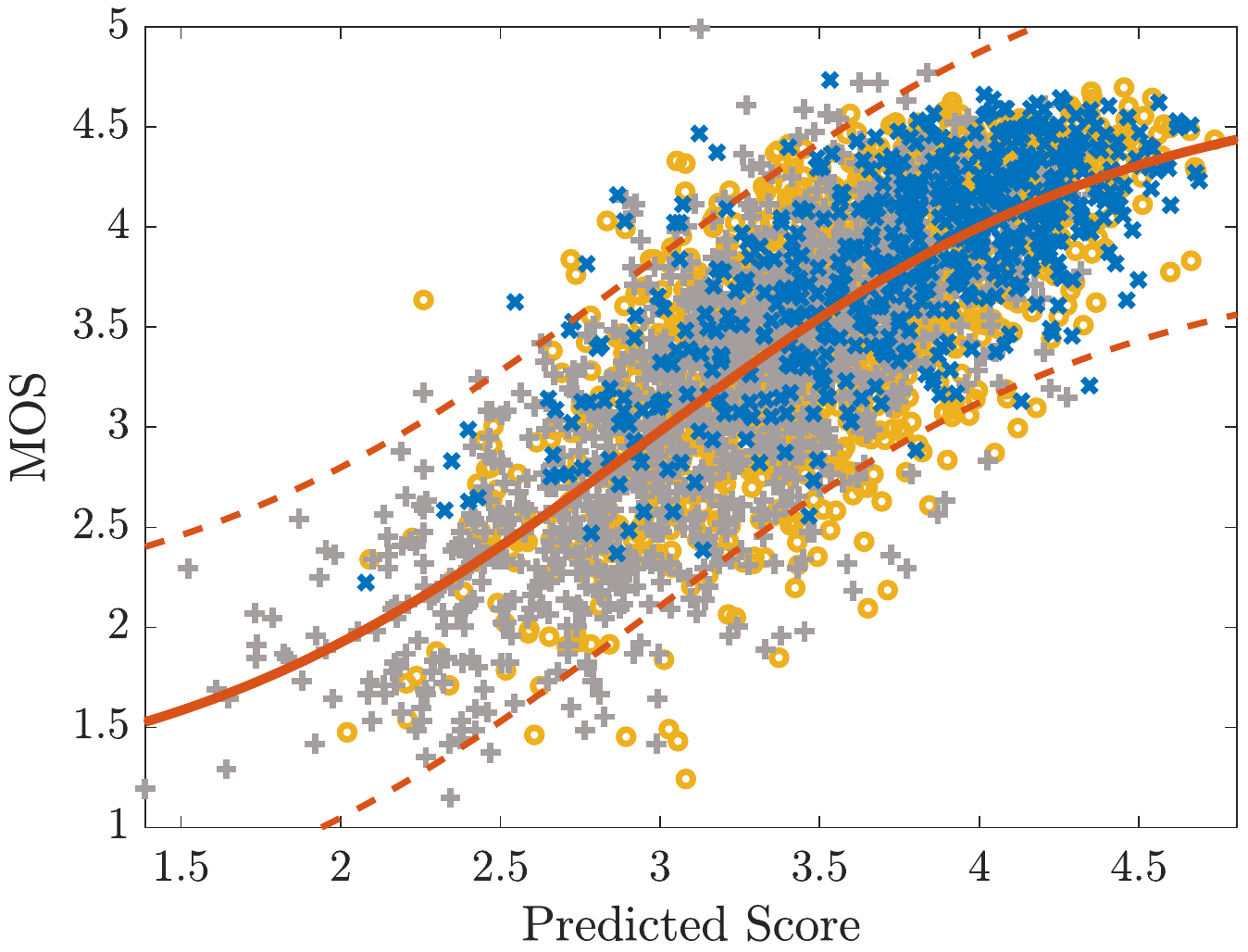}
\label{fig1h}} 
\hspace{\hswidth}
\subfloat[KonCept512]{\includegraphics[width=\xwidth\textwidth]{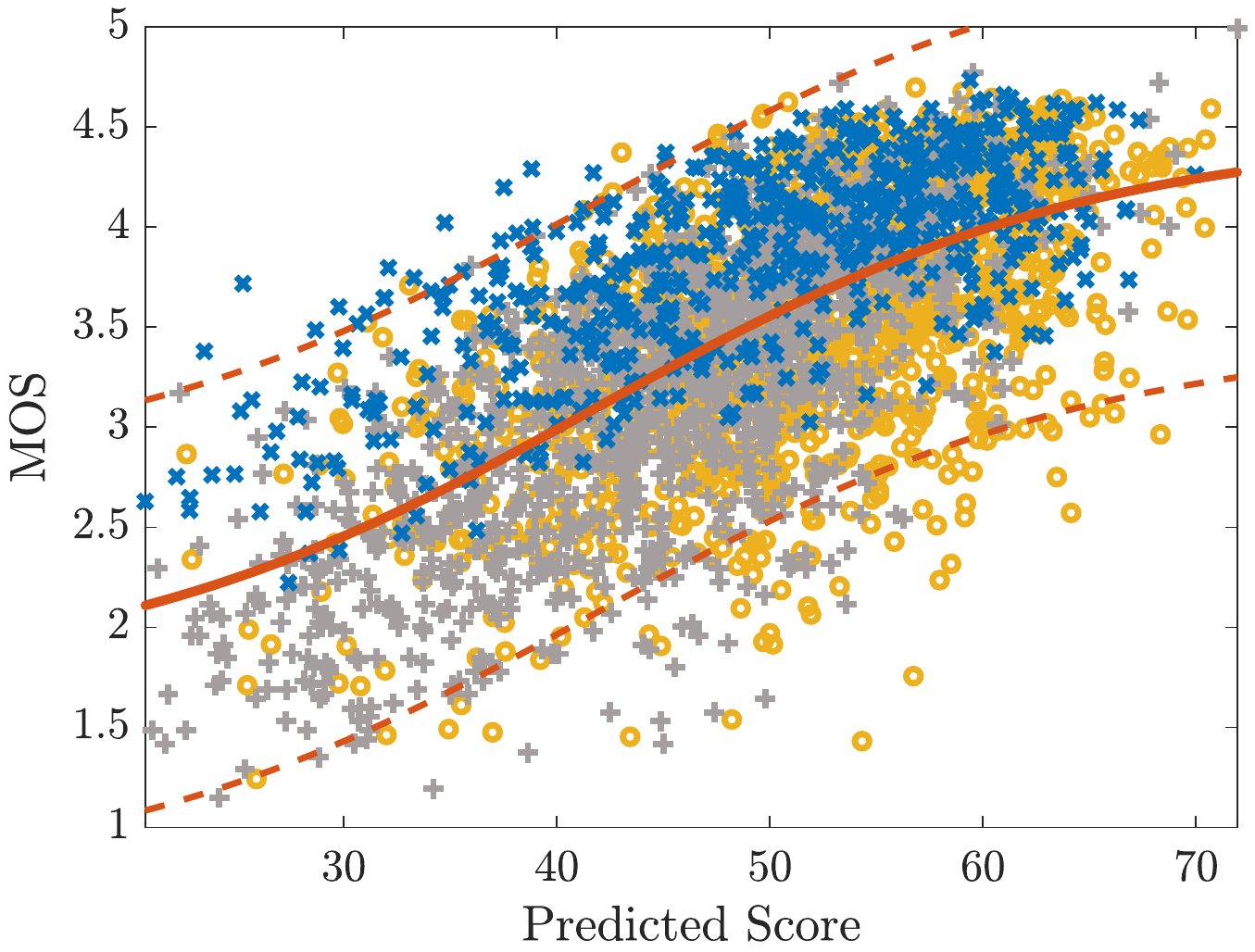}
\label{fig1i}} 
\hspace{\hswidth}
\subfloat[PaQ-2-PiQ]{\includegraphics[width=\xwidth\textwidth]{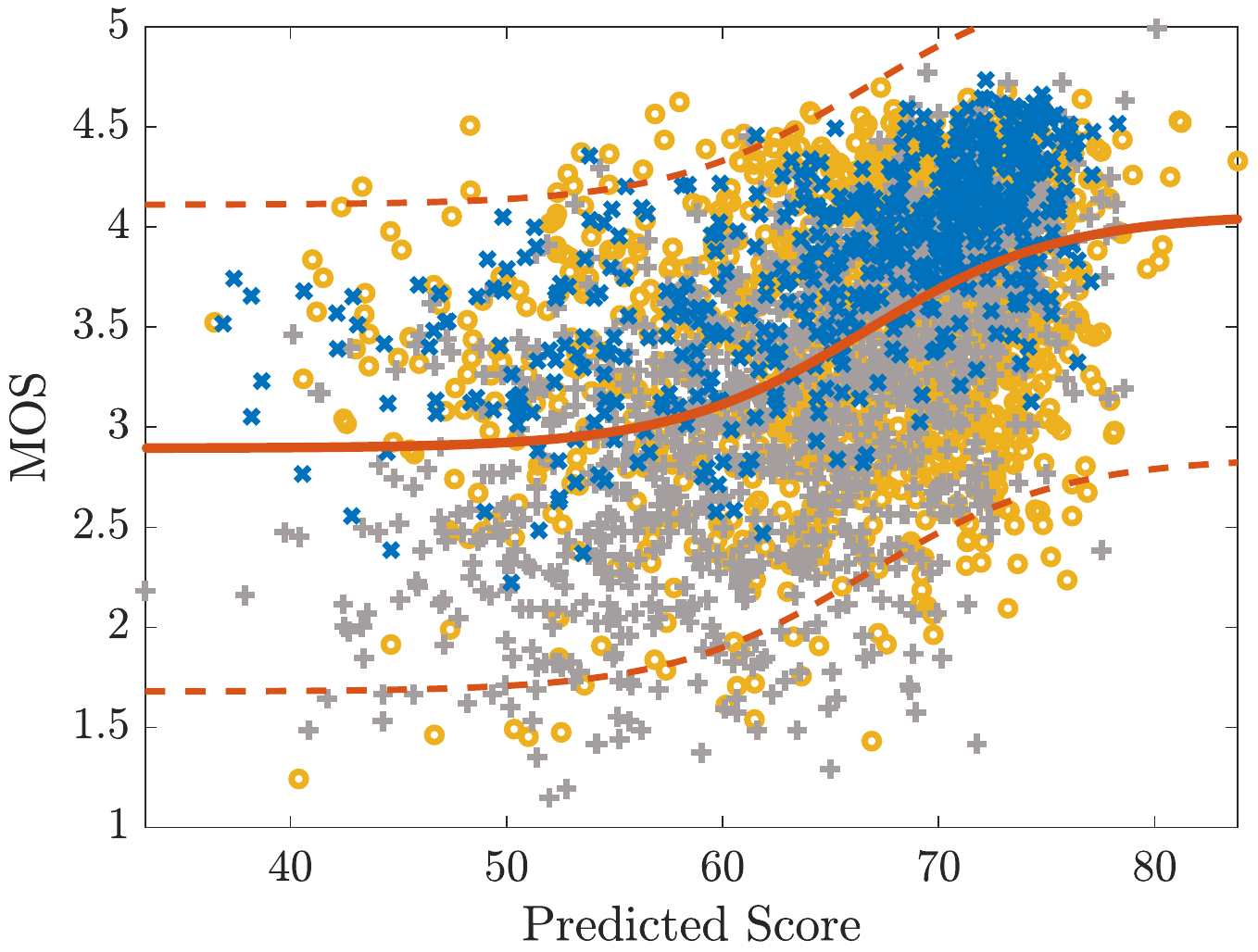}
\label{fig1j}} \\ [-2ex]

\subfloat[V-BLIINDS]{\includegraphics[width=\xwidth\textwidth]{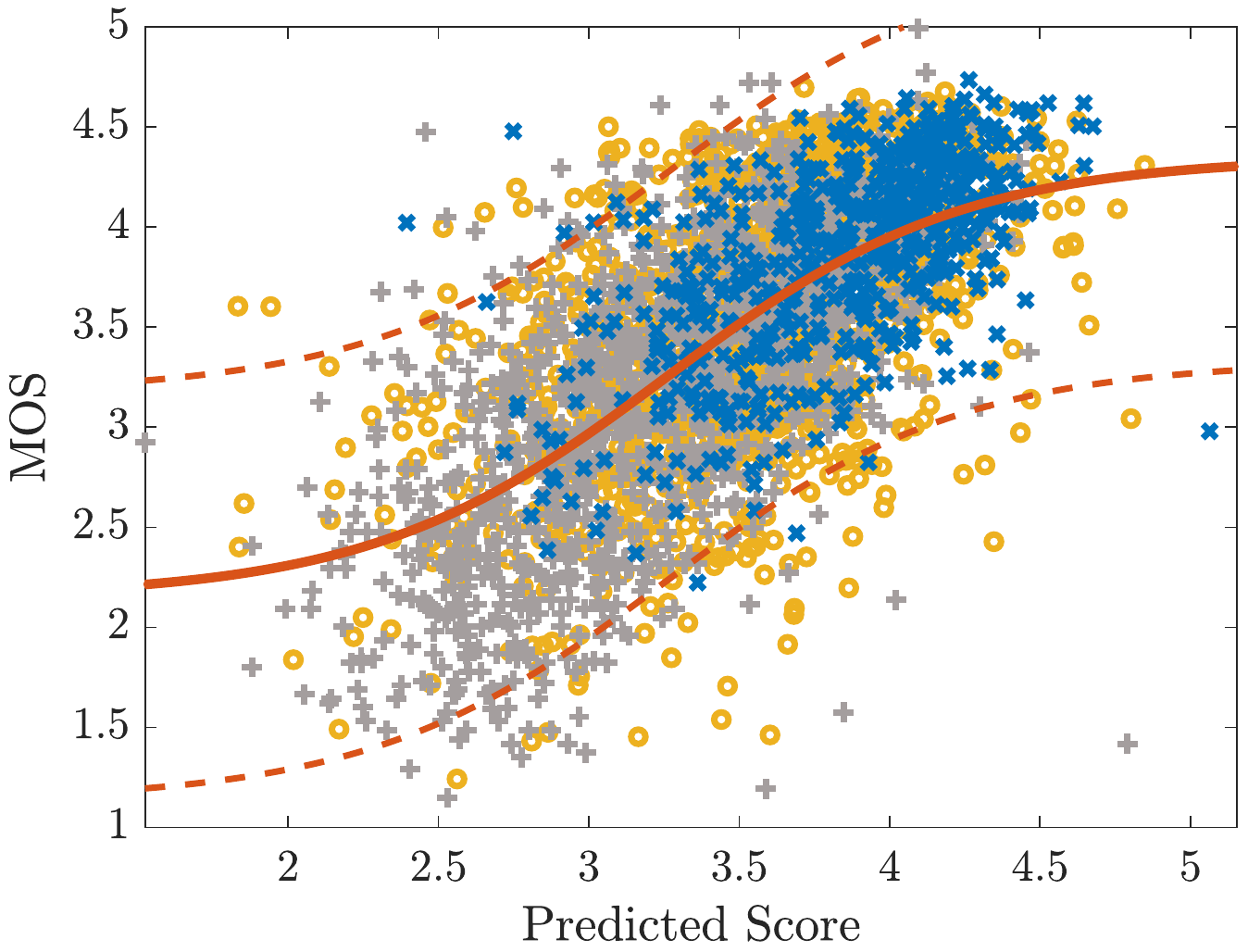}
\label{fig1k}}
\hspace{\hswidth}
\subfloat[TLVQM]{\includegraphics[width=\xwidth\textwidth]{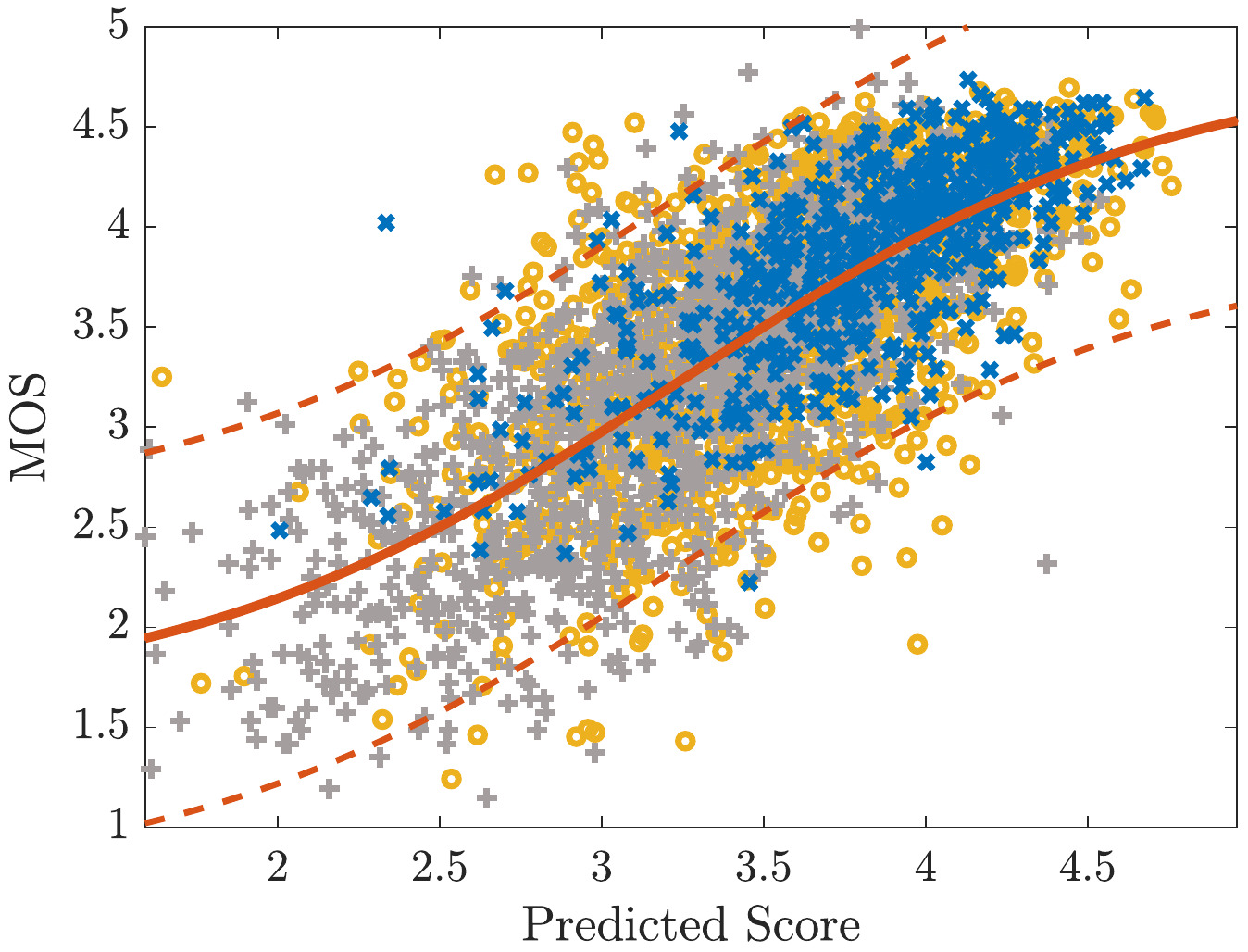}
\label{fig1l}}
\hspace{\hswidth}
\subfloat[{VIDEVAL}]{\includegraphics[width=\xwidth\textwidth]{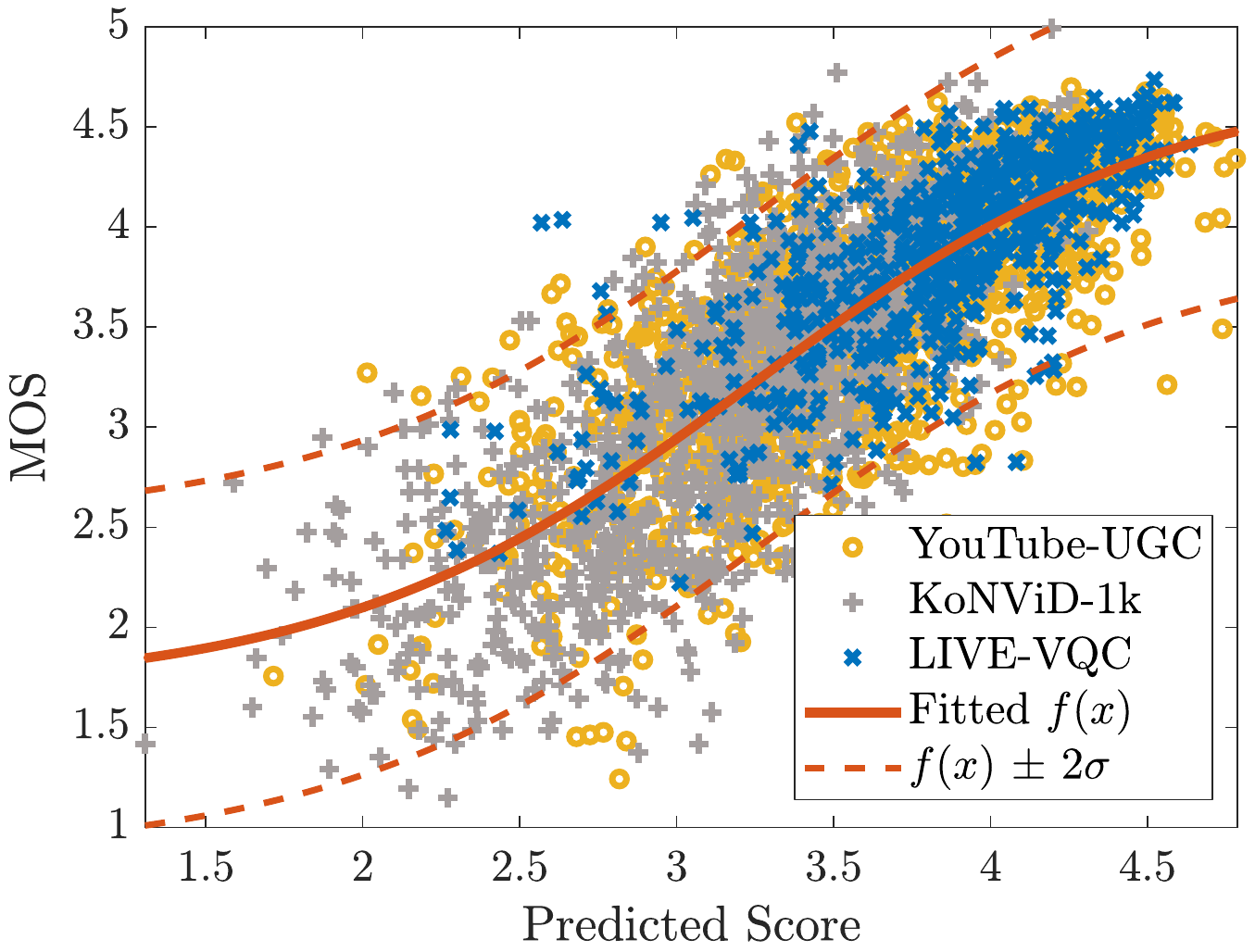}
\label{fig1m}}
\caption{Scatter plots and nonlinear logistic fitted curves of VQA models versus MOS trained with a grid-search SVR using $k$-fold cross-validation on the All-Combined\textsubscript{c} set. (a) BRISQUE (1 fr/sec), (b) GM-LOG (1 fr/sec), (c) HIGRADE (1 fr/sec), (d) FRIQUEE (1 fr/sec), (e) CORNIA (1 fr/sec), (f) HOSA (1 fr/sec), (g) VGG-19 (1 fr/sec), (h) ResNet-50 (1 fr/sec), (i) KonCept512 (1 fr/sec), (j) PaQ-2-PiQ (1 fr/sec), (k) V-BLIINDS, (l) TLVQM, and (m) VIDEVAL.}
\label{fig:kvcv_draw}
\end{figure*}

\section{Experimental Results}
\label{sec:exp}

\subsection{Evaluation Protocol}
\label{ssec:eval_proto}
\textbf{UGC Dataset Benchmarks.} To conduct BVQA performance evaluation, we used the three UGC-VQA databases: KoNViD-1K \cite{hosu2017konstanz}, LIVE-VQC \cite{sinno2018large}, and YouTube-UGC \cite{wang2019youtube}. We found that the YouTube-UGC dataset contains 57 grayscale videos, which yield numerical errors when computing the color model FRIQUEE. Therefore, we extracted a subset of 1,323 color videos from YouTube-UGC, which we denote here as the YouTube-UGC\textsubscript{c} set, for the evaluation of color models. In order to study overall model performances on all the databases, we created a large composite benchmark, which is referred to here as All-Combined\textsubscript{c}, using the iterative nested least squares algorithm (INLSA) suggested in \cite{pinson2003objective}, wherein YouTube-UGC is selected as the anchor set, and the objective MOS from the other two sets, KoNViD-1k and LIVE-VQC, are linearly mapped onto a common scale ($[1,5]$). Figure \ref{fig:inlsa} shows scatter plots of MOS versus NIQE scores before (Figure \ref{fig:inlsa-a}) and after (Figure \ref{fig:inlsa-b}) INLSA linear mapping, calibrated by NIQE \cite{mittal2012making} scores. The All-Combined\textsubscript{c} (3,108) dataset is simply the union of KoNViD-1k (1,200), LIVE-VQC (575), and YouTube-UGC\textsubscript{c} (1,323) after MOS calibration:
\begin{equation}
\label{eq:mos_cal_kon}
y_\mathrm{adj}=5-4\times\left[(5-y_\mathrm{org})/4\times1.1241-0.0993\right]
\end{equation}
\begin{equation}
\label{eq:mos_cal_live}
y_\mathrm{adj}=5-4\times\left[(100-y_\mathrm{org})/100\times0.7132+0.0253\right]
\end{equation}
where (\ref{eq:mos_cal_kon}) and (\ref{eq:mos_cal_live}) are for calibrating KoNViD-1k and LIVE-VQC, respectively. $y_\mathrm{adj}$ denotes the adjusted scores, while $y_\mathrm{org}$ is the original MOS.

\textbf{BVQA Model Benchmarks.} We include a number of representative BVQA/BIQA algorithms in our benchmarking evaluation as references to be compared against. These baseline models include NIQE \cite{mittal2012making}, ILNIQE \cite{zhang2015feature}, VIIDEO \cite{mittal2015completely}, BRISQUE \cite{mittal2012no}, GM-LOG \cite{xue2014blind}, HIGRADE \cite{kundu2017no}, FRIQUEE \cite{ghadiyaram2017perceptual}, CORNIA \cite{ye2012unsupervised}, HOSA \cite{xu2016blind}, KonCept512 \cite{hosu2020koniq}, PaQ-2-PiQ \cite{ying2019patches}, V-BLIINDS \cite{saad2014blind}, and TLVQM \cite{korhonen2019two}. Among these, NIQE, ILNIQE, and VIIDEO are ``completely blind'' (opinion-unaware (OU)), since no training is required to build them. The rest of the models are all training-based (opinion-aware (OA)) and we re-train the models/features when evaluating on a given dataset. We also utilized the well-known deep CNN models VGG-19 \cite{simonyan2014very} and ResNet-50 \cite{he2016deep} as additional CNN-based baseline models, where each was pretrained on the ImageNet classification task. The fully-connected layer (4,096-dim) from VGG-19 and average-pooled layer (2,048-dim) from ResNet-50 served as deep feature descriptors, by operating on 25 227$\times$227 random crops of each input frame, then average-pooled into a single feature vector representing the entire frame \cite{kim2017deep}. Two SOTA deep BIQA models, KonCept512 \cite{hosu2020koniq} and PaQ-2-PiQ \cite{ying2019patches}, were also included in our evaluations. We implemented the feature extraction process for each evaluated BVQA model using its initial released implementation in MATLAB R2018b, except that VGG-19 and ResNet-50 were implemented in TensorFlow, while KonCept512\footnote{\url{https://github.com/ZhengyuZhao/koniq-PyTorch}} and PaQ-2-PiQ\footnote{\url{https://github.com/baidut/paq2piq}} were implemented in PyTorch. All the feature-based BIQA models extract features at a uniform sampling rate of one frame per second, then temporally average-pooled to obtain the overall video-level feature.

\textbf{Regression Models.} We used a support vector regressor (SVR) as the back-end regression model to learn the feature-to-score mappings, since it achieves excellent performance in most cases \cite{korhonen2019two, saad2014blind, kim2017deep, ghadiyaram2017perceptual, mittal2012no, xu2014no}. The effectiveness of SVR, however, largely depends on the selection of its hyperparameters. As recommended in \cite{chang2011libsvm}, we optimized the SVR parameter values $(C,\gamma)$ by a grid-search of $10\times 10$ exponentially growing sequences (in our experiments, we used a grid of  $C=2^1,2^2,...,2^{10},\gamma=2^{-8},2^{-7},...,2^{1}$) using cross-validation on the training set. The pair $(C,\gamma)$ yielding the best cross-validation performance, as measured by the root mean squared error (RMSE) between the predicted scores and the MOS, is picked. Afterward, the selected model parameters are applied to re-train the model on the entire training set, and we report the evaluation results on the test set. This kind of cross-validation procedure can prevent over-fitting, thus providing fair evaluation of the compared BVQA models. We chose the linear kernel for CORNIA, HOSA, VGG-19, and ResNet-50, considering their large feature dimension, and the radial basis function (RBF) kernel for all the other algorithms. We used Python 3.6.7 with the scikit-learn toolbox to train and test all the evaluated learning-based BVQA models.

\textbf{Performance Metrics.} Following convention, we randomly split the dataset into non-overlapping training and test sets ($80\%/20\%$), where the regression model was trained on the training set, and the performance was reported on the test set. This process of random split was iterated 100 times and the overall median performance was recorded. For each iteration, we adopted four commonly used performance criteria to evaluate the models: The Spearman Rank-Order Correlation Coefficient (SRCC) and the Kendall Rank-Order Correlation Coefficient (KRCC) are non-parametric measures of prediction monotonicity, while the Pearson Linear Correlation Coefficient (PLCC) with corresponding Root Mean Square Error (RMSE) are computed to assess prediction accuracy. Note that PLCC and RMSE are computed after performing a nonlinear four-parametric logistic regression to linearize the objective predictions to be on the same scale of MOS \cite{seshadrinathan2010study}.

\subsection{Performance on Individual and Combined Datasets}
\label{ssec:performance_diff_datasets}

Table \ref{table:compete-blind} shows the performance evaluation of the three ``completely blind'' BVQA models, NIQE, ILNIQE, and VIIDEO on the four UGC-VQA benchmarks. None of these methods performed very well, meaning that we still have much room for developing OU ``completely blind'' UGC video quality models.

Table \ref{table:eval_svr} shows the performance evaluation of all the learning-based BVQA models trained with SVR on the four datasets in our evaluation framework. For better visualization, we also show box plots of performances as well as scatter plots of predictions versus MOS on the All-Combined\textsubscript{c} set, in Figures \ref{fig:all_boxplot} and \ref{fig:kvcv_draw}, respectively. Overall, VIDEVAL achieves SOTA or near-SOTA performance on all the test sets. On LIVE-VQC, however, TLVQM outperformed other BVQA models by a notable margin, while it significantly underperformed on the more recent YouTube-UGC database. We observed in Section \ref{ssec:observation} that LIVE-VQC videos generally contain more (camera) motions than KoNViD-1k and YouTube-UGC, and TLVQM computes multiple motion relevant features. Moreover, the only three BVQA models containing temporal features (V-BLIINDS, TLVQM, and VIDEVAL) excelled on LIVE-VQC, which suggests that it is potentially valuable to integrate at least a few, if not many, motion-related features into quality prediction models, when assessing on videos with large (camera) motions.

\begin{table}[!t]
\setlength{\tabcolsep}{4pt}
\renewcommand{\arraystretch}{1.1}
\centering
\caption{Performances on different resolution subsets: 1080p (427), 720p (566), and $\le$480p (448).}
\label{table:resolution_breakdown}
\begin{tabular}{lccccccccccccccccccccccccc}
\toprule
\textsc{Subset} & \multicolumn{2}{c}{1080p} & & \multicolumn{2}{c}{720p}  & & \multicolumn{2}{c}{$\le$480p} \\ \cline{2-3}\cline{5-6}\cline{8-9}\\[-1.em]
\textsc{Model} & SRCC & PLCC & & SRCC &  PLCC & & SRCC & PLCC \\ \hline\\[-1.em]
BRISQUE   & 0.4597 & 0.4637 & & 0.5407 & 0.5585 & & 0.3812 & 0.4065  \\
GM-LOG    & 0.4796 & 0.4970 & & 0.5098 & 0.5172 & & 0.3685 & 0.4200  \\
HIGRADE   & 0.5142 & 0.5543 & & 0.5095 & 0.5324 & & 0.4650 & 0.4642  \\
FRIQUEE   & 0.5787 & 0.5797 & & 0.5369 & 0.5652 & & 0.5042 & 0.5363  \\
CORNIA    & 0.5951 & \textbf{0.6358} & & 0.6212 & 0.6551 & & 0.5631 & 0.6118  \\
HOSA      & 0.5924 & 0.6093 & & \textbf{\underline{0.6651}} & \textbf{0.6739} & & \textbf{0.6514} & \textbf{0.6652}  \\
VGG-19     & \textbf{0.6440} & 0.6090 & & 0.6158 & \textbf{0.6568} & & \textbf{0.5845} & \textbf{0.6267}  \\
ResNet-50  & \textbf{\underline{0.6615}} & \textbf{\underline{0.6644}} & & \textbf{0.6645} & \textbf{\underline{0.7076}} & & \textbf{\underline{0.6570}} & \textbf{\underline{0.6997}}  \\
{KonCept512} & \textbf{0.6332} & \textbf{0.6336} & & 0.6055 & 0.6514 & & 0.4271 & 0.4612 \\
{PaQ-2-PiQ} & 0.5304  & 0.5176 & & 0.5768 & 0.5802 & &  0.3646 & 0.4748  \\
V-BLIINDS & 0.4449 & 0.4491 & & 0.5546 & 0.5719 & & 0.4484 & 0.4752  \\
TLVQM     & 0.5638 & 0.6031 & & \textbf{0.6300} & 0.6526 & & 0.4318 & 0.4784 \\
VIDEVAL    & 0.5805 & 0.6111 & & {0.6296} & {0.6393} & & 0.5014 & 0.5508 \\
\bottomrule
\end{tabular}
\end{table}

\begin{table}[!t]
\setlength{\tabcolsep}{4pt}
\renewcommand{\arraystretch}{1.1}
\centering
\caption{Performances on different content subsets: screen content (163), animation (81), and gaming (209).}
\label{table:content_breakdown}
\begin{tabular}{lccccccccccccccccccccccccc}
\toprule
\textsc{Subset} & \multicolumn{2}{c}{Screen Content} & & \multicolumn{2}{c}{Animation}  & & \multicolumn{2}{c}{Gaming} \\ \cline{2-3}\cline{5-6}\cline{8-9}\\[-1.em]
\textsc{Model} & SRCC & PLCC & & SRCC &  PLCC & & SRCC & PLCC \\ \hline\\[-1.em]
BRISQUE   & 0.2573 & 0.3954 & & 0.0747 & 0.3857 & & 0.2717 & 0.3307  \\
GM-LOG    & 0.3004 & 0.4244 & & 0.2009 & 0.4129 & & 0.3371 & 0.4185  \\
HIGRADE   & 0.4971 & 0.5652 & & 0.1985 & 0.4140 & & 0.6228 & 0.6832  \\
FRIQUEE   & \textbf{0.5522} & \textbf{0.6160} & & 0.2377 & 0.4574 & & \textbf{0.6919} & \textbf{0.7193}  \\
CORNIA    & 0.5105 & 0.5667 & & 0.1936 & 0.4627 & & 0.5741 & 0.6502  \\
HOSA      & 0.4667 & 0.5255 & & 0.1048 & 0.4489 & & 0.6019 & \textbf{0.6998}  \\
VGG-19     & 0.5472 & 0.6229 & & 0.1973 & 0.4700 & & 0.5765 & 0.6370  \\
ResNet-50  & \textbf{\underline{0.6199}} & \textbf{\underline{0.6676}} & & \textbf{0.2781} & \textbf{0.4871} & & \textbf{0.6378} & 0.6779  \\
{KonCept512} & 0.4714 & 0.5119 & & \textbf{0.2757} & \textbf{0.5229} & & 0.4780 & 0.6240  \\
{PaQ-2-PiQ} & 0.3231 & 0.4312 & & 0.0208 & 0.4630 & & 0.2169 & 0.3874 \\
V-BLIINDS & 0.3064 & 0.4155 & & 0.0379 & 0.3917 & & 0.5473 & 0.6101  \\
TLVQM     & 0.3843 & 0.4524 & & 0.2708 & 0.4598 & & 0.5749 & 0.6195 \\
VIDEVAL    & \textbf{{0.6033}} & \textbf{{0.6610}} & & \textbf{\underline{0.3492}}  & \textbf{\underline{0.5274}} & & \textbf{\underline{0.6954}} & \textbf{\underline{0.7323}}  \\
\bottomrule
\end{tabular}
\end{table}

It is also worth mentioning that the deep CNN baseline methods (VGG-19 and ResNet-50), despite being trained as picture-only models, performed quite well on KoNViD-1k and All-Combined\textsubscript{c}. This suggests that transfer learning is a promising technique for the blind UGC-VQA problem, consistent with conclusions drawn for picture-quality prediction \cite{kim2017deep}. Deep models will perform even better, no doubt, if trained on temporal content and distortions. 

{The two most recent deep learning picture quality models, PaQ-2-PiQ, and KonCept512, however, did not perform very well on the three evaluated video datasets. The most probable reason would be that these models were trained on picture quality datasets \cite{ying2019patches, hosu2020koniq}, which contain different types of (strictly spatial) distortions than UGC-VQA databases. Models trained on picture quality sets do not necessarily transfer very well to UGC video quality problems. In other words, whatever model should be either trained or fine-tuned on UGC-VQA datasets in order to obtain reasonable performance. Indeed, if temporal distortions (like judder) are present, they may severely underperform if the frame quality is high \cite{madhusudana2020subjective}.}

\begin{table}[!t]
\setlength{\tabcolsep}{4pt}
\renewcommand{\arraystretch}{1.1}
\centering
\caption{Performances on different quality subsets: low quality (1558) and high quality (1550).}
\label{table:quality_breakdown}
\begin{tabular}{lccccccccccccccccccccccccc}
\toprule
\textsc{Subset} & \multicolumn{2}{c}{Low Quality} & & \multicolumn{2}{c}{High Quality} \\ \cline{2-3}\cline{5-6}\\[-1.em]
\textsc{Model} & SRCC & PLCC & & SRCC &  PLCC  \\ \hline\\[-1.em]
BRISQUE   & 0.4312 & 0.4593 & & 0.2813 & 0.2979   \\
GM-LOG    & 0.4221 & 0.4715 & & 0.2367 & 0.2621   \\
HIGRADE   & 0.5057 & 0.5466 & & 0.4714 & 0.4799   \\
FRIQUEE   & \textbf{0.5460} & \textbf{0.5886} & & \textbf{0.5061} & \textbf{0.5152}   \\
CORNIA    & 0.4931 & 0.5435 & & 0.3610 & 0.3748   \\
HOSA      & \textbf{0.5348} & \textbf{0.5789} & & 0.4208 & 0.4323   \\
VGG-19     & 0.3710 & 0.4181 & & 0.3522 & 0.3614   \\
ResNet-50  & 0.3881 & 0.4250 & & 0.2791 & 0.3030   \\
{KonCept512} & 0.3428 & 0.4497 & & 0.2245 & 0.2597 & \\
{PaQ-2-PiQ} & 0.2438 & 0.2713 & & 0.2013 & 0.2252 \\
V-BLIINDS & 0.4703 & 0.5060 & & 0.3207 & 0.3444   \\
TLVQM     & 0.4845 & 0.5386 & & \textbf{0.4783} & \textbf{0.4860}   \\
VIDEVAL   & \textbf{\underline{0.5680}} & \textbf{\underline{0.6056}} & & \textbf{\underline{0.5546}} & \textbf{\underline{0.5657}} \\
\bottomrule
\end{tabular}
\end{table}

\begin{table}[!t]
\setlength{\tabcolsep}{3.5pt}
\renewcommand{\arraystretch}{1.1}
\centering
\caption{Best model in terms of SRCC for cross dataset generalization evaluation.}
\label{table:cross_dataset_srcc}
\begin{tabular}{lcccccccc}
\toprule
\textsc{Train}\textbackslash\textsc{Test}  & LIVE-VQC &  KoNViD-1k &  YouTube-UGC\textsubscript{c} \\
\hline\\[-1.em]
 LIVE-VQC & - &  ResNet-50 (0.69)  &  ResNet-50 (0.33)   \\
 KoNViD-1k & ResNet-50 (0.70) & - & VIDEVAL (0.37)  \\
 YouTube-UGC\textsubscript{c} & HOSA (0.49)  &  VIDEVAL (0.61)  & -    \\
\bottomrule
\end{tabular}
\end{table}

\begin{table}[!t]
\setlength{\tabcolsep}{3.5pt}
\renewcommand{\arraystretch}{1.}
\centering
\caption{Best model in terms of PLCC for cross dataset generalization evaluation.}
\label{table:cross_dataset_plcc}
\begin{tabular}{lcccccccc}
\toprule
\textsc{Train}\textbackslash\textsc{Test}  & LIVE-VQC &  KoNViD-1k &  YouTube-UGC\textsubscript{c} \\
\hline\\[-1.em]
 LIVE-VQC & - &   ResNet-50 (0.70)  & VIDEVAL (0.35)    \\
 KoNViD-1k & ResNet-50 (0.75) & - & VIDEVAL (0.39)  \\
 YouTube-UGC\textsubscript{c} & HOSA (0.50)  &  VIDEVAL (0.62)  & -    \\
\bottomrule
\end{tabular}
\end{table}

\subsection{Performance Evaluation on Categorical Subsets}

We propose three new categorical evaluation methodologies - resolution, quality, and content-based category breakdown. These will allow us to study the compared BVQA models from additional and practical aspects in the context of real-world UGC scenarios, which have not been, nor can it be accounted in previous legacy VQA databases or studies.

\begin{table*}[!t]
\setlength{\tabcolsep}{4pt}
\renewcommand{\arraystretch}{1.1}
\centering
\caption{{Performance comparison of a total of eleven temporal pooling methods using TLVQM and VIDEVAL as testbeds on KoNViD-1k, LIVE-VQC, and YouTube-UGC. The three best results along each column are \textbf{boldfaced}.}}
\label{table:pooling}
\begin{tabular}{lccccccccccccccccccccccc}
\toprule
\textsc{Database} & \multicolumn{5}{c}{KoNViD-1k} & & \multicolumn{5}{c}{LIVE-VQC} & & \multicolumn{5}{c}{YouTube-UGC} \\ \cline{2-6}\cline{8-12}\cline{14-18}\\[-1.em]

\textsc{Model} & \multicolumn{2}{c}{TLVQM} & & \multicolumn{2}{c}{VIDEVAL} & & \multicolumn{2}{c}{TLVQM} & & \multicolumn{2}{c}{VIDEVAL} & & \multicolumn{2}{c}{TLVQM} & & \multicolumn{2}{c}{VIDEVAL} \\
\cline{2-3}\cline{5-6}\cline{8-9}\cline{11-12}\cline{14-15}\cline{17-18}\\[-1.em]

\textsc{Pooling} & SRCC & PLCC & & SRCC &  PLCC & & SRCC & PLCC & & SRCC & PLCC & & SRCC & PLCC & & SRCC & PLCC \\ \hline\\[-1.em]

Mean & \textbf{0.7511} & \textbf{0.7475} & & 0.7749 & \textbf{0.7727} & & \textbf{0.7917} &  \textbf{0.7984} & &  \textbf{0.7396} &  0.7432 & & \textbf{0.6369} & \textbf{0.6310} & & 0.7447 & 0.7332\\
Median & 0.7483 & 0.7437 & & 0.7650 & 0.7698 & & 0.7708 & 0.7887 & & 0.7236 & 0.7308 & & 0.6127 & 0.6090 & & 0.7452 & \textbf{0.7448} \\
Harmonic & 0.7458 & 0.7392 & & \textbf{0.7772} & 0.7681 & &  0.7845 & 0.7890 & & 0.7312 & 0.7250 & & 0.6119 & 0.6038 & & 0.7449 & 0.7318 \\
Geometric & 0.7449 & 0.7461 & & 0.7566 & 0.7592 & &  \textbf{0.7878} & \textbf{0.7964} & & \textbf{0.7412} & 0.7487 & & 0.6347 & 0.6236 & & \textbf{0.7508} & \textbf{0.7437} \\
Minkowski & 0.7498 & \textbf{0.7481} & & \textbf{0.7775} & \textbf{0.7727} & &  0.7863 & 0.7908 & & 0.7371 & \textbf{0.7558} & & \textbf{0.6368} & \textbf{0.6311} & & \textbf{0.7542} & \textbf{0.7508} \\
Percentile & 0.7078 & 0.7000 & & 0.7161 & 0.7049 & &  0.7378 & 0.7313 & & 0.6596 & 0.6576 & & 0.4871 & 0.4996 & & 0.6443 & 0.6465 \\
VQPooling & 0.7240 & 0.7196 & & 0.7366 & 0.7296 & & 0.7696 & 0.7895 & & 0.7240 & 0.7311 & & 0.5654 & 0.5618 & & 0.6942 & 0.6862 \\
Primacy & 0.7456 & 0.7451 & & 0.7711 & 0.7700 & & 0.7751 & 0.7851 & & 0.7349 & \textbf{0.7523}  & & 0.5734 & 0.5692 & & 0.7221 & 0.7156 \\
Recency & \textbf{0.7528} & 0.7470 & & 0.7683 & 0.7677 & & 0.7715 & 0.7857 & & \textbf{0.7405} & \textbf{0.7584} & & 0.5821 & 0.5695 & & 0.7176 & 0.7116 \\
Hysteresis & 0.7434 & 0.7430 & & 0.7612 & 0.7554  & & 0.7856 & 0.7901 & & 0.7226 & 0.7433 & & 0.6092 & 0.6109 & & 0.7370 & 0.7306 \\
EPooling & \textbf{0.7641} & \textbf{0.7573} & & \textbf{0.7831} & \textbf{0.7867} & & \textbf{0.7925} & \textbf{0.7917} & & 0.7371 & 0.7372 & & \textbf{0.6452} & \textbf{0.6592} & & \textbf{0.7517} & 0.7379 \\
\toprule
\end{tabular}
\end{table*}

For resolution-dependent evaluation, we divided the All-Combined\textsubscript{c} set into three subsets, based on video resolution: (1) 427 1080p-videos (110 from LIVE-VQC, 317 from YouTube-UGC), (2) 566 720p-videos (316 from LIVE-VQC, 250 from YouTube-UGC), and (3) 448 videos with resolution $\le$480p (29 from LIVE-VQC, 419 from YouTube-UGC), since we are also interested in performance on videos of different resolutions. We did not include 540p-videos, since those videos are almost exclusively from KoNViD-1k. Table \ref{table:resolution_breakdown} shows the resolution-breakdown evaluation results. Generally speaking, learned features (CORNIA, HOSA, VGG-19, KonCept512, and ResNet-50) outperformed hand-designed features, among which ResNet-50 ranked first.

\begin{figure}[!t]
\centering
\footnotesize
\def\xwidth{0.32}
\def\hswidth{0ex}
\def\rdshift{0.3264\linewidth}
\begin{tabular}{cc}
\multirow{1}{*}[\rdshift]{\includegraphics[height=\xwidth\linewidth]{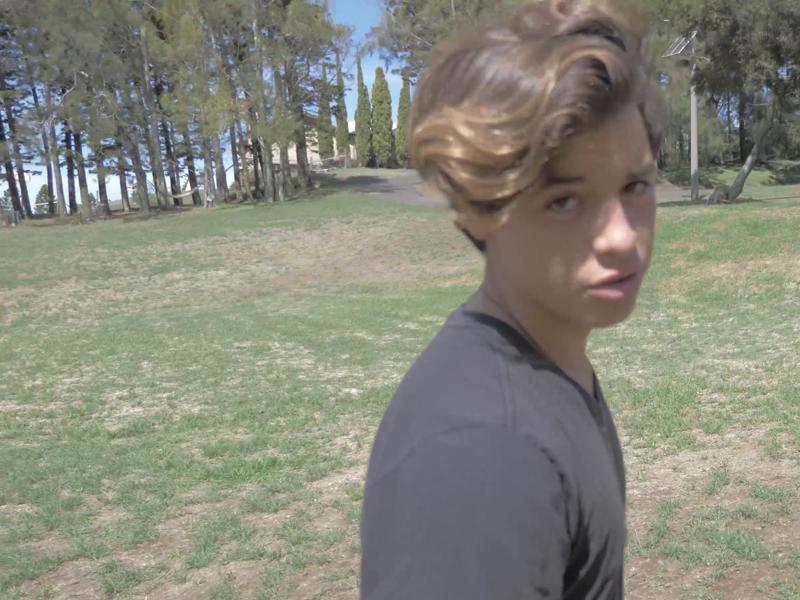}} & 
{\includegraphics[height=0.3668\linewidth]{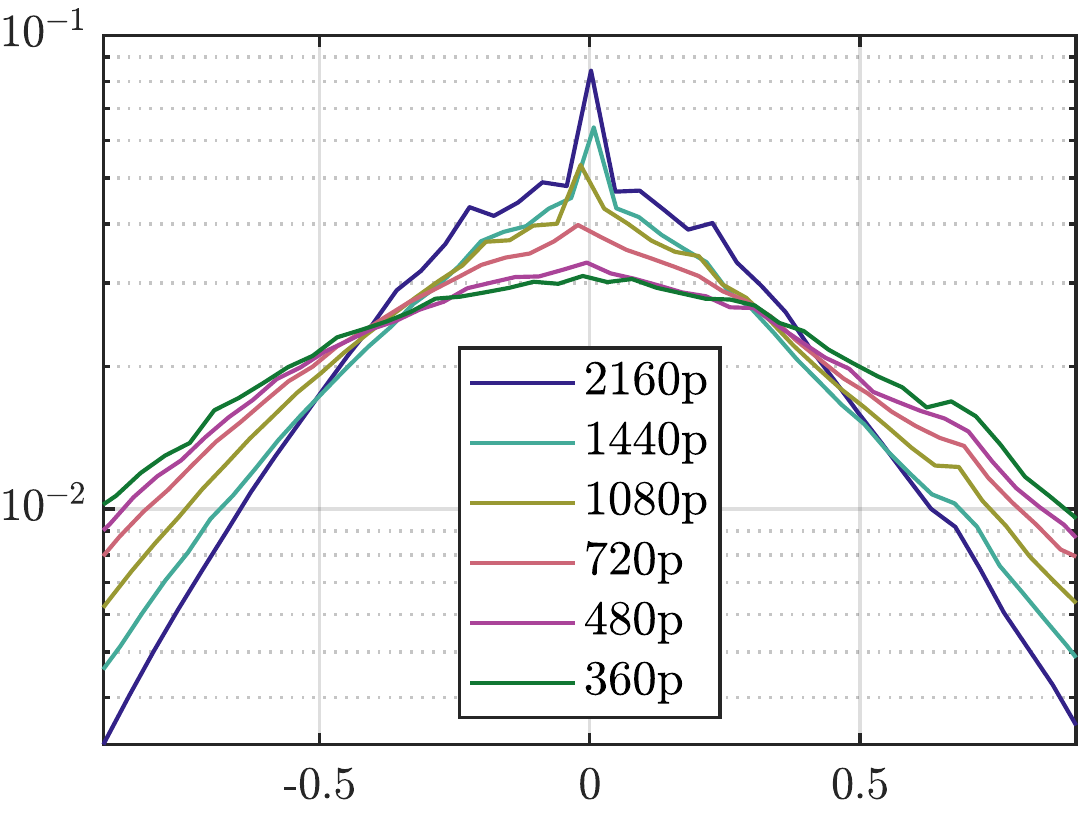}} \\ 
(a) Vlog\_2160P-408f.mkv & (b) MSCN distributions \\
\end{tabular}
\caption{{(a) An examplary 2160p video from YouTube-UGC and (b) the mean-subtracted contrast-normalized (MSCN) distributions of its downscaled versions: 2160p, 1440p, 1080p, 720p, 480p, and 360p.}}
\label{fig:nss}
\end{figure}

{Here we make two arguments to try to explain the observations above: (1) video quality is intrinsically correlated with resolution; (2) NSS features are implicitly \textit{resolution-aware}, while CNN features are not. The first point is almost self-explanatory, no matter to what degree one agrees. To further justify this, we trained an SVR only using resolution (height, width) as features to predict MOS on YouTube-UGC, which contains balanced samples across five different resolutions. This yielded surprisingly high values $0.576 / 0.571$ for SRCC$/$PLCC, indicating the inherent correlation between video quality and resolution. Secondly, we selected one 2160p video from YouTube-UGC, namely `Vlog2160P-408f.mkv,' and plotted, in Figure \ref{fig:nss}, the mean-subtracted contrast-normalized (MSCN) distributions of its downscaled versions: 2160p, 1440p, 1080p, 720p, 480p, and 360p. It may be observed that resolution can be well separated by MSCN statistics, based on which most feature-based methods are built. We may infer, from these two standpoints, that including various resolutions of videos is favorable to the training of NSS-based models, since NSS features are resolution-aware, and resolution is further well correlated with quality. In other words, the resolution-breakdown evaluation shown in Table \ref{table:resolution_breakdown}, which removes this important implicit feature (resolution), would possibly reduce the performance of NSS-based models, such as FRIQUEE and VIDEVAL.}

\begin{table*}[!t]
\setlength{\tabcolsep}{4pt}
\renewcommand{\arraystretch}{1.1}
\centering
\begin{threeparttable}
\caption{Feature description, dimensionality, computational complexity, and average runtime comparison (in seconds evaluated on twenty $1080p$ videos from LIVE-VQC) among MATLAB-implemented BVQA models.}
\label{table:complexity}
\begin{tabular}{llp{5.5cm}cp{5.5cm}c}
\toprule
\textsc{Class} & \textsc{Model} & \textsc{Feature Description} & \textsc{Dim} & \textsc{Computational Complexity} & \textsc{Time (sec)} \\
\hline\\[-1.em]
\multirow{15}{*}{\shortstack{IQA}} & NIQE (1 fr/sec) & Spatial NSS & 1 & $\mathcal{O}(d^2 NT)$ $d$: window size  &  6.3 \\
 & ILNIQE (1 fr/sec) & {Spatial NSS, gradient, log-Gabor, and color statistics} & 1 & {$\mathcal{O}((d^2+h+gh)NT)$ $d$: window size; $h$: filter size; $g$: log-Gabor filter size} & 23.3 \\
 & BRISQUE (1 fr/sec) & Spatial NSS & 36 & $\mathcal{O}(d^2 NT)$ $d$: window size  &  1.7 \\
& GM-LOG (1 fr/sec) & Joint statistics of gradient magnitude and laplacian of gaussian coefficients & 40 & {$\mathcal{O}(((h+k)NT)$ $d$: window size; $k$: probability matrix size}  &  2.1 \\
& HIGRADE (1 fr/sec) & Spatial NSS, and gradient magnitude statistics in LAB color space & 216 & {$\mathcal{O}(3(2d^2+k)NT)$ $d$: window size; $k$: gradient kernel size} & 11.6 \\
& FRIQUEE (1 fr/sec) & Complex streerable pyramid wavelet, luminance, chroma, LMS, HSI, yellow channel, and their transformed domain statistics & 560 & $\mathcal{O}((fd^2 N+4N(\log(N)+m^2))T)$ $d$: window size; $f$: number of color spaces; $m$: neighborhood size in DNT  & 701.2 \\
 & CORNIA (1 fr/sec) & Spatially normalized image patches and max min pooling & 10k & $\mathcal{O}(d^2 KNT)$ $d$: window size $K$: codebook size  & 14.3 \\
  & HOSA (1 fr/sec) & Local normalized image patches based on high order statistics aggregation & 14.7k & $\mathcal{O}(d^2 KNT)$ $d$: window size $K$: codebook size  & 1.2 \\ \\[-1.1em]
\hline\\[-1.em]
\multirow{10}{*}{\shortstack{VQA}}    & VIIDEO  & Frame difference spatial statistics, inter sub-band statistics  & 1 & $\mathcal{O}(N\log(N)T)$ & 674.8 \\
  & V-BLIINDS  & Spatial NSS, frame difference DCT coefficient statistics, motion coherency, and egomotion & 47 & $\mathcal{O}((d^2 N+\log(k)N+k^2 w^3)T)$ $d$: window size; $k$: block size; $w$: motion vector tensor size  & 1989.9 \\
& TLVQM & Captures impairments computed at two computation levels: low complexity and high complexity features & 75 & $\mathcal{O}((h_1^2 N+k^2 K)T_1+(\log(N)+h_2^2) NT_2))$ $h_1,h_2$: filter size; $k$: motion estimation block size; $K$: number of key points & 183.8  \\
& VIDEVAL & Selected combination of NSS features in multiple perceptual spaces and using visual impairment features from TLVQM & 60 & $\mathcal{O}((fh_1^2N+k^2K)T_1+h_2^2NT_2)$ $h_1,h_2$: filter size; $f$: number of color spaces; $k$: motion estimation block size; $K$: number of key points & 305.8 \\

\bottomrule
\end{tabular}
\begin{tablenotes}[para,flushleft]
\item $N$: number of pixels per frame; $T$: number of frames computed for feature extraction. Note that for VIIDEO and V-BLIINDS, $T$ is the total number of frames, whereas for IQA models, $T$ equals the total number of frames sampled at 1 fr/sec. For TLVQM and VIDEVAL, $T_1$ is total number of frames divided by 2, while $T_2$ is the number of frames sampled at 1 fr/sec.
\end{tablenotes}
\end{threeparttable}
\end{table*}

We also divided the All-Combined\textsubscript{c} into subsets based on content category: Screen Content (163), Animation (81), Gaming (209), and Natural (2,667) videos. We only reported the evaluation results on the first three subsets in Table \ref{table:content_breakdown}, since we observed similar results on the Natural subset with the entire combined set. The proposed VIDEVAL model outperformed over all categories, followed by ResNet-50 and FRIQUEE, suggesting that VIDEVAL features are robust quality indicatives across different content categories.

The third categorical division is based on quality scores: we partitioned the combined set into Low Quality (1,558) and High Quality (1,550) halves, using the median quality value 3.5536 as the threshold, to see the model performance only  on high/low quality videos. Performance results are shown in Table \ref{table:quality_breakdown}, wherein VIDEVAL still outperformed the other BVQA models on both low and high quality partitions.

\subsection{Cross Dataset Generalizability}

We also performed a cross dataset evaluation to verify the generalizability of BVQA models, wherein LIVE-VQC, KoNViD-1k, and YouTube-UGC\textsubscript{c} were included. That is, we trained the regression model on one full database and report the performance on another. To retain label consistency, we linearly scaled the MOS values in LIVE-VQC from raw $[0,100]$ to $[1,5]$, which is the scale for the other two datasets. We used SVR for regression and adopted $k$-fold cross validation using the same grid-search as in Section \ref{ssec:eval_proto} for hyperparameter selection. The selected parameter pair were then applied to re-train the SVR model on the full training set, and the performance results on the test set were recorded. Table \ref{table:cross_dataset_srcc} and \ref{table:cross_dataset_plcc} show the best performing methods with cross domain performances in terms of SRCC and PLCC, respectively.

We may see that the cross domain BVQA algorithm generalization between LIVE-VQC and KoNViD-1k was surprisingly good, and was well characterized by pre-trained ResNet-50 features. We also observed better algorithm generalization between KoNViD-1k and YouTube-UGC than LIVE-VQC, as indicated by the performances of the best model, VIDEVAL. This might be expected, since as Figure \ref{fig:tsne} shows, YouTube-UGC and KoNViD-1k share overlapped coverage of content space, much larger than that of LIVE-VQC. Therefore, we may conclude that VIDEVAL and ResNet-50 were the most robust BVQA models among those compared in terms of cross domain generalization capacities.

\subsection{Effects of Temporal Pooling}

Temporal pooling is one of the most important, unresolved problems for video quality prediction \cite{park2012video, tu2020comparative,seshadrinathan2011temporal, korhonen2019two, bampis2018recurrent}. In our previous work \cite{tu2020comparative}, we have studied the efficacy of various pooling methods using scores predicted by BIQA models. Here we extend this to evaluate on SOTA BVQA models. For practical considerations, the high-performing TLVQM and VIDEVAL were selected as exemplar models. Since these two models independently extract features on each one-second block, we applied temporal pooling of chunk-wise quality predictions. A total of eleven pooling methods were tested: three Pythagorean means (arithmetic, geometric, and harmonic mean), median, Minkowski ($p=2$) mean, percentile pooling ($20\%$) \cite{moorthy2009visual}, VQPooling \cite{park2012video}, primacy and recency pooling \cite{murdock1962serial}, hysteresis pooling \cite{seshadrinathan2011temporal}, and our previously proposed ensemble method, EPooling \cite{tu2020comparative}, which aggregates multiply pooled scores by training a second regressor on top of mean, Minkowski, percentile, VQPooling, variation, and hysteresis pooling. We refer the reader to \cite{tu2020comparative} for detailed algorithmic formulations and parameter settings thereof.

It is worth noting that the results in Table \ref{table:pooling} are only \textit{self-consistent}, meaning that they are not comparable to any prior experiments - since we employed chunk-wise instead of previously adopted video-wise quality prediction to be able to apply temporal quality pooling, which may affect the base performance. Here we observed yet slightly different results using BVQA testbeds as compared to what we observed on BIQA \cite{tu2020comparative}. Generally, we found the mean families and ensemble pooling to be the most reliable pooling methods. Traditional sample mean prediction may be adequate in many cases, due to its simplicity. Pooling strategies that more heavily weight low-quality parts, however, were not observed to perform very well on the tested BVQA, which might be attributed to the fact that not enough samples ($8\sim 20$) can be extracted from each video to attain statistically meaningful results.

\subsection{Complexity Analysis and Runtime Comparison}

The efficiency of a video quality model is of vital importance in practical commercial deployments. Therefore, we also tabulated the computational complexity and runtime cost of the compared BVQA models, as shown in Tables \ref{table:complexity}, \ref{table:time_dl}. The experiments were performed in MATLAB R2018b and Python 3.6.7 under Ubuntu 18.04.3 LTS system on a Dell OptiPlex 7080 Desktop with Intel Core i7-8700 CPU@3.2GHz, 32G RAM, and GeForce GTX 1050 Graphics Cards. The average feature computation time of MATLAB-implemented BVQA models on 1080p videos are reported in Table \ref{table:complexity}. The proposed VIDEVAL method achieves a reasonable complexity among the top-performing algorithms, TLVQM, and FRIQUEE. We also present theoretical time complexity in Table \ref{table:complexity} for potential analytical purposes.

{We also provide in Table \ref{table:time_dl} an additional runtime comparison between MATLAB models on CPU and deep learning models on CPU and GPU, respectively. It may be observed that top-performing BVQA models such as TLVQM and VIDEVAL are essentially slower than deep CNN models, but we expect orders-of-magnitude speedup if re-implemented in pure $\text{C}/\text{C}\texttt{++}$. Simpler NSS-based models such as BRISQUE and HIGRADE (which only involve several convolution operations) still show competitive efficiency relative to CNN models even when implemented in MATLAB. We have also seen a $5\sim 10$ times speedup switching from CPU to GPU for the CNN models, among which KonCept512 with PyTorch-GPU was the fastest since it requires just a single pass to the CNN backbone, while the other three entail multiple passes for each input frame.}

Note that the training/test time of the machine learning regressor is approximately proportional to the number of features. Thus, it is not negligible compared to feature computation given a large number of features, regardless of the regression model employed. The feature dimension of each model is listed in Table \ref{table:complexity}. As may be seen, codebook-based algorithms (CORNIA (10$k$) and HOSA (14.7$k$)) require significantly larger numbers of features than other hand-crafted feature based models. Deep ConvNet features ranked second in dimension (VGG-19 (4,080) and ResNet-50 (2,048)). Our proposed VIDEVAL only uses 60 features, which is fairly compact, as compared to other top-performing BVQA models like FRIQUEE (560) and TLVQM (75).

\begin{table}[!t]
\setlength{\tabcolsep}{4pt}
\renewcommand{\arraystretch}{1.1}
\centering
\caption{{Run time comparison of feature-based and deep learning BVQA models (in seconds evaluated on twenty $1080p$ videos from LIVE-VQC). Model loading time for deep models are excluded}.}
\label{table:time_dl}
\begin{tabular}{lrrccccccccccccccccccccccc}
\toprule
\textsc{Model} &  & \textsc{Time (Sec)} \\ \hline\\[-1.em]
BRISQUE (1 fr/sec) & MATLAB-CPU & 1.7 \\
HOSA (1 fr/sec) & MATLAB-CPU & 1.2 \\
TLVQM            & MATLAB-CPU & 183.8 \\
VIDEVAL         & MATLAB-CPU & 305.8 \\
\hline\\[-1.em]
VGG-19 (1 fr/sec) & TensorFlow-CPU  & 27.8 \\
                & TensorFlow-\textit{GPU} & 5.7 \\
ResNet-50 (1 fr/sec) & TensorFlow-CPU  & 9.6 \\
                & TensorFlow-\textit{GPU} & 1.9 \\
\hline\\[-1.em]
KonCept512 (1 fr/sec) & PyTorch-CPU &  2.8 \\
                      & PyTorch-\textit{GPU} & 0.3 \\
PaQ-2-PiQ (1 fr/sec) & PyTorch-CPU & 6.9 \\
                     & PyTorch-\textit{GPU} & 0.8 \\
\bottomrule
\end{tabular}
\end{table}

\subsection{{Ensembling VIDEVAL with Deep Features}}
\label{ssec:ensemble}

We also attempted a more sophisticated ensemble fusion of VIDEVAL and deep learning features to determine whether this could further boost its performance, which could give insights on the future direction of this field. Since PaQ-2-PiQ aimed for local quality prediction, we included the predicted $3\times 5$ local quality scores as well as a single global score, as additional features. For KonCept512, the feature vector (256-dim) immediately before the last linear layer in the fully-connected head was appended. Our own baseline CNN models, VGG-19 and ResNet-50, were also considered, because these are commonly used standards for downstream vision tasks.

The overall results are summarized in Table \ref{table:fusion}. We may observe that ensembling VIDEVAL with certain deep learning models improved the performance by up to $\sim 4\%$ compared to the vanilla VIDEVAL, which is very promising. Fusion with either ResNet-50 or KonCept512 yielded top performance. It should be noted that the number of fused features is also an essential aspect. For example, blending VIDEVAL (60-dim) with VGG-19 (4,096-dim) may not be recommended, since the enormous number of VGG-19 features could possibly dominate the VIDEVAL features, as suggested by some performance drops in Table \ref{table:fusion}.

\begin{table}[!t]
\setlength{\tabcolsep}{3.5pt}
\renewcommand{\arraystretch}{1.1}
\centering
\caption{Performance of the ensemble VIDEVAL models fused with additional deep learning features.}
\label{table:fusion}
\begin{tabular}{llcccc}
\toprule
\textsc{Dataset}  & \textsc{Model} \textbackslash\ \textsc{Metric} & SRCC   & KRCC    & PLCC     & RMSE    \\
\hline\\[-1.em]
\multirow{5}{*}{KoNViD} & 
VIDEVAL  & 0.7832 & 0.5845 & 0.7803 & 0.4024 \\
& VIDEVAL$+$VGG-19 & 0.7827 & 0.5928 & 0.7913 & 0.3897   \\
& VIDEVAL$+$ResNet-50 & 0.8129 & 0.6212 & \textbf{0.8200} & \textbf{0.3659} \\
& VIDEVAL$+$KonCept512  & \textbf{0.8149} & \textbf{0.6251} & 0.8169 & 0.3670  \\
& VIDEVAL$+$PaQ-2-PiQ  & 0.7844 & 0.5891 & 0.7793 & 0.4018  \\
\hline\\[-1.em]
\multirow{5}{*}{LIVE-VQC}   & 
VIDEVAL  & 0.7522 & 0.5639 & 0.7514 & 11.100 \\
& VIDEVAL$+$VGG-19  & 0.7274 & 0.5375 & 0.7717 & 10.749  \\
& VIDEVAL$+$ResNet-50 & 0.7456 & 0.5555 & 0.7810 & 10.385 \\
& VIDEVAL$+$KonCept512  & \textbf{0.7849} & \textbf{0.5953} & \textbf{0.8010} & \textbf{10.145}  \\
& VIDEVAL$+$PaQ-2-PiQ    & 0.7677  & 0.5736  & 0.7686 & 10.787  \\

\hline\\[-1.em]
\multirow{5}{*}{YT-UGC} & 
VIDEVAL  & 0.7787 & 0.5830 & 0.7733 & 0.4049 \\
& VIDEVAL$+$VGG-19 & 0.7868 & 0.5930 & 0.7847 & 0.3993  \\
& VIDEVAL$+$ResNet-50 & \textbf{0.8085} & 0.6128 & \textbf{0.8033} & \textbf{0.3837} \\
& VIDEVAL$+$KonCept512 & 0.8083 & \textbf{0.6139} & 0.8028 & 0.3859 \\
& VIDEVAL$+$PaQ-2-PiQ & 0.7981 & 0.6015 & 0.7941 & 0.3959  \\

\hline\\[-1.em]
\multirow{5}{*}{All-Comb} &
VIDEVAL  & 0.7960 & 0.6032 & 0.7939 & 0.4268 \\
& VIDEVAL$+$VGG-19 & 0.7859 & 0.5912 & 0.7962 & 0.4202   \\
& VIDEVAL$+$ResNet-50 & 0.8115 & \textbf{0.6207} & \textbf{0.8286} & \textbf{0.3871} \\
& VIDEVAL$+$KonCept512 & \textbf{0.8123} & 0.6193  & 0.8168  & 0.4017  \\
& VIDEVAL$+$PaQ-2-PiQ & 0.7962  & 0.5991 & 0.7934  & 0.4229 \\

\toprule
\end{tabular}
\end{table}

\subsection{Summarization and Takeaways}

Finally, we briefly summarize the experimental results and make additional observations:

\begin{enumerate}
    \item Generally, spatial distortions dominated quality prediction on Internet UGC videos like those from YouTube and Flickr, as revealed by the remarkable performances of picture-only models (e.g., HIGRADE, FRIQUEE, HOSA, ResNet-50) on them. Some motion-related features (as in TLVQM) may not apply as well in this scenario.
    \item On videos captured with mobile devices (e.g., those in LIVE-VQC), which often present larger and more frequent camera motions, including temporal- or motion-related features can be advantageous (e.g., V-BLIINDS, TLVQM, VIDEVAL).
    \item Deep CNN feature descriptors (VGG-19, ResNet-50, etc.) pre-trained for other classical vision tasks (e.g. image classification) are transferable to UGC video quality predictions, achieving very good performance, suggesting that using transfer learning to address the general UGC-VQA problem is very promising.
    \item It is still a very hard problem to predict UGC video quality on non-natural or computer-generated video contents: screen contents, animations, gaming, etc. Moreover, there are no sufficiently large UGC-VQA datasets designed for those kinds of contents.
    \item A simple feature engineering and selection implementation built on top of current effective feature-based BVQA models is able to obtain excellent performance, as exemplified by the compact new model (VIDEVAL).
    \item Simple temporal mean pooling of chunk-wise quality predictions by BVQA models yields decent and robust results. Furthermore, an ensemble pooling approach can noticeably improve the quality prediction performance, albeit with higher complexity.
    \item Ensembling scene statistics-based BVQA models with additional deep learning features (e.g., VIDEVAL plus KonCept512) could further raise the performance upper bound, which may be a promising way of developing future BVQA models.
\end{enumerate}

\section{Conclusion}
\label{sec:conclud}
We have presented a comprehensive analysis and empirical study of blind video quality assessment for user-generated content (the \textbf{\textsf{UGC-VQA problem}}). We also proposed a new fusion-based BVQA model, called the VIDeo quality EVALuator (VIDEVAL), which uses a feature ensemble and selection procedure on top of existing efficient BVQA models. A systematic evaluation of prior leading video quality models was conducted within a unified and reproducible evaluation framework and accordingly, we concluded that a selected fusion of simple distortion-aware statistical video features, along with well-defined visual impairment features, is able to deliver state-of-the-art, robust performance at a very reasonable computational cost. The promising performances of baseline CNN models suggest the great potential of leveraging transfer learning techniques for the UGC-VQA problem. We believe that this benchmarking study will help facilitate UGC-VQA research by clarifying the current status of BVQA research and the relative efficacies of modern BVQA models. To promote reproducible research and public usage, an implementation of VIDEVAL has been made available online: \textbf{\url{https://github.com/vztu/VIDEVAL}}. In addition to the software, we are also maintaining an ongoing performance leaderboard on Github: \textbf{\url{https://github.com/vztu/BVQA_Benchmark}}.


%





\ifCLASSOPTIONcaptionsoff
  \newpage
\fi



%

\bibliographystyle{IEEEtran}
\bibliography{ref}

\end{document}